\documentclass{article} %
\usepackage{iclr2027_conference,times}

\usepackage{amsmath,amsfonts,bm}

\def\eqref#1{equation~\ref{#1}}

\def\1{\bm{1}}

\DeclareMathAlphabet{\mathsfit}{\encodingdefault}{\sfdefault}{m}{sl}
\SetMathAlphabet{\mathsfit}{bold}{\encodingdefault}{\sfdefault}{bx}{n}

\usepackage{hyperref}
\usepackage{url}
\newcommand{\thetas}[0]{\boldsymbol{\theta}}
\newcommand{\betas}[0]{\boldsymbol{\beta}}
\newcommand{\transl}[0]{\mathbf{t}}

\newcommand{\ours}[0]{InHabit}
\newcommand{\ourdata}[0]{\emph{InHabitants}}
\newcommand{\posa}[0]{POSA++}
\definecolor{Fgreen}{HTML}{309c59}

\newcommand{\TODO}[1]{\textbf{\color{red}[TODO: #1]}}

\usepackage{comment}
\usepackage{siunitx}
\usepackage{multirow,bigdelim}
\usepackage{color}
\usepackage{pbox}
\usepackage{wrapfig}
\usepackage{csquotes}
\usepackage{booktabs}
\usepackage{pifont}
\usepackage{makecell}
\usepackage{tabularx}
\usepackage{rotating}
\usepackage{xspace}
\usepackage{cleveref}
\crefname{figure}{Fig.}{Figs.}\Crefname{figure}{Fig.}{Figs.}
\crefname{table}{Tab.}{Tabs.}\Crefname{table}{Tab.}{Tabs.}
\crefname{section}{Sec.}{Secs.}\Crefname{section}{Sec.}{Secs.}
\crefname{appendix}{Sec.}{Secs.}\Crefname{appendix}{Sec.}{Secs.}
\crefname{equation}{Eq.}{Eqs.}\Crefname{equation}{Eq.}{Eqs.}
\usepackage{tikz}
\newcommand{\eg}{e.g.\@\xspace}

\newenvironment{teaserfigure}{\begin{figure}[!h]\centering}{\end{figure}}

\renewcommand{\TODO}[1]{}

 \renewcommand{\paragraph}[1]{\vspace{.2em}\noindent\textbf{#1.}}

\hypersetup{colorlinks=true, citecolor=blue!60!black, linkcolor=red!60!black, urlcolor=magenta!70!black}

\title{\ours: Scaling 3D Human-Scene Interaction Data with Image Foundation Models} 

\author{Nikita Kister$^{1}$\thanks{Equal contribution. Correspondence: \texttt{nikita.kister@uni-tuebingen.de}} \quad Pradyumna YM$^{1}$\footnotemark[1] \quad Istv\'an S\'ar\'andi$^{1}$ \\
\textbf{Jiayi Wang}$^{2}$ \quad \textbf{Anna Khoreva}$^{3}$ \quad \textbf{Gerard Pons-Moll}$^{1,4}$ \\
$^1$University of T{\"u}bingen, Germany \quad
$^2$Bosch Center for Artificial Intelligence, Germany \\
$^3$Zalando SE, Germany \quad
$^4$Max Planck Institute for Informatics, Germany
}

\iclrfinalcopy %
\begin{document}

\maketitle

\begin{teaserfigure}
  \vspace{-2em}
  \includegraphics[width=1.0\textwidth, trim=0 60 0 100, clip]{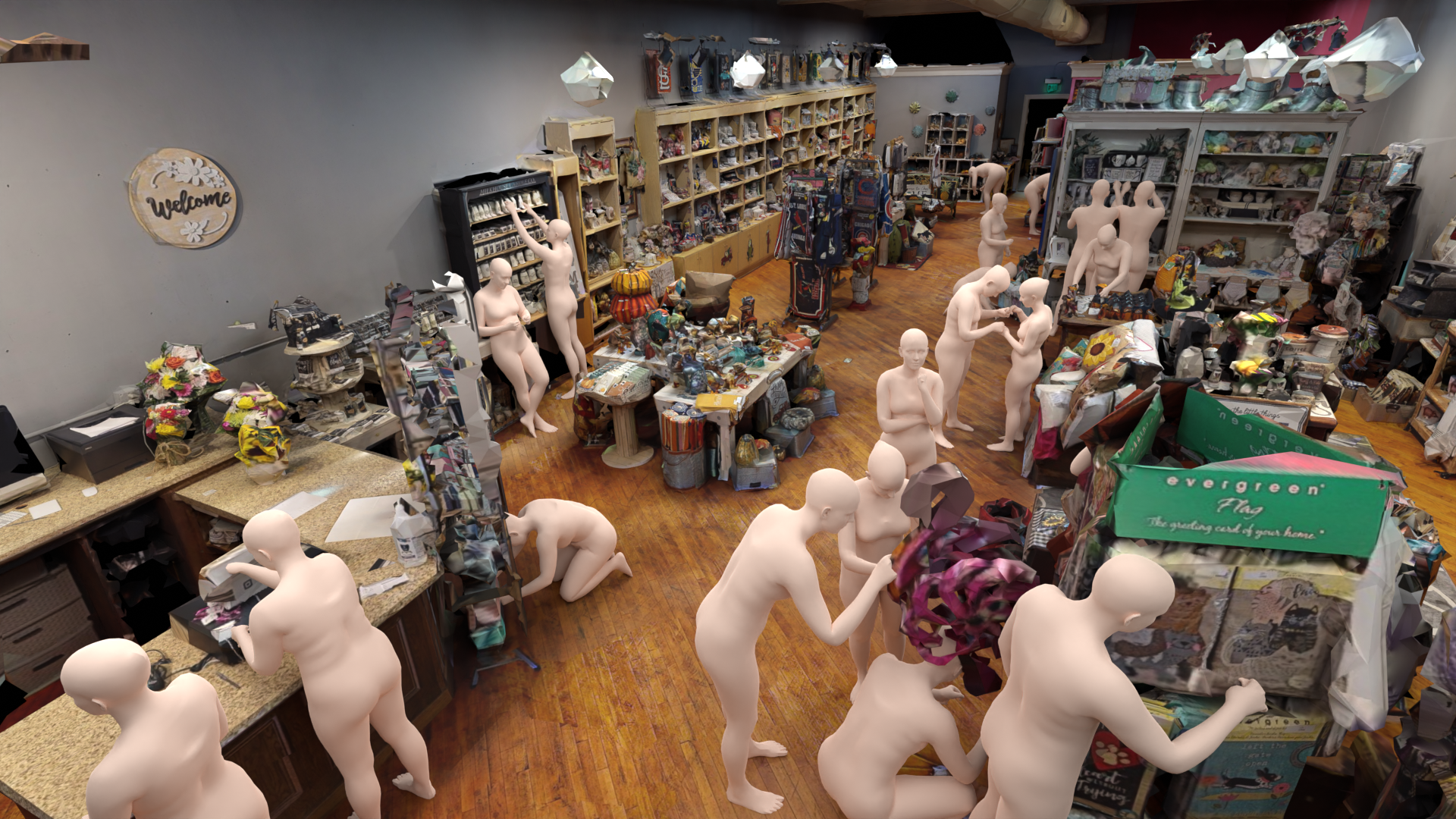}
  \caption{\textbf{\ours{}} generates diverse, scene-aware 3D human placement across varied environments and actions. In this craft shop, it produces realistic behaviors such as browsing, leaning, crouching, and reaching. Such interactions are specific to this environment and are hard to reenact in studio capture.
  Videos and interactive 3D visualizations of our data are available on the project page.}
  \label{fig:teaser}
\end{teaserfigure}

\begin{abstract}
Embodied AI and human-centric 3D scene understanding require data of people meaningfully interacting with diverse environments, yet such data remains scarce. Real-world capture has steadily pushed toward more scenes, but is bound to studios and at most around a hundred scenes, while synthetic datasets scale easily but place people without regard to what the scene affords.
We argue that useful human--scene interaction data requires diversity in both scenes and interactions, and we show this with our proposed data engine~\ours{}.
\ours{} is a fully automatic engine that renders an existing 3D scene, uses 2D foundation models to propose and depict context-appropriate interactions, and lifts them into SMPL-X bodies, where the known scene geometry makes lifting well-constrained.
Applied to Habitat-Matterport3D, \ours{} produces \ourdata{}, which pairs photorealistic images with SMPL-X bodies and complete scene geometry across $\sim$6.6k rooms, about 60$\times$ more scenes than the largest captured dataset.
We illustrate the quality and value of such large-scale human--scene interaction data through a variety of experiments. 
In a perceptual study, our interactions are preferred over two prior placement methods in 78\% of cases.
Across three tasks---1) contact estimation, 2) human--scene reconstruction, and 3) generating humans in scenes---training with \ourdata{} outperforms training on synthetic data with exact ground truth (BEDLAM), motion-capture data (TRUMANS), and real RGB-D captures (PROX); \eg, GRAFT trained on \ourdata{} reaches a contact F1 of 0.594 on PROX, versus 0.492 when trained on PROX itself.
Project page: \url{https://virtualhumans.mpi-inf.mpg.de/inhabit/}
\end{abstract}

\section{Introduction}
\label{sec:intro}
Understanding 3D scenes from a human-centric perspective is essential for robotics and embodied AI~\citep{puig2023habitat3,wang2024embodiedscan,zhen2024threedvla}.
Large collections of 3D scenes~\citep{dai2017scannet,hm3d} and of captured human motion~\citep{mahmood2019amass} exist separately, but datasets of people interacting with 3D environments remain small: capture-based efforts~\citep{hassan2019resolving,huang2022capturing,guzov2021human,araujo2023circle} cover fewer than 30 scenes each.
Synthetic datasets avoid the cost of capture by placing pre-recorded motions into scenes~\citep{black2023bedlam,wang2022humanise}, but yield people who sit where there is no chair or hover above the furniture they should rest on, rather than engaging with it.

More limiting than their size is what these datasets contain.
Capture setups are typically built from generic furniture, so the recorded interactions are those that any room allows: sitting on a horizontal surface, lying on a large one, reaching toward a point.
The interactions that make a space what it is are instead defined by what its objects are \emph{for}: playing pool, riding a gym bike, measuring a distance on a wall map, standing on a chair to change a light bulb, or carrying a table together with someone.
Recording such behavior would require capture sessions in pool halls, gyms, workshops, and every other type of environment, so it is essentially absent from existing data.
Existing data therefore spans only a narrow band of the interaction space, and models trained on it inherit that narrowness, generalizing poorly to how people actually occupy a scene.
Covering it requires a way to produce human--scene interaction (HSI) data in arbitrary scenes, without capture.

To produce such data, we build on the observation that 2D foundation models have seen millions of images of exactly these activities, and that this knowledge can be transferred into 3D by rendering a scene we already have in 3D.
We introduce \ours{}, a fully automatic engine that follows a \emph{render--generate--lift} principle: we render a view of an existing 3D scene, a vision--language model~\citep{gemini2025} proposes interactions the visible environment affords, an image-editing model~\citep{gemini2025} inserts one or more people performing them, and an optimization-based step~\citep{ym2025physic} recovers SMPL-X~\citep{pavlakos2019expressive} bodies in the scene.
Two obstacles stand in the way. Lifting a single edited view is ill-posed, which we address by optimizing against the complete scene geometry with an explicit human scale, rather than against partial depth.
And naively asking an editing model for a person yields generic standing poses, which we address by letting the VLM propose both scene-afforded and action-driven interactions, preserving the long tail.

Applied to Habitat-Matterport 3D~\citep{hm3d}, \ours{} produces \ourdata{}, spanning $\sim$6.6k rooms, about 60$\times$ more scenes than the largest captured dataset. Every sample pairs a photorealistic image with SMPL-X bodies and the complete scene geometry, and includes single- and multi-person interactions, all without manual annotation.
 In a perceptual study, our placements are preferred over prior work in $78\%$ of cases.

Data generated this way is not only plausible, it is better training data than what capture provides.
On established benchmarks for contact estimation, human--scene reconstruction and generating humans in scenes, training with \ourdata{} outperforms training on motion-capture data (TRUMANS~\citep{jiang2024scaling}), real RGB-D captures (PROX~\citep{hassan2019resolving}), and synthetic data with exact ground truth (BEDLAM~\citep{black2023bedlam}), improving SOTA methods for contact estimation~\citep{tripathi2023deco} and HSI reconstruction~\citep{graft,chen2025human3r}.
In an ablation over size-matched subsets (\Cref{sec:what-makes-useful}), the breadth of interactions contributes most to these gains.
Notably, GRAFT~\citep{graft} trained on \ourdata{} reaches a contact F1 of $0.594$ on PROX, against $0.492$ when trained on PROX itself.

In summary, our contributions are:
\begin{enumerate}
\setlength{\itemsep}{0pt}\setlength{\parskip}{0pt}
\item \ours{}, a fully automatic engine that populates any indoor 3D scene with diverse, scene-appropriate human interactions, requiring no manual annotation or human in the loop.
\item \ourdata{}, an HSI dataset spanning $\sim$6.6k rooms with complete scene geometry, SMPL-X bodies and photorealistic images, including multi-person interactions.
\item Evidence that fully automatically generated data competes with and surpasses motion-capture, real-world and exactly-labeled synthetic training data across three tasks.
\end{enumerate}
We will release our dataset and code publicly.

\section{Related Work}
\paragraph{Human--Scene Interaction Datasets}
Pioneering work on datasets, such as PiGraphs~\citep{savva2016pigraphs} and PROX~\citep{hassan2019resolving}, obtains human--scene data from real RGB-D recordings in only a handful of scenes.
Later works~\citep{huang2022capturing,guzov2021human,guzov2024interaction,zhang2022couch} use MoCap within pre-scanned scenes, but their dependency on complex MoCap and scanning infrastructure still limits their scalability, reaching at most 30 scenes.
TRUMANS~\citep{jiang2024scaling} covers 100 scenes by procedurally rearranging objects, but remains MoCap-dependent for human poses and is semantically constrained.
Synthetic datasets~\citep{black2023bedlam,wang2022humanise,cao2020longterm} are scalable but place prerecorded human motions into scenes, producing humans that are merely physically valid in the scene, but are not truly engaging with it.
CIRCLE~\citep{araujo2023circle} combines VR and MoCap to capture richer behaviors, but contains only 9 scenes.
Across these datasets the same tradeoff recurs: data is either grounded in real interactions but confined to a few scenes, or scalable but populated with people who ignore their surroundings. Using \ours{}, we introduce \ourdata{}, spanning $\sim$6.6k rooms, where every interaction is proposed for the specific environment it occurs in and lifted against the complete scene geometry. An ablation in \Cref{sec:what-makes-useful} indicates that this breadth contributes most to its value as training data.

\paragraph{Human Placement in 3D Scenes}
The dominant approach to placing humans in 3D scenes relies on geometric features and optimization for physical constraints such as contact and penetration~\citep{place}.
POSA~\citep{posa} learns a data-driven contact prior from captured interactions, but requires a pre-specified input pose, and only processes scene semantics through coarse-grained labels.
GenZI~\citep{genzi} generates 3D human-scene interactions from text but requires the user to specify the location of the interaction and relies on a brittle multi-view consensus optimization.
While these methods produce physically plausible configurations, they neither reason about what a scene affords nor run without per-interaction human input, so none of them scales to generating a dataset.
\ours{} addresses both by letting 2D foundation models decide what people do where, removing the human from the loop entirely and populating $\sim$6.6k rooms automatically.

\paragraph{Leveraging 2D Image Generation for 3D Tasks}
A recent line of work uses 2D image or video generation models to solve 3D tasks.
HumanWild~\citep{humanwild2024} generates humans in scenes using diffusion models and recovers 3D annotations from the generated images, showing that 2D generation can substitute for expensive real-world capture.
ComA~\citep{coma2024} uses 2D inpainting to synthesize human--object interaction images, then lifts these to 3D affordance maps, showing that interaction knowledge latent in 2D models can be grounded in 3D.
DAViD~\citep{david2025} generates 2D human--object interaction videos using video diffusion and lifts them to 4D interaction samples.
While these works are promising and show that 2D generative models can be a source of 3D human data, they operate only on isolated human--object pairs or single tasks without reasoning about the broader scene context, and none can generate HSI data at full building scale.
We take this principle further, to entire indoor environments, producing scene-appropriate interaction data at unprecedented scale fully automatically.

\begin{figure*}[t!]
\vspace{-1em}
    \centering
    \includegraphics[width=0.85\linewidth]{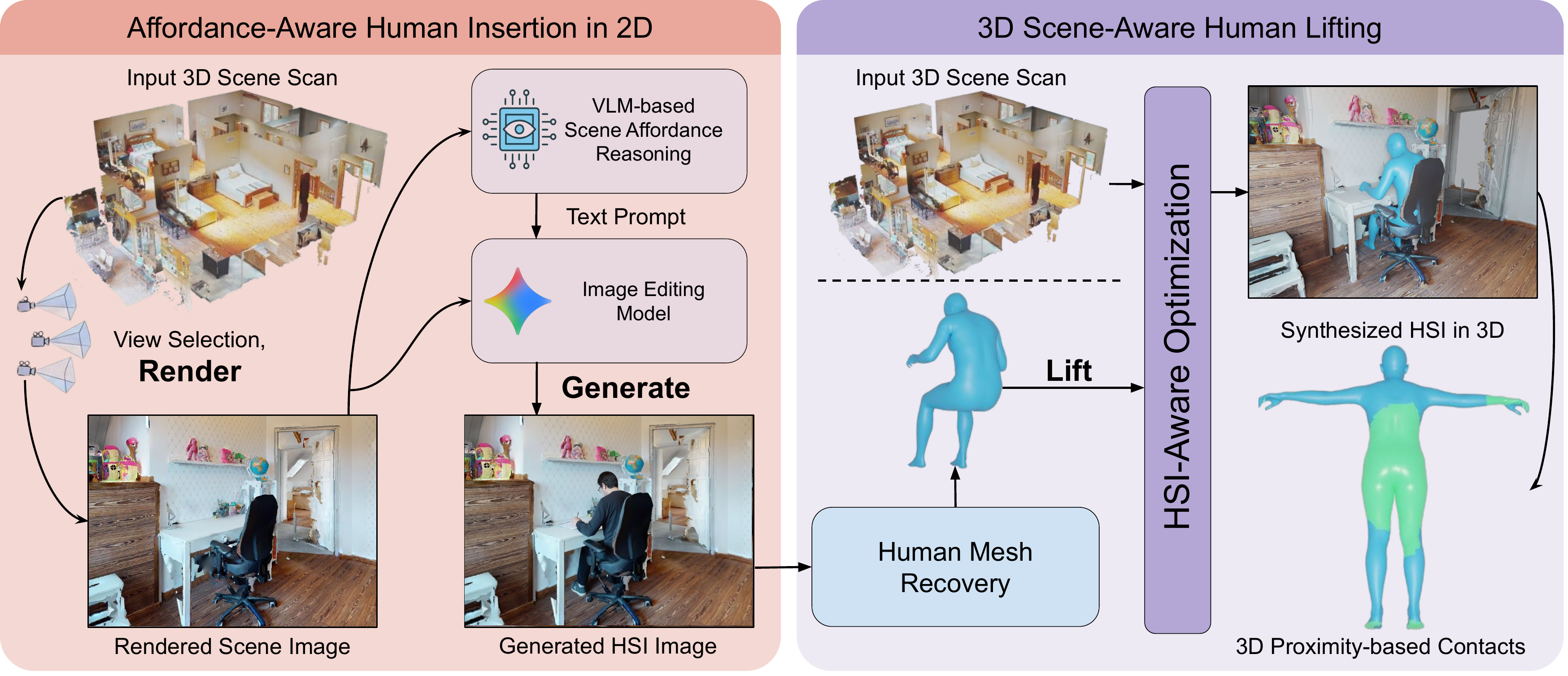}
    \caption{\textbf{Scalable 2D-to-3D human--scene interaction generation.} Overview of \ours{}. \emph{Left}: affordance-aware 2D insertion---from rendered scene views, a VLM proposes context-consistent interactions and an image-editing model synthesizes humans in the image. \emph{Right}: 3D scene-aware lifting---the inserted humans are recovered as SMPL-X meshes and optimized with scene geometry and proximity-based contact constraints for physically plausible 3D HSI.}
    \label{fig:overview}
\end{figure*}
\section{Method}

Our goal is to generate large-scale 3D human--scene interaction (HSI) data with natural, diverse, and context-aware behaviors. 
We formulate data generation as a \emph{render–generate–lift loop} (\Cref{fig:overview}): we render an existing 3D scene, use image foundation models to generate realistic humans in image space, and lift the results back into physically grounded 3D bodies aligned with the known scene. Rendering is what connects image foundation models to 3D: since each image comes from a known scene and camera, the diverse, context-aware humans these models insert are registered to metric geometry, turning ill-posed monocular reconstruction into a well-constrained fit.

Given any static 3D indoor scene (e.g., large-scale scanned environments such as HM3D~\citep{hm3d}), \ours{} has four stages. 1) We automatically sample camera viewpoints, render RGB/depth views, and perform scene-aware human insertion in 2D image space (\Cref{sec:scene_insertion}), where foundation models generate interaction proposals and edited images. 2) We reconstruct and place inserted humans in 3D with geometric and physical constraints from the full scene mesh (\Cref{sec:lifting}). 3) We scale this process through automated multi-view generation and batching (\Cref{sec:scaling}). 4) Because each stage can introduce artifacts, we apply post-reconstruction quality control to retain only physically plausible samples (\Cref{sec:filtering}). The final output is a dataset of RGB-D images paired with complete 3D scene geometry and aligned SMPL-X human meshes, ready to be used as training data in downstream tasks.

\subsection{Generating Scene-aware Humans in 2D}
\label{sec:scene_insertion}

The first stage provides the behavior: it populates rendered scene views with humans performing diverse, contextually meaningful interactions. This requires understanding both \emph{what} interactions a scene affords and \emph{how} they would look, which we obtain from two foundation models, a vision--language model (VLM) and an image-editing model. A naive idea would be to instruct the editing model to ``Add a person to this image.''
However, this yields mostly generic standing poses, ignoring rich affordances visible in the scene---a kitchen with a stove affords cooking, a living room with a sofa facing a TV affords watching, yet none of this is captured by a generic instruction.

Instead, we first query a VLM to analyze the rendered view and propose plausible activities for the visible scene.
The VLM's world knowledge lets it reason about the functions of objects and typical human behaviors---e.g., that a kitchen with a microwave affords operating it, or that a bedroom affords lying down.
We call this \emph{affordance-driven prompting} as it produces semantically appropriate interactions for the scene without predefined text prompts or manual supervision.
However, even affordance-driven proposals tend to converge on the most common activities and miss long-tail, rare ones.
We therefore also use an \emph{action-driven} strategy: given a set of generic action categories (\eg, reaching, leaning, pushing), we ask the VLM to tailor each to the current scene.
For example, ``reaching'' in a room with a ceiling light may become ``standing on a chair to replace the light bulb.''
These categories apply to any indoor scene without forcing implausible interactions, and yield more diverse interactions.
We also prompt for multi-person interactions, \eg, a couple dancing or moving furniture (\Cref{fig:multi:person}).

Given the VLM's textual interaction proposals, the image-editing model serves as the visual realization engine, generating body poses, proportions, and spatial placements adapted to the local scene context.
As seen from \Cref{fig:dataset-samples}, a person washing their hands is placed in front of the sink with hands below the tap.
These scene-adapted pose variations and multi-person interactions would be impossible to generate with pure optimization-based methods~\citep{genzi, posa, place} that rely on geometric features without visual scene understanding (\Cref{fig:qualitative-results}).
Prompt and design details are provided in \Cref{suppl:sec:prompting}.

\subsection{Lifting Inserted Humans to 3D}
\label{sec:lifting}
The second stage provides the geometry, building on physically grounded human--scene reconstruction such as PhySIC~\citep{ym2025physic} and PROX~\citep{hassan2019resolving}.
PhySIC estimates both 3D humans and scenes from RGB images by optimizing the human pose to satisfy geometric constraints with the predicted depth map, whereas PROX fits humans to RGB-D frames using physical constraints.
Since we take a complete known 3D scene geometry as input, our lifting task is better constrained, allowing more precise human placement and alignment with scene geometry.
In \Cref{suppl:sec:full,suppl:sec:scale} we ablate both choices; without the complete scene geometry, for example, the contact accuracy of the lifted bodies drops substantially (\Cref{tab:abl:rec}).

We optimize the human parameters of pose $\thetas$, shape $\betas$, camera-relative translation $\transl$, and scale $s$ with the following objective inspired by PhySIC:
\begin{equation}
\begin{aligned}
\operatorname{minimize}_{\boldsymbol{\theta},\, \boldsymbol{\beta},\, \transl,\, s} \quad
& \lambda_{\text{proj}} \mathcal{L}_{\text{proj}} +
  \lambda_{\text{depth}} \mathcal{L}_{\text{depth}} +
  \lambda_{\text{con}} \mathcal{L}_{\text{contact}} \\
& + \lambda_{\text{pen}} \mathcal{L}_{\text{penetration}} +
  \lambda_{\text{reg}} \mathcal{L}_{\text{reg}},
\end{aligned}
\end{equation}
where $\mathcal{L}_{\text{proj}}$ enforces 2D reprojection consistency, $\mathcal{L}_{\text{depth}}$ aligns the visible human surface with the estimated depth, $\mathcal{L}_{\text{contact}}$ encourages plausible contact with scene surfaces, and $\mathcal{L}_{\text{penetration}}$ penalizes intersections with the scene geometry.
Full loss definitions and weights are provided in \Cref{suppl:sec:recon}.

\subsection{Generating the InHabitants Dataset}
\label{sec:scaling}
\begin{figure*}[t]
\vspace{-1em}
  \centering
  \begin{minipage}[c]{0.63497\textwidth}
    \centering
    \includegraphics[width=\linewidth, clip]{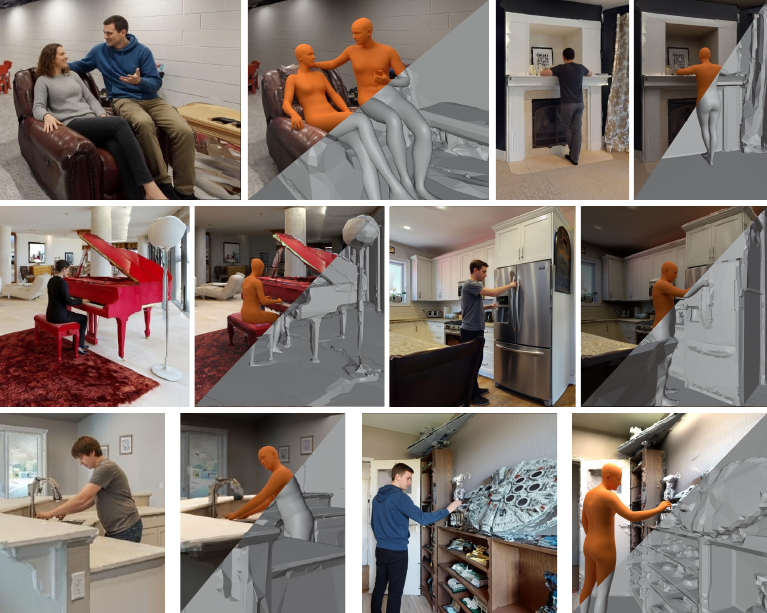}
  \end{minipage}\hspace{2.4pt}%
  \begin{minipage}[c]{0.35899\textwidth}
    \centering
    \setlength{\parskip}{0pt}\setlength{\lineskip}{0pt}\offinterlineskip
    \includegraphics[width=\linewidth,trim={8 8 8 36.6},clip]{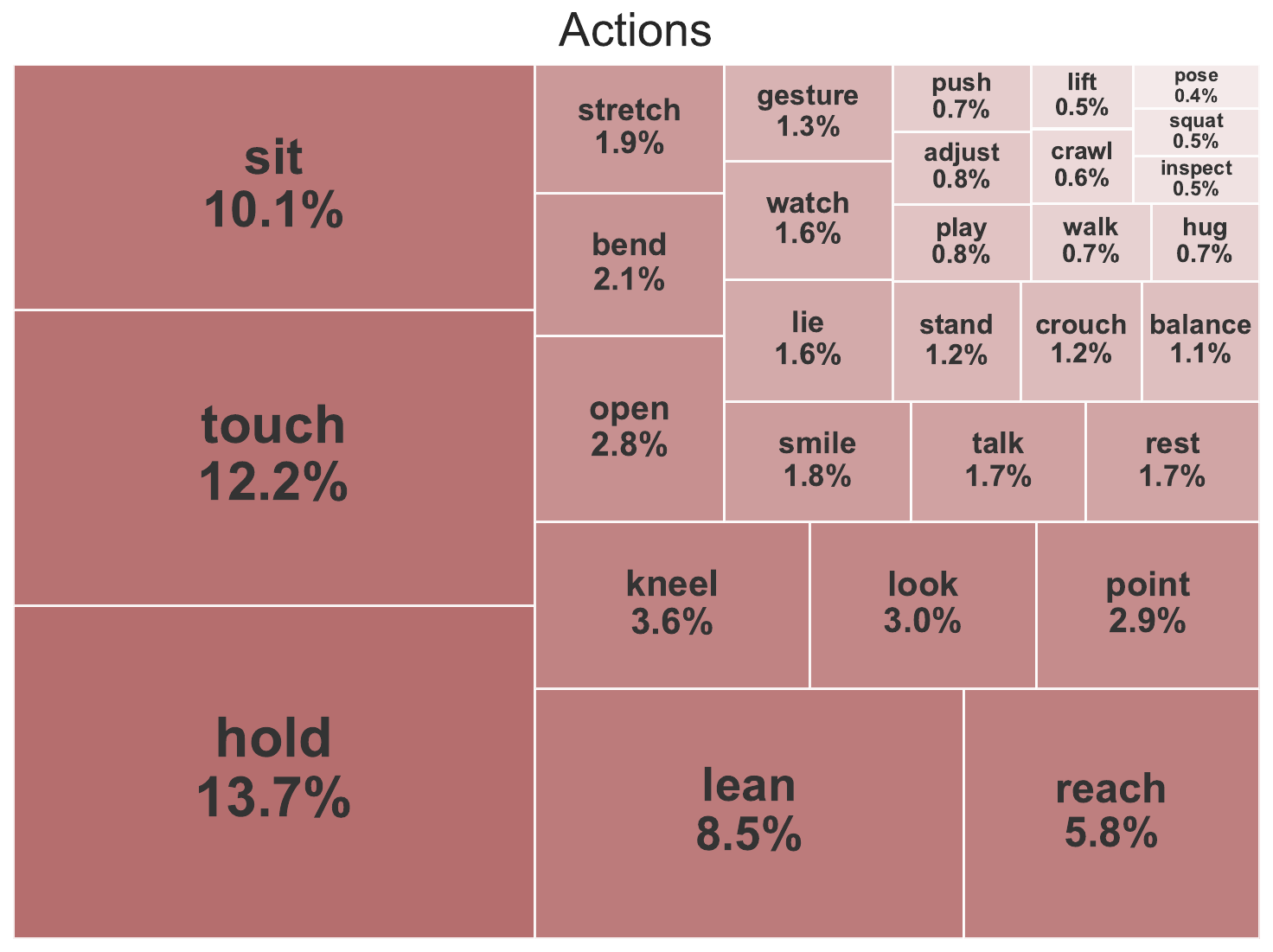}%
    \par\vspace{2pt}%
    \includegraphics[width=\linewidth,trim={8 8 8 36.9},clip]{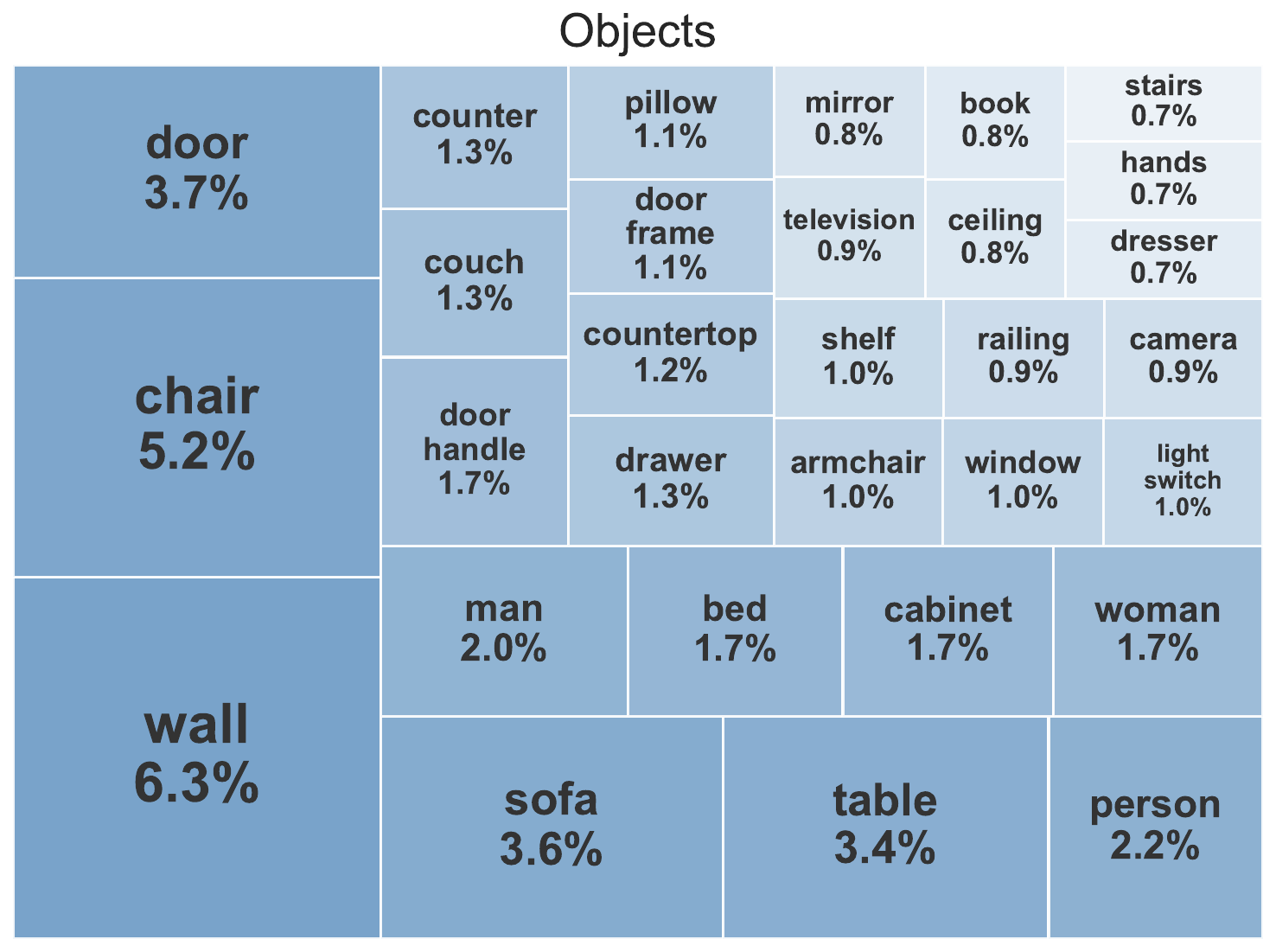}%
  \end{minipage}
  \caption{\emph{Left}: representative samples from \ourdata{}. Each pair shows the edited RGB image (left) and the corresponding lifted 3D human in the 3D scene (right). Our dataset contains diverse, context-aware interactions across varied indoor environments. \emph{Right}: distribution of \ourdata{} by action type (\textit{top}) and by interacted object category (\textit{bottom}), showing broad coverage of both diverse activities and scene elements.}
  \label{fig:dataset-samples}
  \label{fig:dataset}
\end{figure*}

\ours{} is agnostic to the source of 3D scenes and can operate on any textured scene mesh.
In this work, we instantiate it on the Habitat-Matterport 3D dataset (HM3D)~\citep{hm3d}, containing $\sim$800 building-scale scans spanning residential, commercial, and institutional interiors with multiple floors and rooms, containing kitchens, bedrooms, offices, hallways, stairwells, etc.
This enables a diverse distribution of human--scene interactions in the resulting dataset (\Cref{fig:dataset-samples}). As detailed in \Cref{suppl:sec:runtime}, the full dataset is generated in about 19 hours on 32 consumer GPUs.

For scalability, we need automatic viewpoint selection, but random camera poses would often look directly at a nearby wall, be obstructed by furniture or show only ceilings or narrow spaces where inserting a human is infeasible.
We therefore filter viewpoints using a depth heuristic: we require 1)~floor visibility in the bottom of the image, so that humans can be added without truncation, and 2)~sufficient free space ahead, allowing an unobstructed view.
Both criteria are evaluated by threshold checks on the depth map, with details in \Cref{suppl:sec:views}.

Since some rooms can be largely empty and offer limited affordances (\eg, bare corridors, unfurnished storage rooms), the VLM's affordance-driven prompting (\Cref{sec:scene_insertion}) would only yield a narrow set of generic activities.
To achieve pose diversity even then, we augment our prompting with fine-grained text descriptions of body poses from the PoseScript dataset~\citep{posescript}.

\paragraph{Post-reconstruction Filtering}
\label{sec:filtering}
Image editing models are not perfect and the output is not always physically consistent with the underlying 3D scene geometry.
To remove implausible or incomplete human insertions and ensure high-quality data, we apply three complementary, automated post-reconstruction filters. %
We discard 1) examples with strongly truncated humans, 2) generations where the edited image contains unintended scene changes besides inserting the human, and 3) human mesh reconstructions with implausibly small volumes. Details can be found in \Cref{suppl:sec:qc}; the resulting poses are also physically more stable than those of captured datasets (\Cref{suppl:sec:stability}).

\paragraph{Dataset Diversity}
To characterize the dataset, we use a VLM~\citep{gemini3} to classify each generated sample by its action and the primary object involved, in an open-ended manner. \Cref{fig:dataset-samples} shows the distribution over the most frequent actions and objects, illustrating the breadth of interactions in \ourdata{}.
We further quantify our diversity in \Cref{tab:diversity}, showing higher interaction diversity, target-object diversity, and Average Pose Distance (APD) than TRUMANS and PROX-D.

\begin{table}[t]
  \centering
      \caption{\textbf{Dataset diversity.} We compare the diversity of interactions and objects with other datasets.
      We extract actions and target objects with a VLM and compute average cosine distance to measure their diversity. Despite containing the fewest images, \ourdata{} has the most diverse interactions, target objects and poses, and covers about 60$\times$ more scenes than TRUMANS.}
  \setlength{\tabcolsep}{6pt}
  \renewcommand{\arraystretch}{0.95}
  \begin{tabular*}{\textwidth}{@{\extracolsep{\fill}}lccccc@{}}
    \toprule
    Dataset & Images & Scenes & Action Diversity\ $\uparrow$ & Target Obj.\ Div.\ $\uparrow$ & APD $\uparrow$ \\
    \midrule
    TRUMANS              & 1.6M & 100  & 0.406          & 0.452          & 25.71 \\
    PROX-D               & 88k  & 12   & 0.410          & 0.445          & 32.81 \\
    \ourdata{} (Ours)    & 78k  & \textbf{6.6k} & \textbf{0.485} & \textbf{0.493} & \textbf{35.42} \\
    \bottomrule
  \end{tabular*}

  \label{tab:diversity}
\end{table}

\section{Experiments}
In this section we show how \ourdata{} contributes to advances of state-of-the-art methods on established benchmarks for three different tasks:
1) image-based human--scene reconstruction, 2) generating 3D humans in scenes, and 3) body contact surface estimation.
The scale and diversity of \ourdata{} are well suited to these tasks, as they would require expensive motion capture or tedious manual annotations otherwise. 
Furthermore, we demonstrate that our generation framework creates poses that are both more natural and diverse than other 3D human placement methods.

\subsection{Human--Scene Reconstruction}
\begin{table}[t]
\centering
\caption{\textbf{\ourdata{} improves human--scene reconstruction methods substantially}. Best result per model in bold, second best underlined.}
\setlength{\tabcolsep}{5pt}
\renewcommand{\arraystretch}{0.95}
\begin{tabular*}{\textwidth}{@{\extracolsep{\fill}}c@{\hspace{4pt}}l ccc @{\hskip 0.3cm} ccc@{}}
\toprule
\multirow{2}{*}{Model} & \multirow{2}{*}{Training Data} & \multicolumn{3}{c}{PROX Quantitative} & \multicolumn{3}{c}{RICH-100} \\
\cline{3-5} \cline{6-8}
& & \scriptsize{F1 score $\uparrow$} & \scriptsize{V2S $\downarrow$} & \scriptsize{PA-MPJPE $\downarrow$} & \scriptsize{F1 score $\uparrow$ }& \scriptsize{V2S $\downarrow$ }& \scriptsize{PA-MPJPE $\downarrow$ }\\
\midrule
\multirow{4}{*}{Human3R} & BEDLAM & 0.298 & 199.0 & \underline{57.8} & \underline{0.402} & 272.6 & \underline{47.0} \\
 & PROX & 0.215 & 250.1 & 66.6 & 0.279 & 302.1 & 73.5 \\
 & \ourdata{} (Ours) & \underline{0.373} & \textbf{177.0} & 59.5 & \underline{0.402} & \underline{257.9} & 48.1 \\
 & \makecell[r]{+ BEDLAM} & \textbf{0.415} & \underline{179.4} & \textbf{57.2} & \textbf{0.459} & \textbf{249.2} & \textbf{46.0} \\
\midrule
\multirow{4}{*}{GRAFT} & BEDLAM & 0.396 & \underline{183.79} & 84.0 & 0.469 & 230.14 & 65.2 \\
 & PROX & 0.492 & \textbf{166.51} & 72.5 & 0.378 & 270.14 & 84.4 \\
 & \ourdata{} (Ours) & \textbf{0.594} & 184.36 & \textbf{49.7} & \textbf{0.578} & \underline{224.66} & \underline{46.3} \\
 & \makecell[r]{ + BEDLAM} & \underline{0.564} & 185.75 & \underline{51.0} & \underline{0.561} & \textbf{221.52} & \textbf{45.0} \\
\bottomrule
\end{tabular*}%

\label{table:hsr}
\end{table}

\begin{figure*}[t]
\vspace{-1em}
  \centering
  {\offinterlineskip%
  \def\wqcell#1{\includegraphics[width=0.21\linewidth]{assets/wild_qual/#1}}%
  \def\wqhd#1{\makebox[0.21\linewidth][c]{\footnotesize #1}}%
  \def\wqgap{\hspace{0.015\linewidth}}%
  \mbox{\wqhd{Input}\wqgap\wqhd{Ours}\wqgap\wqhd{BEDLAM}\wqgap\wqhd{PROX}}\par
  \vspace{0.003636\linewidth}
\mbox{\wqcell{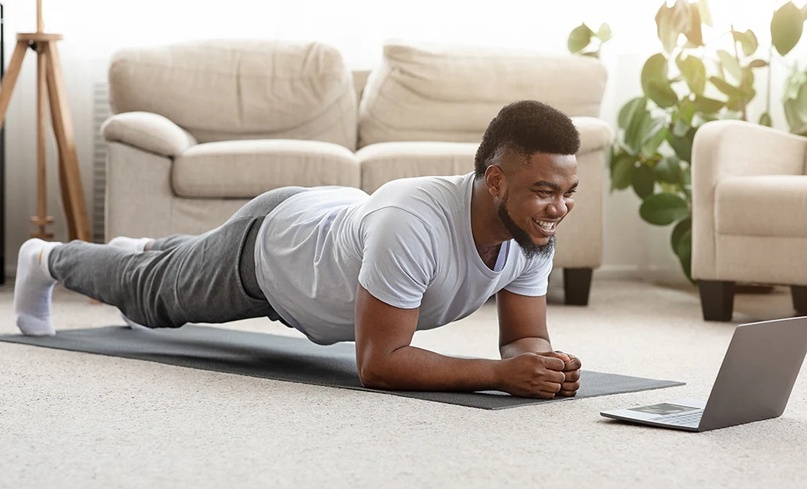}\wqgap\wqcell{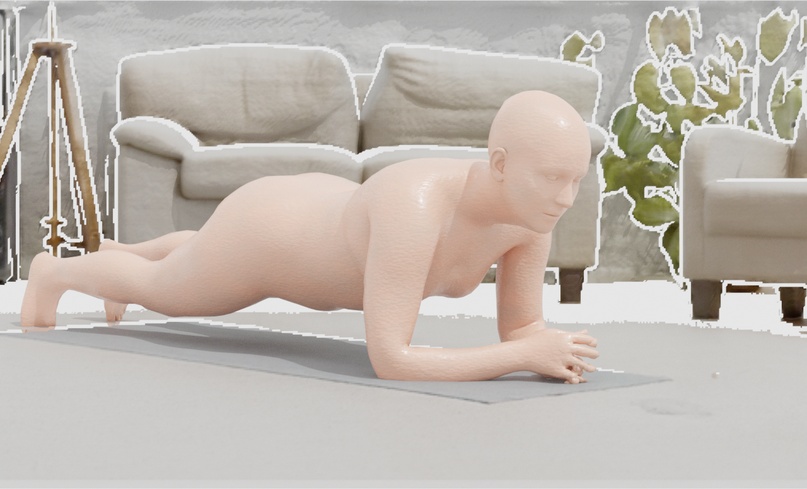}\wqgap\wqcell{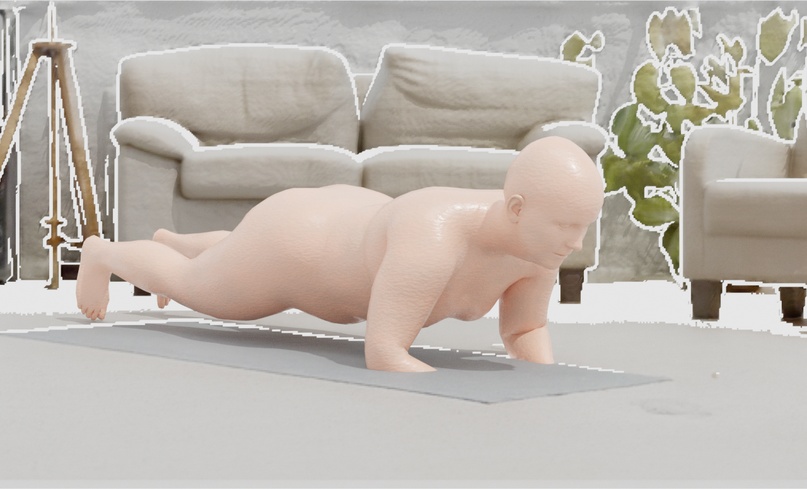}\wqgap\wqcell{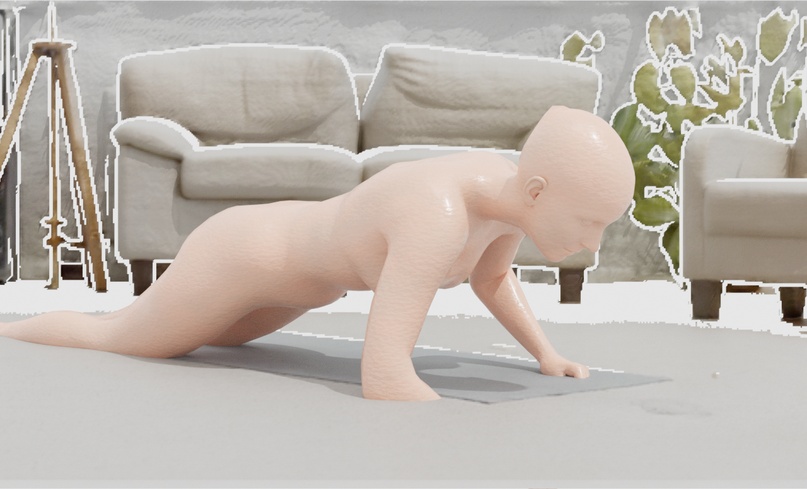}}\par  %
\vspace{0.007273\linewidth}
\mbox{\wqcell{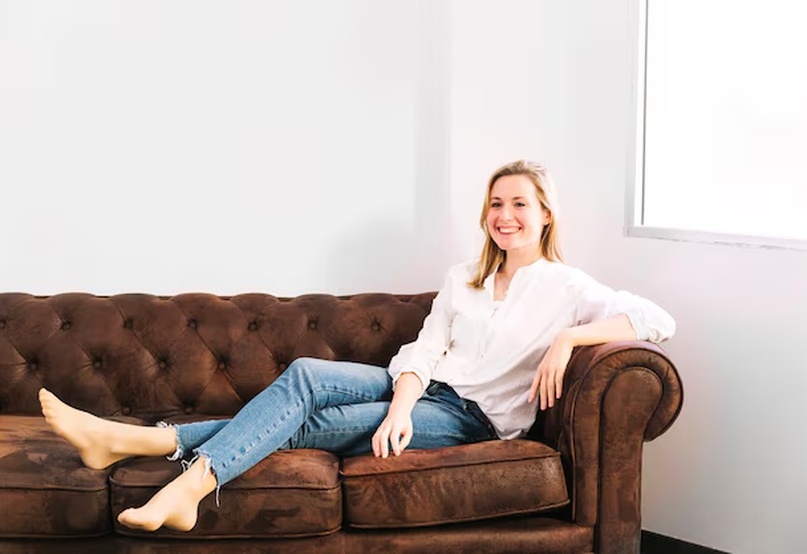}\wqgap\wqcell{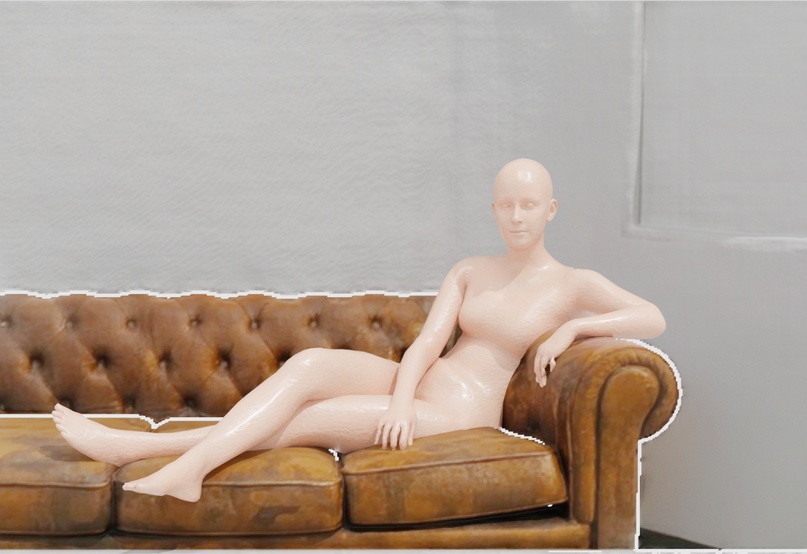}\wqgap\wqcell{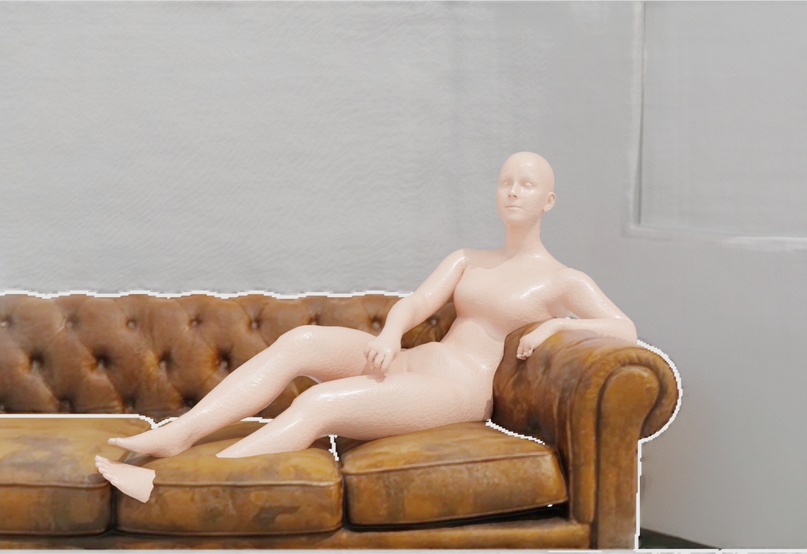}\wqgap\wqcell{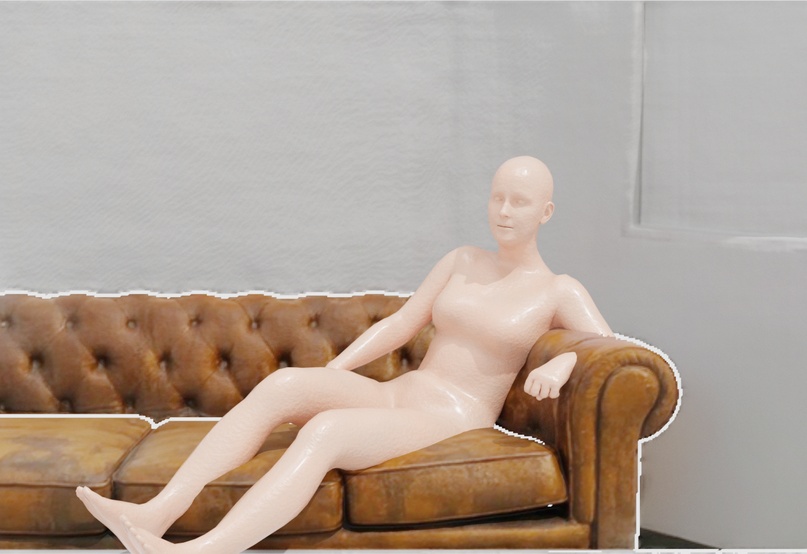}}\par  %
  }%
  \caption{\textbf{GRAFT trained on \ourdata{} reconstructs bodies with more realistic scene interactions.} We apply GRAFT trained on \ourdata{}, BEDLAM and PROX to unconstrained internet images. Training on \ourdata{} yields bodies that rest on the surfaces that support them, while others more often intersect or float relative to them, such as the legs sinking into the sofa for BEDLAM.}
  \label{fig:suppl:wildqual}
\end{figure*}

Human--scene reconstruction (HSR) jointly recovers, from a single image, the 3D body and the surrounding scene geometry in a spatially consistent way (e.g., feet on the floor, hand on the table). Diverse real-world training data with both signals paired is scarce, exactly the gap \ourdata{} fills.

We train two recent image-based HSR methods, GRAFT~\citep{graft} and Human3R~\citep{chen2025human3r} on \ourdata{} and evaluate on PROX Quant.~\citep{hassan2019resolving} and RICH-100~\citep{ym2025physic, huang2022capturing}.
We compare \ourdata{} against two commonly used datasets: BEDLAM~\citep{black2023bedlam} and PROX~\citep{hassan2019resolving}.
BEDLAM is a synthetic dataset that provides images, 3D body annotations and depth maps. However, it places human motion randomly into the scene, resulting in mostly walking poses and interactions that are not supported by the scene, such as sitting in the air.
PROX, on the other hand, records people interacting in various ways (e.g., sitting or lying), but captures only 12 small scenes.
Our HSR metrics are F1-score for scene contacts and V2S~\citep{graft} to evaluate the alignment of the body with the scene. We also report PA-MPJPE, measuring pose accuracy.

Training on \ourdata{} improves HSR metrics regardless of the underlying method used and outperforms using existing datasets overall when compared on both generalization benchmarks (see \Cref{table:hsr}).
This is due to the larger number of scenes and the diverse interactions our dataset provides, which PROX and BEDLAM lack. Human3R trained exclusively on our data is on par with BEDLAM on RICH, since RICH sequences are restricted to feet and hand contacts. The advantage is more apparent on PROX-Quant., which contains more diverse scene interactions. However, combining both BEDLAM and \ourdata{} results in the best performance for Human3R, highlighting that our interactions are complementary to BEDLAM's. GRAFT, a method that explicitly reasons about contacts in 3D space, trained on our data alone improves substantially in HSR and pose metrics compared to the other datasets. %
It also generalizes better to in-the-wild images as can be seen in \Cref{fig:suppl:wildqual}.
Notice how the bodies rest on the surface with minimal penetration. \ourdata{} is well suited for this approach and advances the state of HSR.

\subsection{Learning to Generate Humans in Scenes}
\label{sec:exp-generative}
\begin{table}[t]
  \centering
  \caption{\textbf{SceneDiffuser achieves improved results on ScanNet when trained on \ourdata{}}.}
  \setlength{\tabcolsep}{6pt}
  \renewcommand{\arraystretch}{0.95}
  \begin{tabular*}{\textwidth}{@{\extracolsep{\fill}}lccccc@{}}
    \toprule
    \makecell[l]{Training Data} & \makecell{Contact $\uparrow$} & \makecell{Penetration\\(in L) $\downarrow$} & \makecell{Foot-Contact $\uparrow$} & \makecell{Physical\\Stability $\uparrow$} & \makecell{Diversity $\uparrow$} \\
    \midrule
    PROX                & 0.570 & \textbf{5.2} & 0.04 & 0.14 & 2.76 \\
    TRUMANS             & 0.733 & 7.1 & 0.21 & 0.27 & 3.70 \\
    \ourdata{}  (Ours) & \textbf{0.795} & 6.8 & \textbf{0.29} & \textbf{0.32} & \textbf{4.02} \\
    \bottomrule
  \end{tabular*}

  \label{tab:scenediff:scannet}
\end{table}

\begin{figure*}[t]
\vspace{-1em}
  \centering
  \definecolor{sdOurs}{HTML}{B47B4B}
  \definecolor{sdTrumans}{HTML}{4C9466}
  \definecolor{sdProx}{HTML}{B5528A}
  \begin{tikzpicture}[
      img/.style={anchor=north west, inner sep=0pt, outer sep=0pt},
      frame/.style={draw=#1, line width=0.8pt},
      head/.style={anchor=south, inner sep=1.5pt, font=\small, text height=1.6ex, text depth=0.4ex},
    ]
    \pgfmathsetlengthmacro{\gap}{1.5pt}                 %
    \pgfmathsetlengthmacro{\gutter}{5pt}                %
    \pgfmathsetlengthmacro{\W}{(\textwidth-2*\gutter)/3} %
    \pgfmathsetlengthmacro{\H}{\W/530*364}
    \pgfmathsetlengthmacro{\Z}{(\W-2*\gap)/3}           %
    \pgfmathsetlengthmacro{\ZH}{\Z/400*364}             %
    \pgfmathsetlengthmacro{\zpx}{\Z/400}                %
    \pgfmathsetlengthmacro{\Zfull}{\Z/400*530}          %

    \def\sdmain#1#2#3#4{%
      \node[img] (main#1) at ({#1*(\W+\gutter)},0) {\includegraphics[width=\W]{assets/scene_diff/#2}};
      \draw[frame=#3] ([shift={(0.4pt,0.4pt)}]main#1.south west) rectangle ([shift={(-0.4pt,-0.4pt)}]main#1.north east);
      \node[head] at (main#1.north) {#4};
    }
    \def\sdzoom#1#2#3#4#5{%
      \coordinate (zc) at ({#1*(\W+\gutter)+#2*(\Z+\gap)}, {-\H-\gap});
      \begin{scope}
        \clip (zc) rectangle ++(\Z,-\ZH);
        \node[img] at ([xshift=-#4*\zpx]zc) {\includegraphics[width=\Zfull]{assets/scene_diff/#3}};
      \end{scope}
      \draw[frame=#5] ([shift={(0.4pt,-0.4pt)}]zc) rectangle ([shift={(\Z-0.4pt,-\ZH+0.4pt)}]zc);
    }

    \sdmain{0}{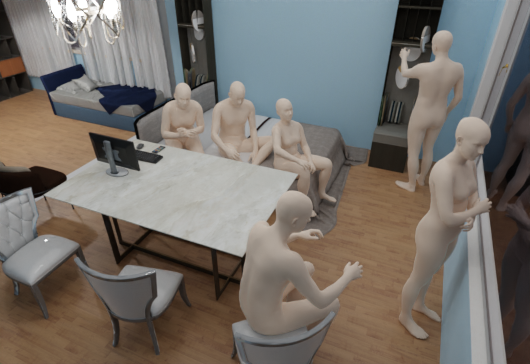}{sdOurs}{\ourdata{} (Ours)}
    \sdmain{1}{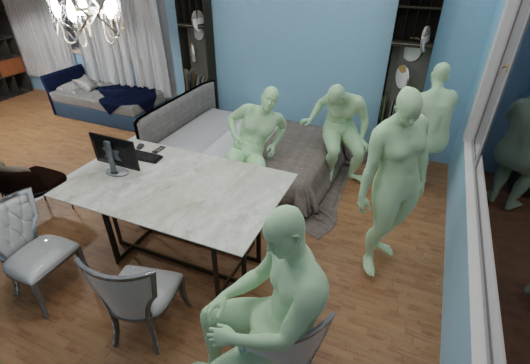}{sdTrumans}{TRUMANS}
    \sdmain{2}{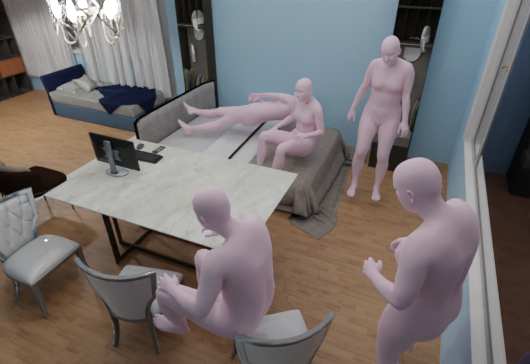}{sdProx}{PROX}

    \sdzoom{0}{0}{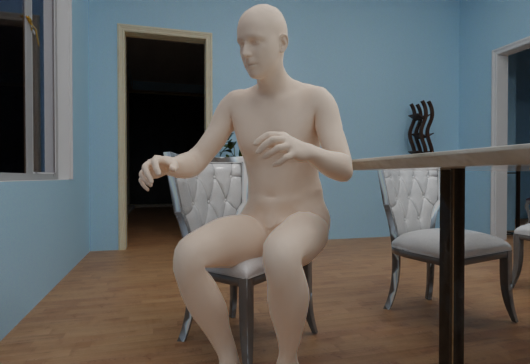}{85}{sdOurs}
    \sdzoom{0}{1}{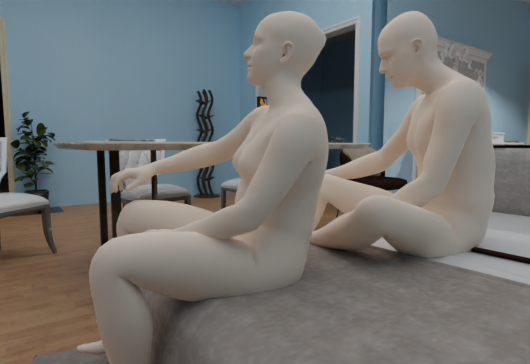}{95}{sdOurs}
    \sdzoom{0}{2}{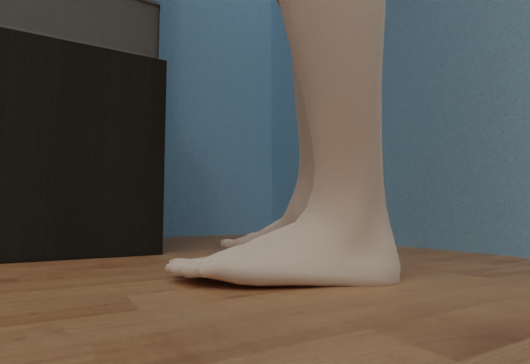}{70}{sdOurs}
    \sdzoom{1}{0}{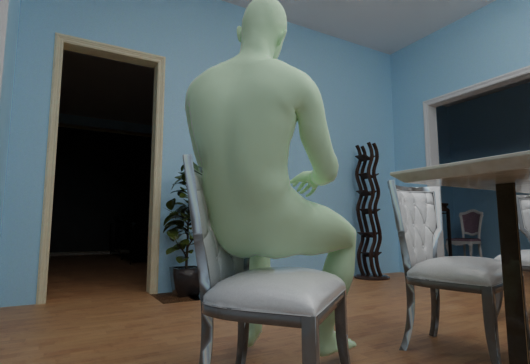}{85}{sdTrumans}
    \sdzoom{1}{1}{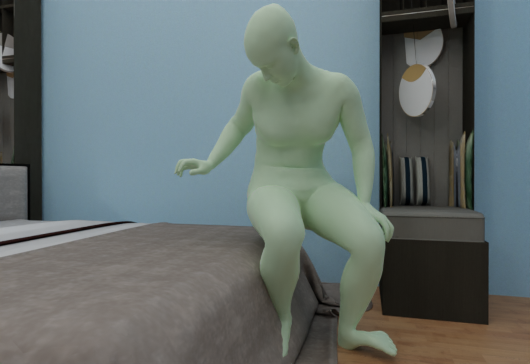}{75}{sdTrumans}
    \sdzoom{1}{2}{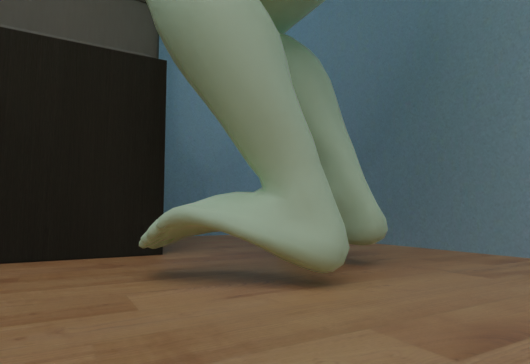}{70}{sdTrumans}
    \sdzoom{2}{0}{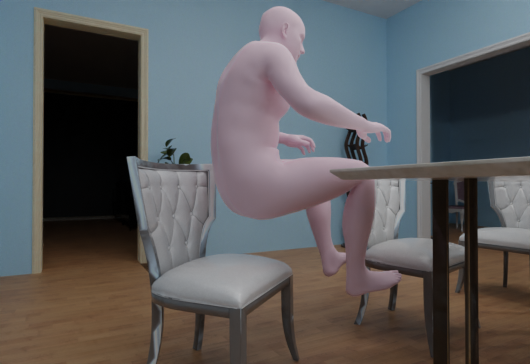}{85}{sdProx}
    \sdzoom{2}{1}{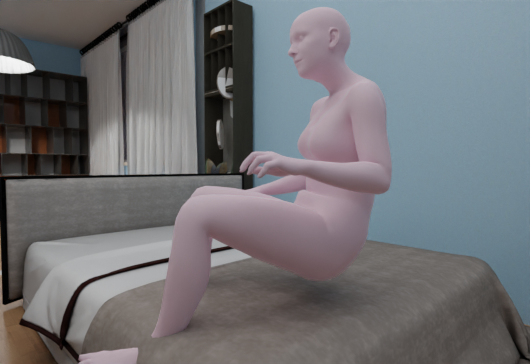}{100}{sdProx}
    \sdzoom{2}{2}{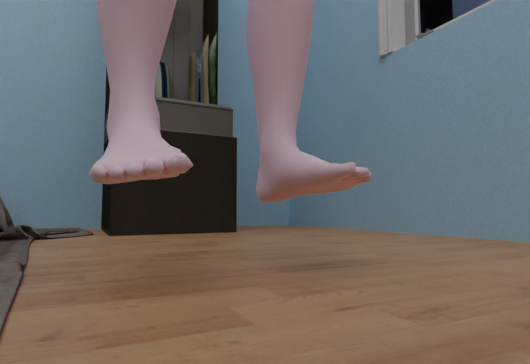}{95}{sdProx}  %
  \end{tikzpicture}
  \caption{\textbf{SceneDiffuser trained on \ourdata{} generates stable, diverse poses with realistic scene contact compared to baselines.} Models trained on TRUMANS and PROX produce more floating bodies and interactions that are physically unstable. Notice the floating feet for TRUMANS and PROX, whereas ours are in close contact to the floor, highlighting the quality of our data.}
  \label{fig:scene-diffuser-qualitative}
\end{figure*}

\ourdata{} is generated with a complex pipeline involving 2D generation and 3D reconstruction. However, once generated, this data can be used to train a much faster model that generates SMPL bodies given a 3D scene as input, distilling the entire \ours{} pipeline into a single diffusion model.
We adopt SceneDiffuser~\citep{scenediffuser} for this, a pose diffusion model conditioned on the input point cloud of the scene.
We train it on \ourdata{} and compare against models trained on PROX and TRUMANS~\citep{jiang2024scaling}.
We evaluate all models on 638 ScanNet~\citep{dai2017scannet} scenes, an out-of-distribution setting for all trained models.
Following prior work~\citep{scenediffuser}, we measure scene contact and pose diversity, and additionally report penetration volume (SMPL–scene intersection), foot contact, and the percentage of physically stable poses (\Cref{suppl:sec:stability}).

The model trained on \ourdata{} outperforms both baselines in terms of contact, foot contact, and physical stability, while still providing a higher level of diversity (see \Cref{tab:scenediff:scannet} and \Cref{fig:scene-diffuser-qualitative}).
The high diversity of scenes and interactions allows it to generalize better to out-of-distribution data.
The models trained on PROX and TRUMANS tend to generate more floating bodies as indicated by the lower contact and foot contact numbers. Especially the model trained on PROX struggles to create realistic contact, leading to its lower penetration volume. 
Conversely, our higher contact comes with some penetration, e.g., bodies sinking into the bed (\Cref{fig:scene-diffuser-qualitative}), which may partly stem from our reconstruction deliberately allowing penetration to account for soft surfaces such as sofas and beds.
Overall, our data provides high-quality, diverse interactions to help further research.

\subsection{Human--Scene Contact Estimation}
\begin{table}[t]
\centering
\begin{minipage}[t]{0.53\linewidth}
  \centering\small
  \caption{\textbf{\ourdata{} improves contact estimation} performance of DECO on DAMON-test.}
  \setlength{\tabcolsep}{2pt}
  \renewcommand{\arraystretch}{1.05}
  \begin{tabular*}{\linewidth}{@{\extracolsep{\fill}}lcccc@{}}
    \toprule
    Training Data & Prec. $\uparrow$ & Rec. $\uparrow$ & F1 $\uparrow$ & geo.\@ err $\downarrow$ \\
    \midrule
    DAMON & 0.639 & 0.523 & 0.512 & 25.06 \\
    DAMON+TRUMANS & 0.670 & 0.481 & 0.497 & 18.87 \\
    DAMON+\ourdata{}  & \textbf{0.670} & \textbf{0.532} & \textbf{0.534} & \textbf{17.60} \\
    \bottomrule
  \end{tabular*}
  
  \label{tab:damon-contact}
\end{minipage}\hfill
\begin{minipage}[t]{0.45\linewidth}
  \centering\small
  \caption{\textbf{Placement evaluation} regarding semantic prompt alignment and contacts.}
  \setlength{\tabcolsep}{2pt}
  \renewcommand{\arraystretch}{1.05}
  \begin{tabular}{lcccc}
    \toprule
    Method & CLIP $\uparrow$ & Prec. $\uparrow$ & Rec. $\uparrow$ & F1 $\uparrow$ \\
    \midrule
    GenZI & 0.252 & 0.500 & 0.052 & 0.060 \\
    \posa & 0.249 & 0.715 & 0.134 & 0.195 \\
    \ours{} (Ours) & \textbf{0.266} & \textbf{0.720} & \textbf{0.229} & \textbf{0.267} \\
    \bottomrule
  \end{tabular}
  
  \label{tab:ours:quant}
\end{minipage}
\end{table}

Body contact estimation predicts, from a single RGB image, which body-surface vertices touch the scene. It is a key signal for human--scene reasoning, but is bottlenecked by the cost of per-vertex annotation on real images. We show that \ourdata{} provides high-quality contact supervision that directly improves a SOTA estimator. Since we recover both the human and the surrounding geometry, we read off per-vertex contacts directly from spatial proximity, yielding image-to-contact pairs without manual labeling. We use these to train DECO~\citep{tripathi2023deco}, a SOTA model for in-the-wild body contact estimation, and evaluate on the DAMON benchmark.

We follow the standard DECO protocol for both training and evaluation, including its metrics. Following TRUMANS~\citep{jiang2024scaling}, the closest prior dataset providing human--scene contact supervision, we train on the DAMON+\allowbreak\ourdata{} combination; results are reported in \Cref{tab:damon-contact}, where the DAMON and DAMON+TRUMANS results are those reported by TRUMANS. 
DAMON+\allowbreak\ourdata{} outperforms DECO trained on DAMON alone, and also outperforms training on DAMON+TRUMANS. This is remarkable, as TRUMANS is a labor-intensive real-world motion capture dataset, whereas our dataset is generated fully automatically. One reason is that TRUMANS is limited to 100 scenes, offers narrower action coverage (see \Cref{tab:diversity}), and contains images that are less realistic than ours.
This shows that our images are aligned with the 3D scene well enough to provide a high-quality supervision signal for fine details such as contacts, which generalizes to the manually annotated labels of DAMON.

\subsection{Comparing \ours{} with Human Placement Baselines}

\begin{figure*}[t]
  \centering
  \setlength{\tabcolsep}{2pt}
  \renewcommand{\arraystretch}{0.5}
  \begin{tabular}{cccccc}

    \rotatebox{90}{\hspace{0.8cm}GenZI} &
    \includegraphics[width=0.16\textwidth]{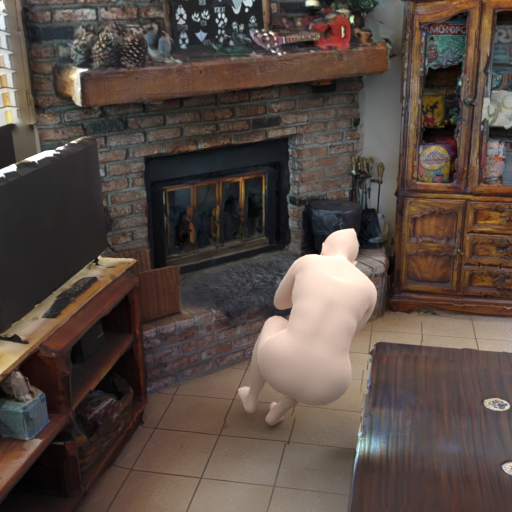} &
    \includegraphics[width=0.16\textwidth]{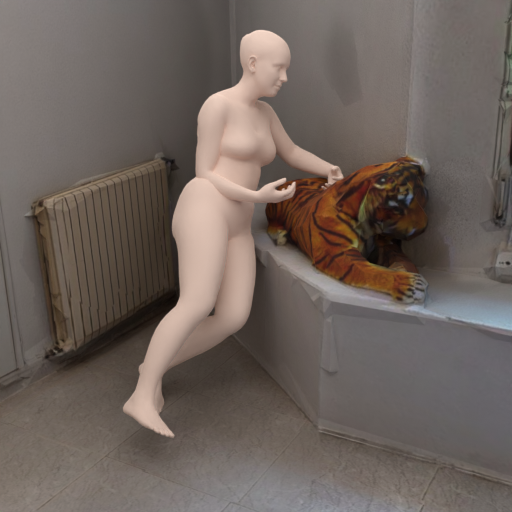} &
    \includegraphics[width=0.16\textwidth]{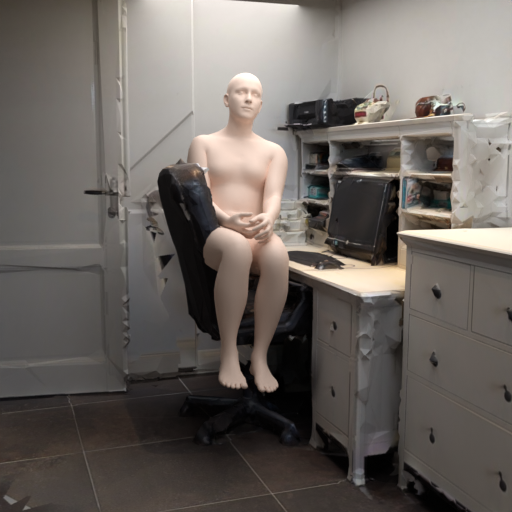}&
    \includegraphics[width=0.16\textwidth]{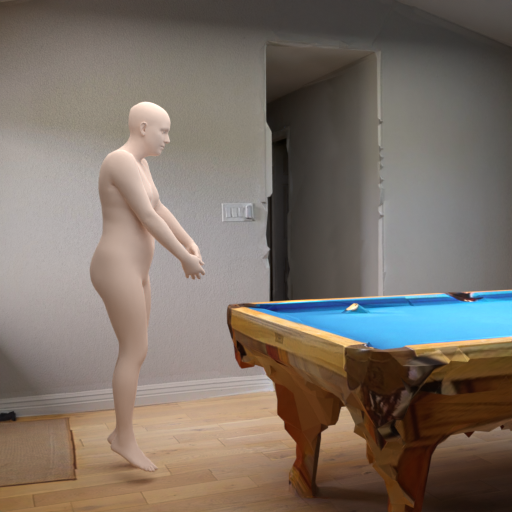} &
    \includegraphics[width=0.16\textwidth]{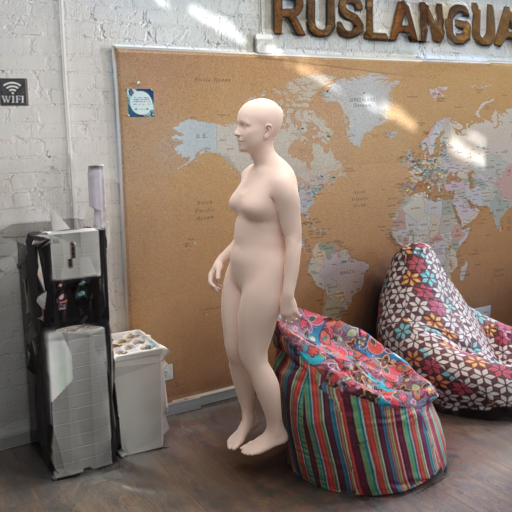} \\

    \rotatebox{90}{\hspace{0.5cm}POSA++} &
    \includegraphics[width=0.16\textwidth]{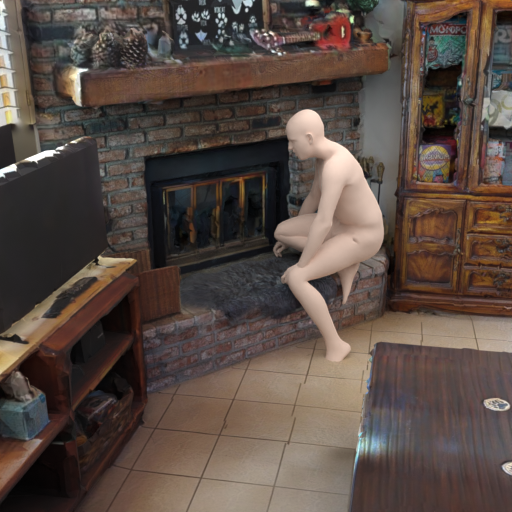} &
    \includegraphics[width=0.16\textwidth]{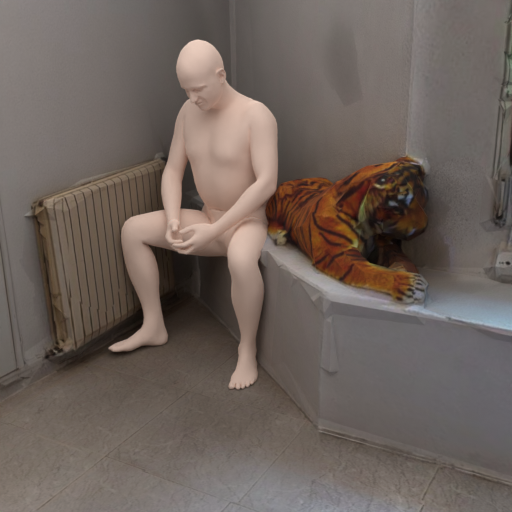} &
    \includegraphics[width=0.16\textwidth]{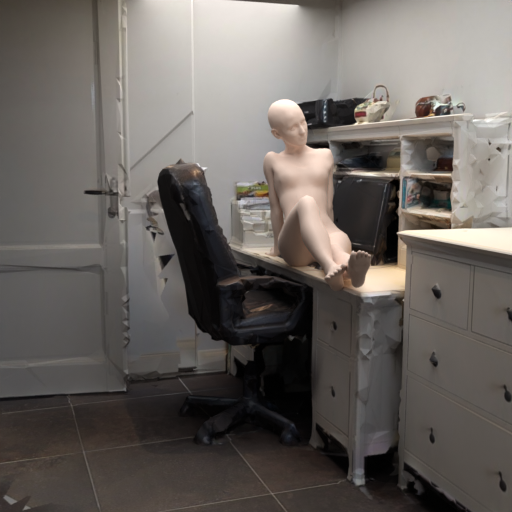} &
    \includegraphics[width=0.16\textwidth]{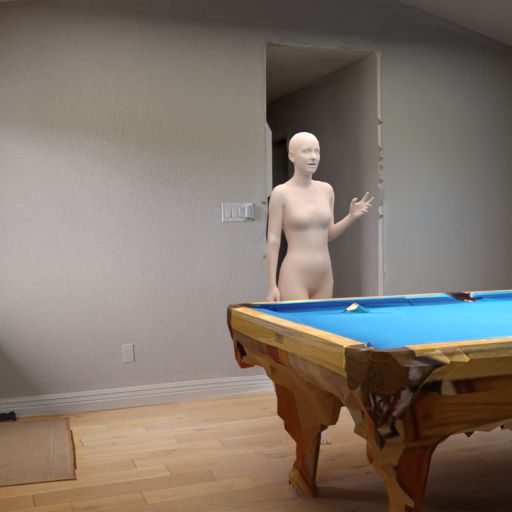} &
    \includegraphics[width=0.16\textwidth]{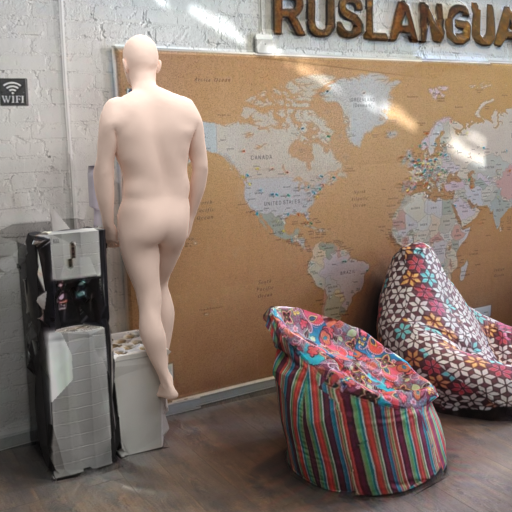} \\

    \rotatebox{90}{\hspace{0.2cm}\ours{} (Ours)} &
    \includegraphics[width=0.16\textwidth]{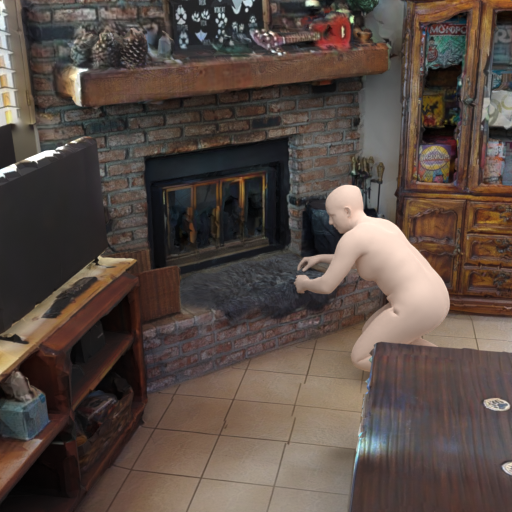} &
    \includegraphics[width=0.16\textwidth]{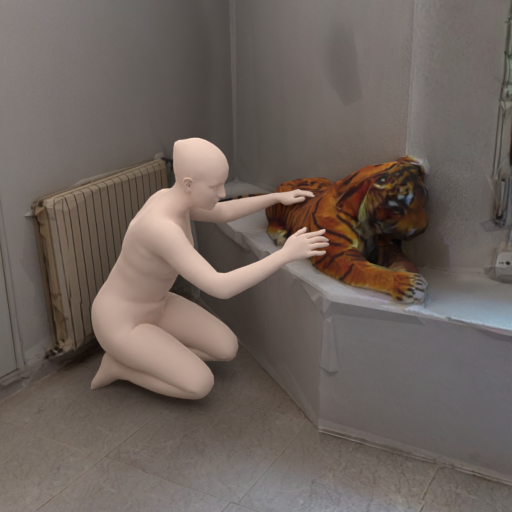} &
    \includegraphics[width=0.16\textwidth]{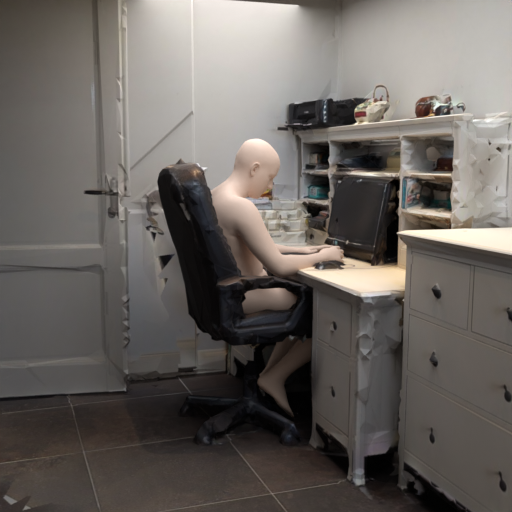} &
    \includegraphics[width=0.16\textwidth]{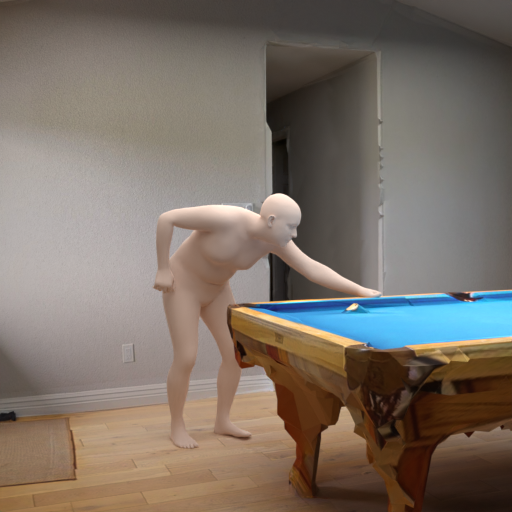} &
    \includegraphics[width=0.16\textwidth]{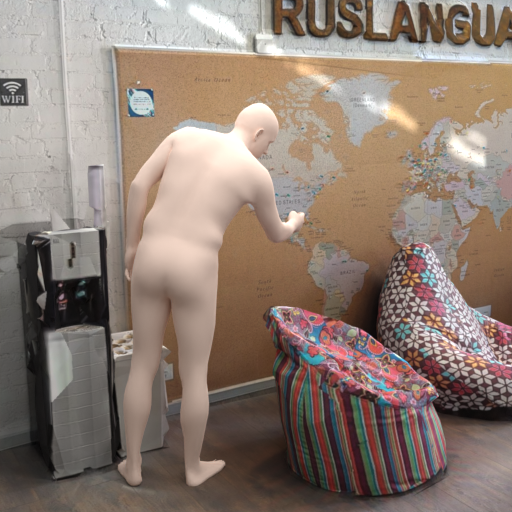} \\
    &
    \makecell{\small Kneeling on\\ \small a fireplace} &
    \makecell{\small Kneeling and \\ \small stroking a tiger} &
    \makecell{\small Sitting in an\\ \small office chair} &
    \makecell{\small Leaning over a \\ \small table with a cue} &
    \makecell{\small Pointing at\\ \small a map} \\

  \end{tabular}
  \caption{Qualitative comparison. Our method generates diverse interactions, including those with rare objects (tiger, fireplace) and touch-less interactions (map). It also produces natural poses fitting for the context of the object (office chair, pool table).
 }
  \label{fig:qualitative-results}
\end{figure*}

While \ours{} is fully automated and does not require user-specified text prompts, for comparability with prior work, we evaluate the text-to-interaction step of our pipeline against the state of the art. 
\emph{GenZI}~\citep{genzi} lifts multi-view Stable Diffusion inpaintings to 3D, given a text prompt and a target location.
\emph{POSA}~\citep{posa} predicts per-vertex semantic contacts  and places the body by contact optimization with the scene geometry. As POSA does not support text input, we extend it to \posa{} by adding text-based pose retrieval \citep{agora, poseembroider}.
We use 200 semantically annotated scenes from HM3D~\citep{puig2023habitat3}, to accommodate \posa{} and sample 150 diverse prompt-location pairs from our generated affordance-aware prompts.
To evaluate prompt consistency we follow the procedure of \cite{genzi} and compute CLIP score.
To assess whether contacts are both physically plausible and semantically meaningful, we introduce semantic contact metrics:
an LLM predicts the body parts expected to be involved in each interaction (e.g., feet and hands for \textit{leaning on a counter}), which serve as pseudo-GT for computing Precision, Recall and F1-score of the body-scene contacts.

\begin{wrapfigure}{r}{0.16\textwidth}
  \vspace{-\intextsep}
  \centering
  \begin{tikzpicture}[x=0.5\linewidth, y=0.5\linewidth]
  \useasboundingbox (-1,-1.778) rectangle (1,1); %
  \definecolor{pieOurs}{HTML}{0077BB}
  \definecolor{pieGenZi}{HTML}{33BBEE}
  \definecolor{piePosa}{HTML}{009988}
  \definecolor{pieDark}{HTML}{1F2937}
  \fill[pieOurs]  (0,0) -- (168.843:1) arc[start angle=168.843, end angle=450,     radius=1] -- cycle;
  \fill[pieGenZi] (0,0) -- (144.545:1) arc[start angle=144.545, end angle=168.843, radius=1] -- cycle;
  \fill[piePosa]  (0,0) -- (90:1)      arc[start angle=90,      end angle=144.545, radius=1] -- cycle;
  \foreach \a in {90, 144.545, 168.843} \draw[white, line width=0.25pt] (0,0) -- (\a:1);
  \begin{scope}[every node/.style={font=\fontsize{5.5}{6}\selectfont\bfseries, inner sep=0pt}]
    \node[text=white]    at (309.422:0.7) {78.1\%};
    \node[text=pieDark]  at (156.694:0.7) {6.7\%};
    \node[text=white]    at (117.273:0.7) {15.2\%};
  \end{scope}
  \begin{scope}[every node/.style={font=\fontsize{6.6}{7}\selectfont, anchor=west, inner sep=0pt}]
    \fill[pieOurs]  (-0.901,-1.080) rectangle (-0.676,-1.211); \node at (-0.522,-1.145) {Ours};
    \fill[pieGenZi] (-0.901,-1.364) rectangle (-0.676,-1.495); \node at (-0.522,-1.429) {GenZI};
    \fill[piePosa]  (-0.901,-1.647) rectangle (-0.676,-1.778); \node at (-0.522,-1.712) {POSA++};
  \end{scope}
\end{tikzpicture}%
  \caption{\raggedright \ours{}'s poses are preferred 78\% of the time.}
  \label{fig:user-study-pie}
\end{wrapfigure}
As shown in \Cref{tab:ours:quant}, \ours{} achieves the highest CLIP score as well as the best semantic contact metrics.
\posa{}'s pure optimization based approach cannot adapt to the scene, and interactions without scene contact, such as playing pool or pointing at a map, cannot be modeled (\Cref{fig:qualitative-results}).
GenZI, in turn, suffers from inconsistencies across its generated views, which lead to inaccurate lifting and implausible configurations (e.g., a foot attached to a wall), reflected in its lower contact recall.
\ours{} generates interactions in image space and reasons about the scene prior to lifting, yielding poses physically and semantically consistent with the scene.

We conducted a perceptual user study, where 41 participants were shown triplets of generations made with \ours{}, GenZI and \posa{} at corresponding scene locations, in random order, as interactive 3D visualizations (interface shown in \Cref{fig:suppl:userstudy}). They were asked to select the single best result, considering the realism of human--scene interaction, physical feasibility and pose naturalness.
Across 726 total responses, \ours{} was preferred 78.1\% of the time (\Cref{fig:user-study-pie}).

\section{Conclusion}
We have presented \ours{}, a fully automatic engine that populates 3D scenes with diverse, scene-appropriate human interactions by transferring the interaction knowledge of 2D foundation models into 3D.
Applied to Habitat-Matterport 3D, it produces \ourdata{}, spanning $\sim$6.6k rooms, where every sample pairs a photorealistic image with SMPL-X bodies and the complete scene geometry.
Training with \ourdata{} improves contact estimation, human--scene reconstruction and generative models of humans in scenes over motion-capture, real captured and exactly-labeled synthetic training data. Currently, \ours{} generates static poses within static 3D scenes, but the principle extends to video generation and dynamic scenes as the underlying generative models continue to evolve.
More broadly, our results show that the coverage of human--scene interactions that capture cannot reach can now be produced automatically, at a scale that makes it useful for training.

\subsubsection*{Acknowledgments}
Special thanks to the RVH members for their help and discussions.
Nikita Kister thanks the European Laboratory for Learning and Intelligent Systems
(ELLIS) PhD program for support. The authors thank the International Max Planck
Research School for Intelligent Systems (IMPRS-IS) for supporting Pradyumna. Istv\'an
S\'ar\'andi and Gerard Pons-Moll were supported by the German Federal Ministry of
Education and Research (BMBF): T{\"u}bingen AI Center, FKZ: 01IS18039A, by the
Deutsche Forschungsgemeinschaft (DFG, German Research Foundation) -- 409792180
(Emmy Noether Programme, project: Real Virtual Humans). GPM is a member of
the Machine Learning Cluster of Excellence, EXC number 2064/1 -- Project number
390727645 and is supported by the Carl Zeiss Foundation.

\bibliography{main}

\begin{thebibliography}{41}
\providecommand{\natexlab}[1]{#1}
\providecommand{\url}[1]{\texttt{#1}}
\expandafter\ifx\csname urlstyle\endcsname\relax
  \providecommand{\doi}[1]{doi: #1}\else
  \providecommand{\doi}{doi: \begingroup \urlstyle{rm}\Url}\fi

\bibitem[Araujo et~al.(2023)Araujo, Li, Vetrivel, Agarwal, Gopinath, Wu, Clegg,
  and Liu]{araujo2023circle}
Joao~Pedro Araujo, Jiaman Li, Karthik Vetrivel, Rishi Agarwal, Deepak Gopinath,
  Jiajun Wu, Alexander Clegg, and C.~Karen Liu.
\newblock {CIRCLE}: Capture in rich contextual environments.
\newblock In \emph{CVPR}, 2023.

\bibitem[Baradel et~al.(2024)Baradel, Armando, Galaaoui, Br{\'e}gier,
  Weinzaepfel, Rogez, and Lucas]{baradel2024multihmr}
Fabien Baradel, Matthieu Armando, Salma Galaaoui, Romain Br{\'e}gier, Philippe
  Weinzaepfel, Gr{\'e}gory Rogez, and Thomas Lucas.
\newblock Multi-{HMR}: Multi-person whole-body human mesh recovery in a single
  shot.
\newblock In \emph{ECCV}, 2024.

\bibitem[Black et~al.(2023)Black, Patel, Tesch, and Yang]{black2023bedlam}
Michael~J. Black, Priyanka Patel, Joachim Tesch, and Jinlong Yang.
\newblock {BEDLAM}: A synthetic dataset of bodies exhibiting detailed lifelike
  animated motion.
\newblock In \emph{CVPR}, 2023.

\bibitem[Cao et~al.(2020)Cao, Gao, Mangalam, Cai, Vo, and
  Malik]{cao2020longterm}
Zhe Cao, Hang Gao, Karttikeya Mangalam, Qi-Zhi Cai, Minh Vo, and Jitendra
  Malik.
\newblock Long-term human motion prediction with scene context.
\newblock In \emph{ECCV}, 2020.

\bibitem[Chen et~al.(2026)Chen, Chen, Xue, Chen, Xiu, and
  {Pons-Moll}]{chen2025human3r}
Yue Chen, Xingyu Chen, Yuxuan Xue, Anpei Chen, Yuliang Xiu, and Gerard
  {Pons-Moll}.
\newblock {Human3R}: Everyone everywhere all at once.
\newblock In \emph{ICLR}, 2026.

\bibitem[Comanici et~al.(2025)Comanici, Bieber, Schaekermann, Pasupat,
  Sachdeva, Dhillon, Blistein, Ram, Zhang, Rosen, et~al.]{gemini2025}
Gheorghe Comanici, Eric Bieber, Mike Schaekermann, Ice Pasupat, Noveen
  Sachdeva, Inderjit Dhillon, Marcel Blistein, Ori Ram, Dan Zhang, Evan Rosen,
  et~al.
\newblock Gemini 2.5: Pushing the frontier with advanced reasoning,
  multimodality, long context, and next generation agentic capabilities.
\newblock \emph{arXiv preprint arXiv:2507.06261}, 2025.

\bibitem[Dai et~al.(2017)Dai, Chang, Savva, Halber, Funkhouser, and
  Nie{\ss}ner]{dai2017scannet}
Angela Dai, Angel~X. Chang, Manolis Savva, Maciej Halber, Thomas Funkhouser,
  and Matthias Nie{\ss}ner.
\newblock {ScanNet}: Richly-annotated {3D} reconstructions of indoor scenes.
\newblock In \emph{CVPR}, 2017.

\bibitem[Delmas et~al.(2022)Delmas, Weinzaepfel, Lucas, Moreno-Noguer, and
  Rogez]{posescript}
Ginger Delmas, Philippe Weinzaepfel, Thomas Lucas, Francesc Moreno-Noguer, and
  Gr{\'e}gory Rogez.
\newblock {PoseScript}: {3D} human poses from natural language.
\newblock In \emph{ECCV}, 2022.

\bibitem[Delmas et~al.(2024)Delmas, Weinzaepfel, Moreno-Noguer, and
  Rogez]{poseembroider}
Ginger Delmas, Philippe Weinzaepfel, Francesc Moreno-Noguer, and Gr{\'e}gory
  Rogez.
\newblock {PoseEmbroider}: Towards a {3D}, visual, semantic-aware human pose
  representation.
\newblock In \emph{ECCV}, 2024.

\bibitem[Ge et~al.(2025)Ge, Wang, Chen, Chen, and Shen]{humanwild2024}
Yongtao Ge, Wenjia Wang, Yongfan Chen, Hao Chen, and Chunhua Shen.
\newblock {3D} human reconstruction in the wild with synthetic data using
  generative models.
\newblock \emph{IEEE Transactions on Pattern Analysis and Machine
  Intelligence}, 2025.

\bibitem[{Google DeepMind}(2025)]{gemini3}
{Google DeepMind}.
\newblock {Gemini} 3.
\newblock \url{https://deepmind.google/models/gemini/}, 2025.

\bibitem[Guzov et~al.(2021)Guzov, Mir, Sattler, and
  {Pons-Moll}]{guzov2021human}
Vladimir Guzov, Aymen Mir, Torsten Sattler, and Gerard {Pons-Moll}.
\newblock Human {POSEitioning} system ({HPS}): {3D} human pose estimation and
  self-localization in large scenes from body-mounted sensors.
\newblock In \emph{CVPR}, 2021.

\bibitem[Guzov et~al.(2024)Guzov, Chibane, Marin, He, Saracoglu, Sattler, and
  {Pons-Moll}]{guzov2024interaction}
Vladimir Guzov, Julian Chibane, Riccardo Marin, Yannan He, Yunus Saracoglu,
  Torsten Sattler, and Gerard {Pons-Moll}.
\newblock Interaction replica: Tracking human-object interaction and scene
  changes from human motion.
\newblock In \emph{3DV}, 2024.

\bibitem[Hassan et~al.(2019)Hassan, Choutas, Tzionas, and
  Black]{hassan2019resolving}
Mohamed Hassan, Vasileios Choutas, Dimitrios Tzionas, and Michael~J. Black.
\newblock Resolving {3D} human pose ambiguities with {3D} scene constraints.
\newblock In \emph{ICCV}, 2019.

\bibitem[Hassan et~al.(2021)Hassan, Ghosh, Tesch, Tzionas, and Black]{posa}
Mohamed Hassan, Partha Ghosh, Joachim Tesch, Dimitrios Tzionas, and Michael~J.
  Black.
\newblock Populating {3D} scenes by learning human-scene interaction.
\newblock In \emph{CVPR}, 2021.

\bibitem[Huang et~al.(2022)Huang, Yi, H{\"o}schle, Safroshkin, Alexiadis,
  Polikovsky, Scharstein, and Black]{huang2022capturing}
Chun-Hao~P. Huang, Hongwei Yi, Markus H{\"o}schle, Matvey Safroshkin,
  Tsvetelina Alexiadis, Senya Polikovsky, Daniel Scharstein, and Michael~J.
  Black.
\newblock Capturing and inferring dense full-body human-scene contact.
\newblock In \emph{CVPR}, 2022.

\bibitem[Huang et~al.(2023)Huang, Wang, Li, Jia, Liu, Zhu, Liang, and
  Zhu]{scenediffuser}
Siyuan Huang, Zan Wang, Puhao Li, Baoxiong Jia, Tengyu Liu, Yixin Zhu, Wei
  Liang, and Song-Chun Zhu.
\newblock Diffusion-based generation, optimization, and planning in {3D}
  scenes.
\newblock In \emph{CVPR}, 2023.

\bibitem[Jiang et~al.(2024)Jiang, Zhang, Li, Ma, Wang, Chen, Liu, Zhu, and
  Huang]{jiang2024scaling}
Nan Jiang, Zhiyuan Zhang, Hongjie Li, Xiaoxuan Ma, Zan Wang, Yixin Chen, Tengyu
  Liu, Yixin Zhu, and Siyuan Huang.
\newblock Scaling up dynamic human-scene interaction modeling.
\newblock In \emph{CVPR}, 2024.

\bibitem[Kim et~al.(2024)Kim, Han, Kwon, and Joo]{coma2024}
Hyeonwoo Kim, Sookwan Han, Patrick Kwon, and Hanbyul Joo.
\newblock Beyond the contact: Discovering comprehensive affordance for {3D}
  objects from pre-trained {2D} diffusion models.
\newblock In \emph{ECCV}, 2024.

\bibitem[Kim et~al.(2025)Kim, Baik, and Joo]{david2025}
Hyeonwoo Kim, Sangwon Baik, and Hanbyul Joo.
\newblock {DAViD}: Modeling dynamic affordance of {3D} objects using
  pre-trained video diffusion models.
\newblock In \emph{ICCV}, 2025.

\bibitem[Li \& Dai(2024)Li and Dai]{genzi}
Lei Li and Angela Dai.
\newblock {GenZI}: Zero-shot {3D} human-scene interaction generation.
\newblock In \emph{CVPR}, 2024.

\bibitem[Liu et~al.(2024)Liu, Zeng, Ren, Li, Zhang, Yang, Li, Yang, Su, Zhu,
  et~al.]{liu2023grounding}
Shilong Liu, Zhaoyang Zeng, Tianhe Ren, Feng Li, Hao Zhang, Jie Yang, Chunyuan
  Li, Jianwei Yang, Hang Su, Jun Zhu, et~al.
\newblock Grounding {DINO}: Marrying {DINO} with grounded pre-training for
  open-set object detection.
\newblock In \emph{ECCV}, 2024.

\bibitem[Mahmood et~al.(2019)Mahmood, Ghorbani, Troje, Pons-Moll, and
  Black]{mahmood2019amass}
Naureen Mahmood, Nima Ghorbani, Nikolaus~F. Troje, Gerard Pons-Moll, and
  Michael~J. Black.
\newblock {AMASS}: Archive of motion capture as surface shapes.
\newblock In \emph{ICCV}, 2019.

\bibitem[Patel et~al.(2021)Patel, Huang, Tesch, Hoffmann, Tripathi, and
  Black]{agora}
Priyanka Patel, Chun-Hao~P. Huang, Joachim Tesch, David~T. Hoffmann, Shashank
  Tripathi, and Michael~J. Black.
\newblock {AGORA}: Avatars in geography optimized for regression analysis.
\newblock In \emph{CVPR}, 2021.

\bibitem[Pavlakos et~al.(2019)Pavlakos, Choutas, Ghorbani, Bolkart, Osman,
  Tzionas, and Black]{pavlakos2019expressive}
Georgios Pavlakos, Vasileios Choutas, Nima Ghorbani, Timo Bolkart, Ahmed~A.
  Osman, Dimitrios Tzionas, and Michael~J. Black.
\newblock Expressive body capture: {3D} hands, face, and body from a single
  image.
\newblock In \emph{CVPR}, 2019.

\bibitem[Puig et~al.(2024)Puig, Undersander, Szot, Cote, Batra, and
  Mottaghi]{puig2023habitat3}
Xavier Puig, Eric Undersander, Andrew Szot, Mikael~Dallaire Cote, Dhruv Batra,
  and Roozbeh Mottaghi.
\newblock Habitat 3.0: A co-habitat for humans, avatars and robots.
\newblock In \emph{ICLR}, 2024.

\bibitem[Ramakrishnan et~al.(2021)Ramakrishnan, Gokaslan, Wijmans, Maksymets,
  Clegg, Turner, Undersander, Galuba, Westbury, Chang, Savva, Zhao, and
  Batra]{hm3d}
Santhosh~Kumar Ramakrishnan, Aaron Gokaslan, Erik Wijmans, Oleksandr Maksymets,
  Alexander Clegg, John~M. Turner, Eric Undersander, Wojciech Galuba, Andrew
  Westbury, Angel~X. Chang, Manolis Savva, Yili Zhao, and Dhruv Batra.
\newblock {Habitat-Matterport} {3D} dataset ({HM3D}): 1000 large-scale {3D}
  environments for embodied {AI}.
\newblock In \emph{NeurIPS Datasets and Benchmarks Track}, 2021.

\bibitem[Ravi et~al.(2025)Ravi, Gabeur, Hu, Hu, Ryali, Ma, Khedr, R{\"a}dle,
  Rolland, Gustafson, Mintun, Pan, Alwala, Carion, Wu, Girshick, Doll{\'a}r,
  and Feichtenhofer]{ravi2024sam2}
Nikhila Ravi, Valentin Gabeur, Yuan-Ting Hu, Ronghang Hu, Chaitanya Ryali,
  Tengyu Ma, Haitham Khedr, Roman R{\"a}dle, Chloe Rolland, Laura Gustafson,
  Eric Mintun, Junting Pan, Kalyan~Vasudev Alwala, Nicolas Carion, Chao-Yuan
  Wu, Ross Girshick, Piotr Doll{\'a}r, and Christoph Feichtenhofer.
\newblock {SAM} 2: Segment anything in images and videos.
\newblock In \emph{ICLR}, 2025.

\bibitem[S\'ar\'andi \& Pons-Moll(2024)S\'ar\'andi and
  Pons-Moll]{sarandi2024nlf}
Istv\'an S\'ar\'andi and Gerard Pons-Moll.
\newblock Neural localizer fields for continuous {3D} human pose and shape
  estimation.
\newblock In \emph{NeurIPS}, 2024.

\bibitem[Savva et~al.(2016)Savva, Chang, Hanrahan, Fisher, and
  Nie{\ss}ner]{savva2016pigraphs}
Manolis Savva, Angel~X. Chang, Pat Hanrahan, Matthew Fisher, and Matthias
  Nie{\ss}ner.
\newblock {PiGraphs}: Learning interaction snapshots from observations.
\newblock \emph{ACM Transactions on Graphics (TOG)}, 35\penalty0 (4), 2016.

\bibitem[Tripathi et~al.(2023{\natexlab{a}})Tripathi, Chatterjee, Passy, Yi,
  Tzionas, and Black]{tripathi2023deco}
Shashank Tripathi, Agniv Chatterjee, Jean-Claude Passy, Hongwei Yi, Dimitrios
  Tzionas, and Michael~J. Black.
\newblock {DECO}: Dense estimation of {3D} human-scene contact in the wild.
\newblock In \emph{ICCV}, 2023{\natexlab{a}}.

\bibitem[Tripathi et~al.(2023{\natexlab{b}})Tripathi, M{\"u}ller, Huang,
  Taheri, Black, and Tzionas]{tripathi2023ipman}
Shashank Tripathi, Lea M{\"u}ller, Chun-Hao~P. Huang, Omid Taheri, Michael~J.
  Black, and Dimitrios Tzionas.
\newblock {3D} human pose estimation via intuitive physics.
\newblock In \emph{CVPR}, 2023{\natexlab{b}}.

\bibitem[Wang et~al.(2025)Wang, Xu, Dong, Deng, Xiang, Lv, Sun, Tong, and
  Yang]{wang2025moge2}
Ruicheng Wang, Sicheng Xu, Yue Dong, Yu~Deng, Jianfeng Xiang, Zelong Lv,
  Guangzhong Sun, Xin Tong, and Jiaolong Yang.
\newblock {MoGe-2}: Accurate monocular geometry with metric scale and sharp
  details.
\newblock In \emph{NeurIPS}, 2025.

\bibitem[Wang et~al.(2024)Wang, Mao, Zhu, Xu, Lyu, Li, Chen, Zhang, Chen, Xue,
  et~al.]{wang2024embodiedscan}
Tai Wang, Xiaohan Mao, Chenming Zhu, Runsen Xu, Ruiyuan Lyu, Peisen Li, Xiao
  Chen, Wenwei Zhang, Kai Chen, Tianfan Xue, et~al.
\newblock {EmbodiedScan}: A holistic multi-modal {3D} perception suite towards
  embodied {AI}.
\newblock In \emph{CVPR}, 2024.

\bibitem[Wang et~al.(2022)Wang, Chen, Liu, Zhu, Liang, and
  Huang]{wang2022humanise}
Zan Wang, Yixin Chen, Tengyu Liu, Yixin Zhu, Wei Liang, and Siyuan Huang.
\newblock {HUMANISE}: Language-conditioned human motion generation in {3D}
  scenes.
\newblock In \emph{NeurIPS}, 2022.

\bibitem[Xu et~al.(2022)Xu, Zhang, Zhang, and Tao]{xu2022vitpose}
Yufei Xu, Jing Zhang, Qiming Zhang, and Dacheng Tao.
\newblock {ViTPose}: Simple vision transformer baselines for human pose
  estimation.
\newblock In \emph{NeurIPS}, 2022.

\bibitem[Yalandur~Muralidhar et~al.(2025)Yalandur~Muralidhar, Xue, Xie,
  Kostyrko, and Pons-Moll]{ym2025physic}
Pradyumna Yalandur~Muralidhar, Yuxuan Xue, Xianghui Xie, Margaret Kostyrko, and
  Gerard Pons-Moll.
\newblock {PhySIC}: Physically plausible {3D} human-scene interaction and
  contact from a single image.
\newblock In \emph{SIGGRAPH Asia}, 2025.

\bibitem[YM et~al.(2026)YM, Xue, Chen, Kister, S{\'a}r{\'a}ndi, and
  Pons-Moll]{graft}
Pradyumna YM, Yuxuan Xue, Yue Chen, Nikita Kister, Istv{\'a}n S{\'a}r{\'a}ndi,
  and Gerard Pons-Moll.
\newblock {GRAFT}: Geometric refinement and fitting transformer for human scene
  reconstruction.
\newblock In \emph{ECCV}, 2026.

\bibitem[Zhang et~al.(2020)Zhang, Zhang, Ma, Black, and Tang]{place}
Siwei Zhang, Yan Zhang, Qianli Ma, Michael~J. Black, and Siyu Tang.
\newblock {PLACE}: Proximity learning of articulation and contact in {3D}
  environments.
\newblock In \emph{3DV}, 2020.

\bibitem[Zhang et~al.(2022)Zhang, Bhatnagar, Guzov, Starke, and
  {Pons-Moll}]{zhang2022couch}
Xiaohan Zhang, Bharat~Lal Bhatnagar, Vladimir Guzov, Sebastian Starke, and
  Gerard {Pons-Moll}.
\newblock {COUCH}: Towards controllable human-chair interactions.
\newblock In \emph{ECCV}, 2022.

\bibitem[Zhen et~al.(2024)Zhen, Qiu, Chen, Yang, Yan, Du, Hong, and
  Gan]{zhen2024threedvla}
Haoyu Zhen, Xiaowen Qiu, Peihao Chen, Jincheng Yang, Xin Yan, Yilun Du, Yining
  Hong, and Chuang Gan.
\newblock {3D-VLA}: A {3D} vision-language-action generative world model.
\newblock In \emph{ICML}, 2024.

\end{thebibliography}
\bibliographystyle{iclr2027_conference}

\newpage
\appendix
\section{Appendix Summary}
In this appendix we provide additional implementation details of \ours{} in \Cref{suppl:sec:impl} and discuss limitations in \Cref{suppl:sec:limits}.
In \Cref{suppl:sec:full} we ablate the use of the complete 3D scene during the lifting step.
In \Cref{suppl:sec:scale} we ablate the explicit scale parameter in the lifting stage.
In \Cref{sec:what-makes-useful} we ablate the interaction and scene diversity of the training data.
In \Cref{suppl:sec:stability} we compare the physical stability of \ourdata{} poses with existing datasets.
We show more qualitative results in \Cref{suppl:sec:data}.
Finally, in \Cref{suppl:sec:failures} we show failure cases of \ours{} and how they can be addressed.
A video showcasing the population of a scene and an interactive visualization of our generations are available on the project page (\url{https://virtualhumans.mpi-inf.mpg.de/inhabit/}).

\section{Limitations}\label{suppl:sec:limits}
In this work we create scene interactions in 3D by first generating them in 2D.
Our goal was to create a dataset of 3D scenes with corresponding realistic images.
To make this tractable, we try to limit the scene changes the image generative models introduce to the image and use filtering as a post-processing step. This discards a lot of useful interactions, e.g., opening shelves, moving chairs.
Extending \ours{} to model changes to the scene is an interesting future direction and would be a valuable addition to the dataset.
Furthermore, we focus on static interactions, so our people are frozen in time, but extending the dataset with dynamic interactions and moving people would create valuable data for applications that need to perceive people in scenes.

\section{Implementation Details}
\label{suppl:sec:impl}

In this section we provide additional implementation details about view selection, prompting, reconstruction, the user study interface and quality control.
\subsection{Automatic View Selection Heuristic}
\label{suppl:sec:views}
The goal of this heuristic is to select views that show the floor and see sufficiently deep into the scene.
Therefore, two regions of the image are important: the bottom strip and the central part.
We define two masks, $M_c$ and $M_b$. The first, $M_c$, covers a circular region in the center; its objective is to reject views with shallow depth.
The second, $M_b$, covers a strip at the bottom of the image to reject views where the floor is occluded.
In both cases, we compare the depth within these masks against thresholds $\tau_c$ and $\tau_b$, respectively.
A rejection score $S_i$ is calculated based on the proportion of pixels below these thresholds.

More formally:
\begin{equation}
S_i = \frac{1}{|M_i|} \sum_{p \in M_i} \mathbf{1}_{D(p) < \tau_i}, \quad i \in \{c, b\}
\end{equation}

where $\mathbf{1}$ is the indicator function and $|M_i|$ is 
the pixel count of the mask. A candidate view is then rejected if the score for either region exceeds a predefined tolerance $\rho_i$:

\begin{equation}
\text{Reject if: } (S_c > \rho_c) \lor (S_b > \rho_b)
\end{equation}

We set the radius of the circle to a quarter of the image height, $\tau_c=2$\,m, $\tau_b$ is set to the height of the camera, $\rho_c=0.5$ and $\rho_b=0.3$.

\subsection{Prompting}
\label{suppl:sec:prompting}

Our affordance-driven prompt is:
``\textit{Give a list of human actions possible within this scene. Only suggest interactions that involve dynamic interactions and or touch. Avoid only standing or walking. Only consider visible elements that are closer to the camera. First think about the scene then generate the list. Each action should include a description of the object and also the description of the body posture. Combine them, do not split them up between different keys. Generate 4 responses with their corresponding probabilities, sampled from the full distribution. Return only VALID JSON. Use only the keys 'action' and 'probability'. Separate the json part from the rest with BEGIN.}''

For our action-driven prompt we first sample three actions from a list of selected actions and include them in the following prompt:
``\textit{Use these actions: 'ACTION-1', 'ACTION-2', 'ACTION-3' and suggest what an action in the scene depicted in the image that involve each of these actions. Only involve visible elements that are closer to the camera. First think about the scene then generate the list. Each action should include one of the given words, a description of the object and also the description of the body posture. Combine them, do not split them up between different keys. Generate 3 responses with their corresponding probabilities, sampled from the full distribution, each response should use only one of the given verbs. Return only VALID JSON. Use only the keys 'action' and 'probability'. Separate the json part from the rest with BEGIN.}''

For multi-person interactions we use:
``\textit{Give a list of human actions possible within this scene. Only suggest interactions that involve dynamic interactions and or touch and multiple people (but at most 4). Avoid only standing or walking. Only consider visible elements that are closer to the camera. First think about the scene then generate the list. Each action should include a description of the object and also the description of the body posture, and the number of people. Combine them, do not split them up between different keys. Generate 2 responses with their corresponding probabilities, sampled from the full distribution. Each response should have only a single interaction. Return only VALID JSON. Use only the keys 'action' and 'probability'. Separate the json part from the rest with BEGIN.}''

We extract the interactions from the JSON string and use them within the following prompt template as the input to the image editing model:
``\textit{Using this provided image of this scene please add \{Interaction\}. Keep the rest of the room unchanged. Do not move any furniture. Keep everything closed. Do not add any other new objects besides what is explicitly requested.}''

\subsection{Reconstruction Optimization}
\label{suppl:sec:recon}
\paragraph{Off-the-shelf components} Humans in the edited image are segmented with SAM~2~\citep{ravi2024sam2} prompted by Grounding DINO~\citep{liu2023grounding} detections (the LangSAM implementation). 2D keypoints are estimated with ViTPose~\citep{xu2022vitpose}, the initial SMPL-X body with NLF~\citep{sarandi2024nlf}, and depth with MoGe-2~\citep{wang2025moge2}.

Our lifting objective follows standard physically grounded human--scene fitting, similar to PhySIC~\citep{ym2025physic} and PROX~\citep{hassan2019resolving}. We optimize pose $\thetas$, shape $\betas$, camera-relative translation $\transl$, and scale $s$:

\begin{equation}
\begin{aligned}
\min_{\boldsymbol{\theta},\,\boldsymbol{\beta},\,\transl,\,s}\quad
& \lambda_{\text{proj}} \mathcal{L}_{\text{proj}} +
  \lambda_{\text{depth}} \mathcal{L}_{\text{depth}} +
  \lambda_{\text{con}} \mathcal{L}_{\text{contact}} \\
& + \lambda_{\text{pen}} \mathcal{L}_{\text{penetration}} +
  \lambda_{\text{reg}} \mathcal{L}_{\text{reg}}.
\end{aligned}
\end{equation}

Let $\mathcal{M}(\thetas,\betas,\transl,s)$ denote the posed SMPL-X mesh~\citep{pavlakos2019expressive}, let $\phi(\cdot)$ denote the signed distance field of the scene, let $\mathcal{C}$ be the set of contact-prone body vertices obtained from the DECO contact labels, and let $\mathcal{P}_{\text{depth}}$ be the partial human point cloud obtained from the estimated human depth map. We denote by $\mathcal{V}_{\text{vis}}$ the subset of mesh vertices visible under ray casting, and by $\mathbf{J}$ and $\widehat{\mathbf{J}}$ the projected and detected 2D joints, respectively.

We define the losses as follows. The contact loss encourages contact-prone vertices to lie close to the scene surface,
\begin{equation}
\mathcal{L}_{\text{contact}} = \frac{1}{|\mathcal{C}|} \sum_{\mathbf{v} \in \mathcal{C}} \left| \phi(\mathbf{v}) \right|,
\end{equation}
while the penetration loss penalizes vertices that lie inside the scene,
\begin{equation}
\mathcal{L}_{\text{penetration}} = \frac{1}{|\mathcal{V}|} \sum_{\mathbf{v} \in \mathcal{V}} \left| \min\big(\phi(\mathbf{v}), 0\big) \right|,
\end{equation}
where $\mathcal{V}$ denotes all mesh vertices. Both terms use an $L_1$ penalty on the scene SDF.

To align the visible body surface with the human depth estimate, we use a standard Chamfer distance between the ray-casting-visible mesh vertices and the depth-derived point cloud:
\begin{equation}
\mathcal{L}_{\text{depth}} = \mathrm{CD}(\mathcal{V}_{\text{vis}}, \mathcal{P}_{\text{depth}}).
\end{equation}
For the 2D term, we project joints into normalized pixel coordinates in $[0,1]$ and penalize reprojection error as
\begin{equation}
\mathcal{L}_{\text{proj}} = \frac{1}{|\mathbf{J}|} \sum_i \left\| \mathbf{J}_i - \widehat{\mathbf{J}}_i \right\|_2^2.
\end{equation}
Finally, we regularize the optimized parameters toward their initialization $(\thetas_0, \betas_0, \transl_0, s_0)$ using an $L_2$ penalty:
\begin{equation}
\mathcal{L}_{\text{reg}} = \left\| \thetas - \thetas_0 \right\|_2^2 + \left\| \betas - \betas_0 \right\|_2^2 + \left\| \transl - \transl_0 \right\|_2^2 + \left| s - s_0 \right|^2.
\end{equation}

In all experiments, we use $\lambda_{\text{con}} = 0.7$, $\lambda_{\text{pen}} = 1.1$, $\lambda_{\text{depth}} = 0.005$, $\lambda_{\text{reg}} = 13$, and $\lambda_{\text{proj}} = 3000$.

\subsection{User Study}
\begin{figure}[t]
    \centering
    \includegraphics[width=0.8\linewidth]{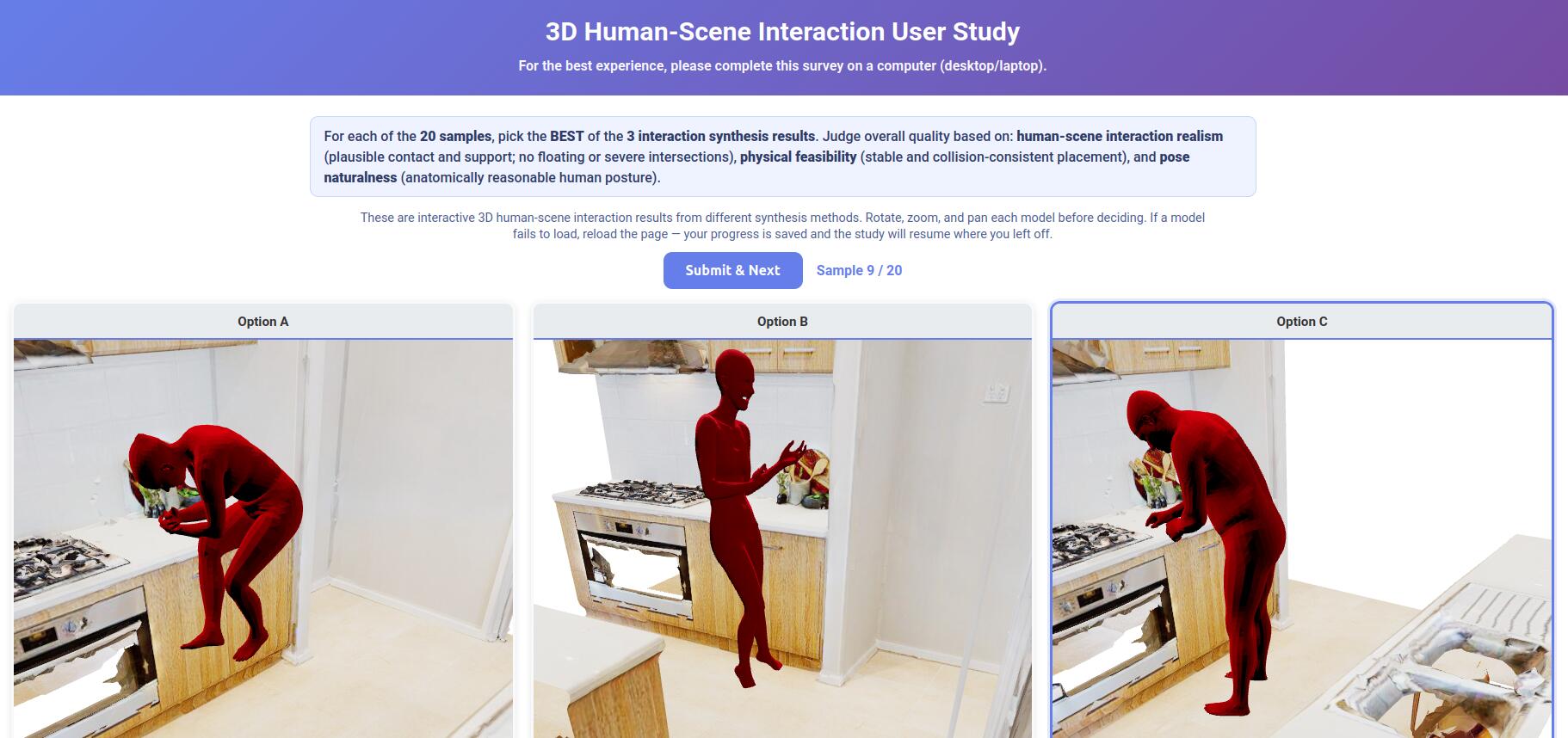}
    \caption{User interface for our user study. The camera automatically orbits around the center of the human bounding box. Additionally, participants can rotate, translate and zoom each scene individually.}
    \label{fig:suppl:userstudy}
\end{figure}

We present the user interface of our user study in \Cref{fig:suppl:userstudy}.

\subsection{Quality Control}
\label{suppl:sec:qc}

\begin{figure*}[t]
  \centering
  \setlength{\tabcolsep}{5pt}
  \renewcommand{\arraystretch}{0.3}
  \begin{tabular}{c c@{\hspace{1pt}} c c@{\hspace{1pt}} c}
    \textbf{Poor Visibility}  & \multicolumn{2}{c}{\textbf{Scene Change}} & \multicolumn{2}{c}{\textbf{Deformed Mesh}}\\
    \includegraphics[width=0.30\textwidth, trim=0cm 0cm 0cm 0.15cm, clip]{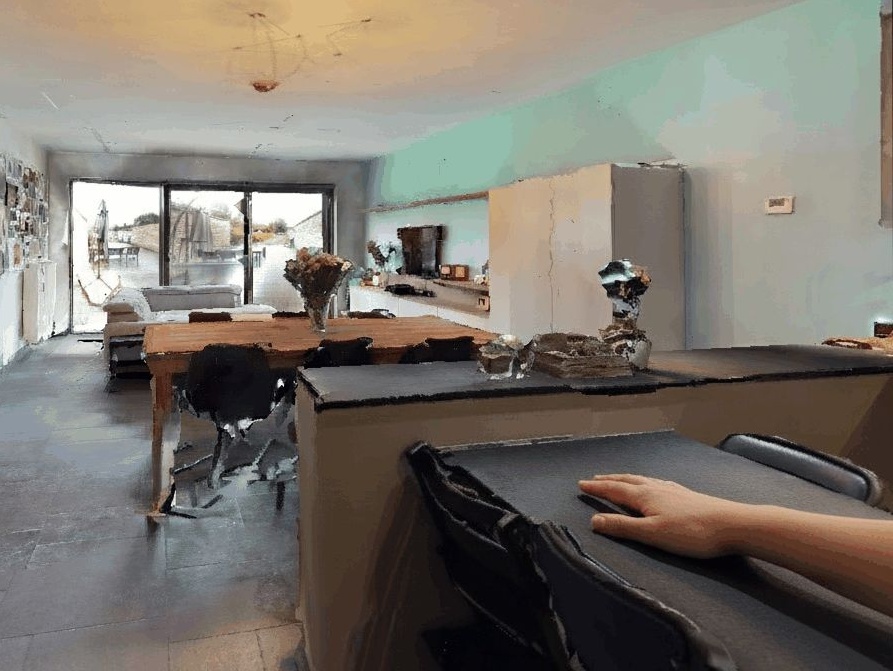} &
    \includegraphics[width=0.15\textwidth, trim=0cm 0cm 0cm 0.19cm, clip]{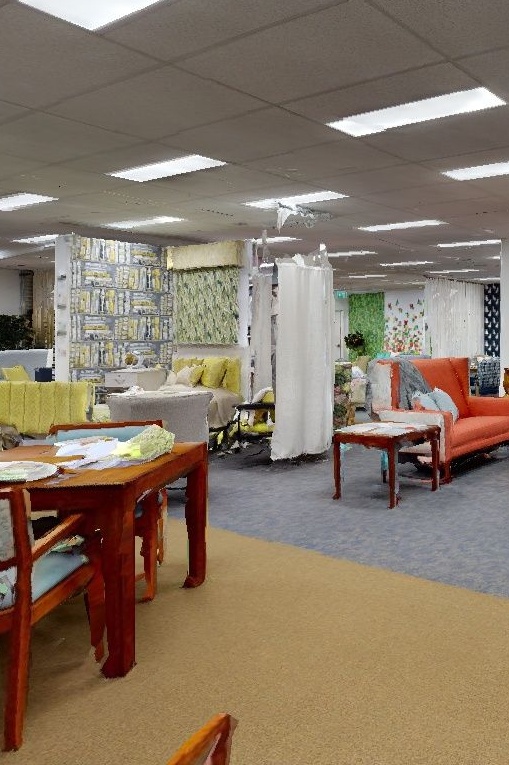} &
    \includegraphics[width=0.15\textwidth, trim=0cm 0cm 0cm 0.21cm, clip]{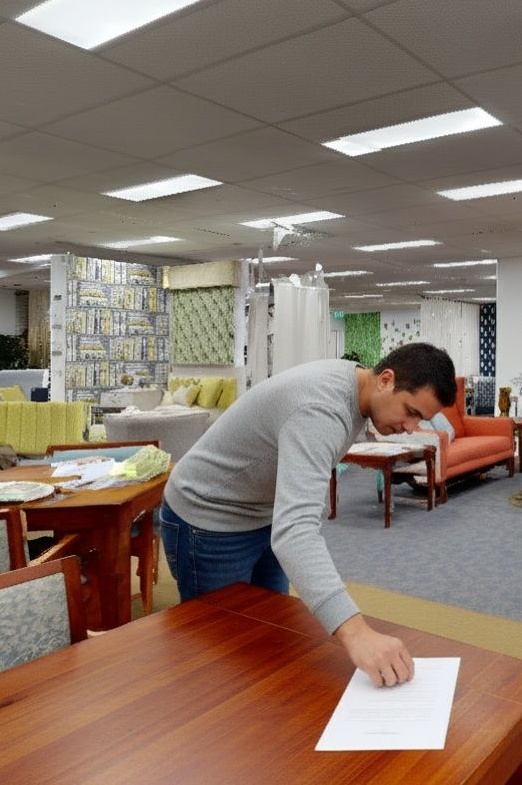} &
    \includegraphics[width=0.15\textwidth]{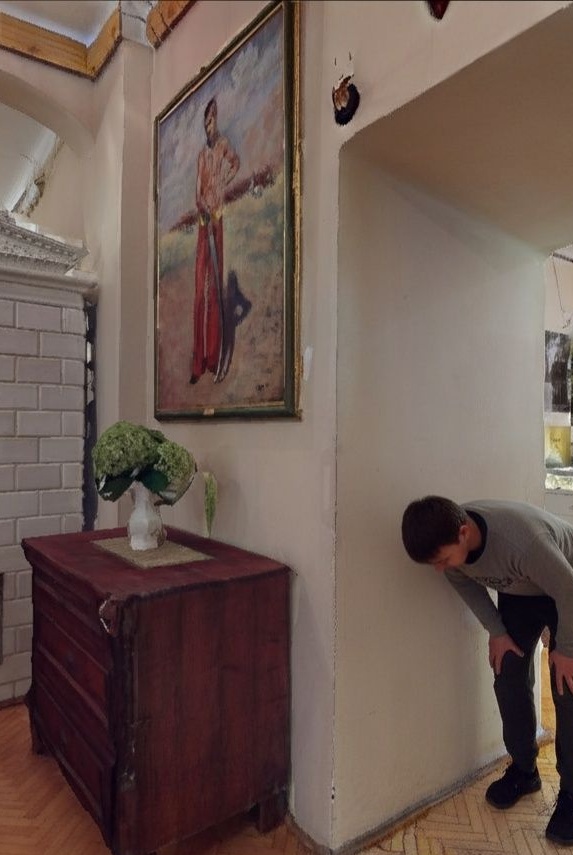} &
    \includegraphics[width=0.15\textwidth, trim=0cm 0cm 0cm 0.19cm, clip]{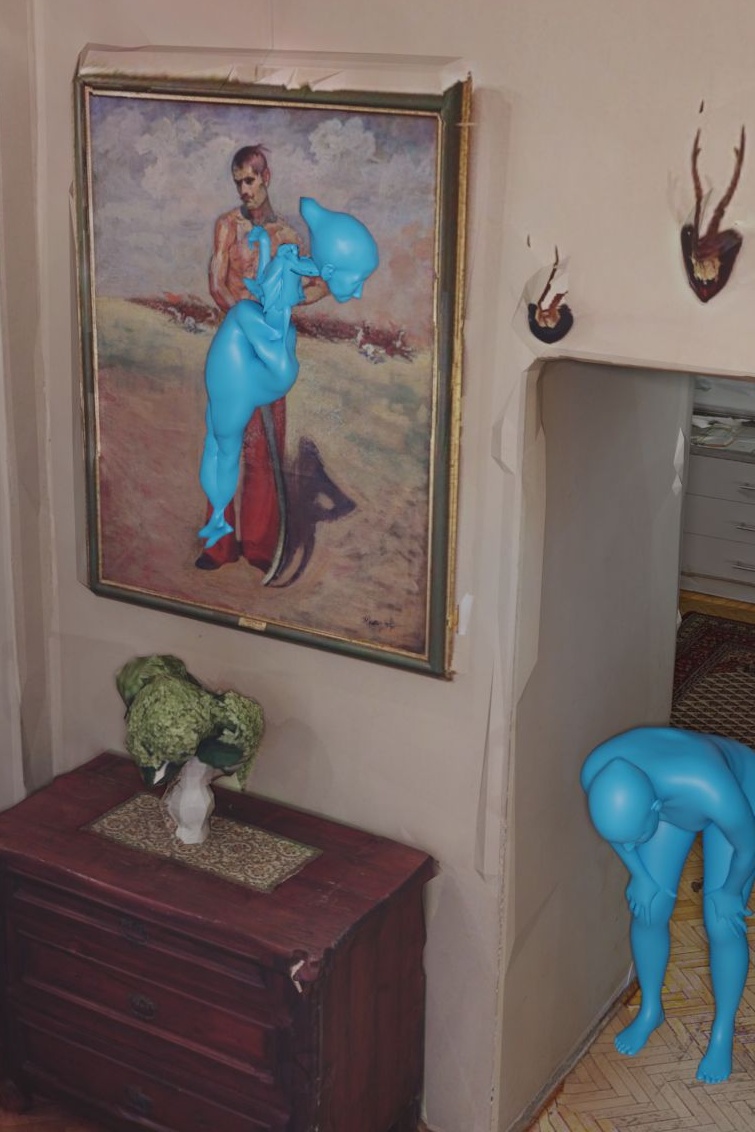} \\
    {\tiny\color{black!70} Edited} & {\tiny\color{black!70} Input} & {\tiny\color{black!70} Edited} & {\tiny\color{black!70} Edited} & {\tiny\color{black!70} Estimation} \\
  \end{tabular}
  \caption{Filtered examples from our quality-control stage. We discard generations that provide too little visible evidence for reliable lifting, alter nearby scene content, or lead to distorted body fits.}
  \label{fig:suppl:quality-control}
\end{figure*}

Some edited images are not suitable for lifting into clean 3D human--scene interactions. We therefore apply three complementary post-reconstruction filters before accepting a sample into the dataset: 1) a \emph{visibility filter} that rejects strongly truncated insertions, 2) a \emph{depth-based filter} that rejects edits which alter local scene structure, and 3) a \emph{volume filter} that rejects distorted SMPL-X fits. We show examples in \Cref{fig:suppl:quality-control}.

\paragraph{Visibility Filter} The editing model can place humans in reflective surfaces (e.g., mirrors or glass) without a corresponding physical person in the scene, or produce severely truncated insertions (e.g., only a hand). To detect these cases, let $\mathcal{H}_i$ denote the reconstructed human mesh for human $i$ in an image. We uniformly sample a set of $N=20{,}000$ points $\{ \mathbf{p}_j \}_{j=1}^{N}$ on $\mathcal{H}_i$ and project them onto the image plane using the camera projection $\pi(\cdot)$. Given a human segmentation mask $M_i$ in the image, we collect the \emph{mask-consistent} points whose projections fall inside $M_i$:
\[
V_i = \{ \mathbf{p}_j : \pi(\mathbf{p}_j) \in M_i \},
\]
and define the \emph{visibility ratio} as
\[
\mathrm{vis}(\mathcal{H}_i) = \frac{|V_i|}{N}.
\]
Humans with $\mathrm{vis}(\mathcal{H}_i) < \tau_{\mathrm{vis}}=0.5$ are discarded. If no humans remain in an image after this step, the image is dropped entirely, indicating the editing model failed to produce a valid insertion.

\paragraph{Depth-Based Filter} The editing model can also introduce unintended local geometry changes near the inserted human, such as hallucinated or displaced furniture. To detect these cases, let $D_{\text{edit}}$ denote the monocular depth map estimated from the edited image, aligned to metric scale using the known scene geometry, and let $M = \bigcup_i M_i$ be the combined binary mask of all humans. We define a thin boundary region around the humans as $M^{\text{b}} = \mathrm{dilate}(M) - M$, where the dilation is performed using OpenCV with a $5\times5$ kernel for $3$ iterations and then a $9\times9$ kernel for another $3$ iterations. We focus on this near-human band because edits are typically local to inserted people; distant regions can be preserved by compositing the near-human edited patch back into the original image. For each pixel $\mathbf{q}$ in this boundary, we compare $D_{\text{edit}}$ against the known scene depth $D_{\text{scene}}$ and compute the fraction of pixels whose \emph{relative} depth discrepancy (normalized by the edited-image boundary depth, so the criterion measures a relative mismatch rather than an absolute metric offset) exceeds a per-pixel tolerance $\tau_{\mathrm{depth}}=0.2$:
\[
\Delta = \frac{\left| \left\{ \mathbf{q} \in M^{\text{b}} : \frac{|D_{\text{edit}}(\mathbf{q}) - D_{\text{scene}}(\mathbf{q})|}{D_{\text{edit}}(\mathbf{q})} > \tau_{\mathrm{depth}} \right\} \right|}{|M^{\text{b}}|}.
\]
If the violation ratio $\Delta$ exceeds an image-level threshold $\tau_{\mathrm{border}}=0.35$, the reconstruction is discarded. This operates as a joint, image-level filter, since geometry edits can extend beyond any single person's boundary.

\paragraph{Volume Filter} The image editing model can occasionally insert children, which fall outside the adult body model we employ (SMPL-X), and the human detector can misfire on depictions of people in portraits, posters, or sculptures. In both cases, fitting an adult SMPL-X mesh to these detections produces highly deformed body shapes: either unnaturally compressed to match a child's proportions or distorted to conform to a non-human surface. We observe that these failure modes consistently yield meshes with anomalously low body volume. Motivated by this, we compute the volume $\mathcal{V}(\mathcal{H}_i)$ of each reconstructed SMPL-X mesh $\mathcal{H}_i$ and discard any instance with $\mathcal{V}(\mathcal{H}_i) < \tau_{\mathrm{vol}}=0.0375\,\si{\meter^3}$. Despite its simplicity, this filter is highly effective: it removes the vast majority of problematic cases and, in practice, performs comparably to more expensive alternatives such as dedicated human detectors (e.g., Multi-HMR~\citep{baradel2024multihmr}) or pose-prior-based filtering (e.g., VPoser~\citep{pavlakos2019expressive}).

Together, these three filters ensure that the final dataset contains only humans that are sufficiently visible, properly aligned, plausibly shaped, and do not introduce implausible scene changes.

\subsection{Runtime and Scalability}
\label{suppl:sec:runtime}
\ours{} is designed for dataset-scale inference. All pre-optimization stages---segmentation, 2D keypoints, 3D pose, contact prediction, and depth---are batched across images and across humans within an image, while the optimization stage processes all humans in a given image in a single forward pass. Each worker fits within 5--8\,GB of VRAM, so the pipeline runs on consumer-grade GPUs (we use 32 RTX 2080s on a shared cluster). Lifting takes $\approx$25\,GPU-seconds per image, with image generation and rendering adding 3--4\,s. Interaction proposals and image edits are obtained through the Gemini API, using Gemini 2.5 Flash~\citep{gemini2025} both to propose interactions and, in its image variant (Gemini 2.5 Flash Image), to edit the images. Reconstructing the full $\sim$78k-image dataset takes $\approx$19\,h of wall-clock time. The low memory footprint and horizontal scaling let \ours{} scale on cheap, abundant hardware.

\begin{table}[t]
  \centering
      \caption{\textbf{Ablation of the reconstruction.} Contact-based Precision, Recall, and F1 of \ours{}'s lifting stage when ablating 1)~the explicit human-scale parameter (\emph{w/o human scale}), and 2)~the use of complete scene geometry, falling back to partial depth (\emph{w/o gt scene}). Since generated images have no ground-truth contacts, we evaluate against coarse, region-level contact labels derived from a VLM; the values are therefore meaningful relative to each other rather than in absolute terms. Both components contribute to final reconstruction quality.}
  \setlength{\tabcolsep}{6pt}
  \renewcommand{\arraystretch}{1.05}
  \begin{tabular}{lccc}
    \toprule
    Reconstruction & Prec. $\uparrow$ & Rec.  $\uparrow$ & F1 $\uparrow$ \\
    \midrule
    \ours{} (Ours)   & \textbf{0.424} & \textbf{0.393} & \textbf{0.368}  \\
    w/o human scale & 0.396 & 0.359 & 0.334  \\
    w/o gt scene    & 0.354 & 0.337 & 0.297  \\
    \bottomrule
  \end{tabular}

  \label{tab:abl:rec}
\end{table}

\section{Full Scene vs Partial Depth}
\label{suppl:sec:full}

\begin{figure*}[t]
  \centering
  \setlength{\tabcolsep}{1pt}
  \renewcommand{\arraystretch}{0.7}
  \begin{tabular}{c c c}
    \textbf{Input} & \textbf{w/ full scene} & \textbf{w/o full scene} \\
    \includegraphics[width=0.18\textwidth]{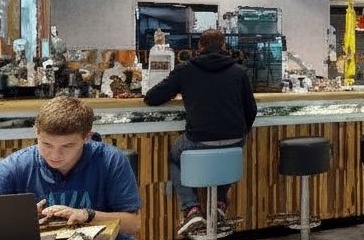} &
    \includegraphics[width=0.18\textwidth]{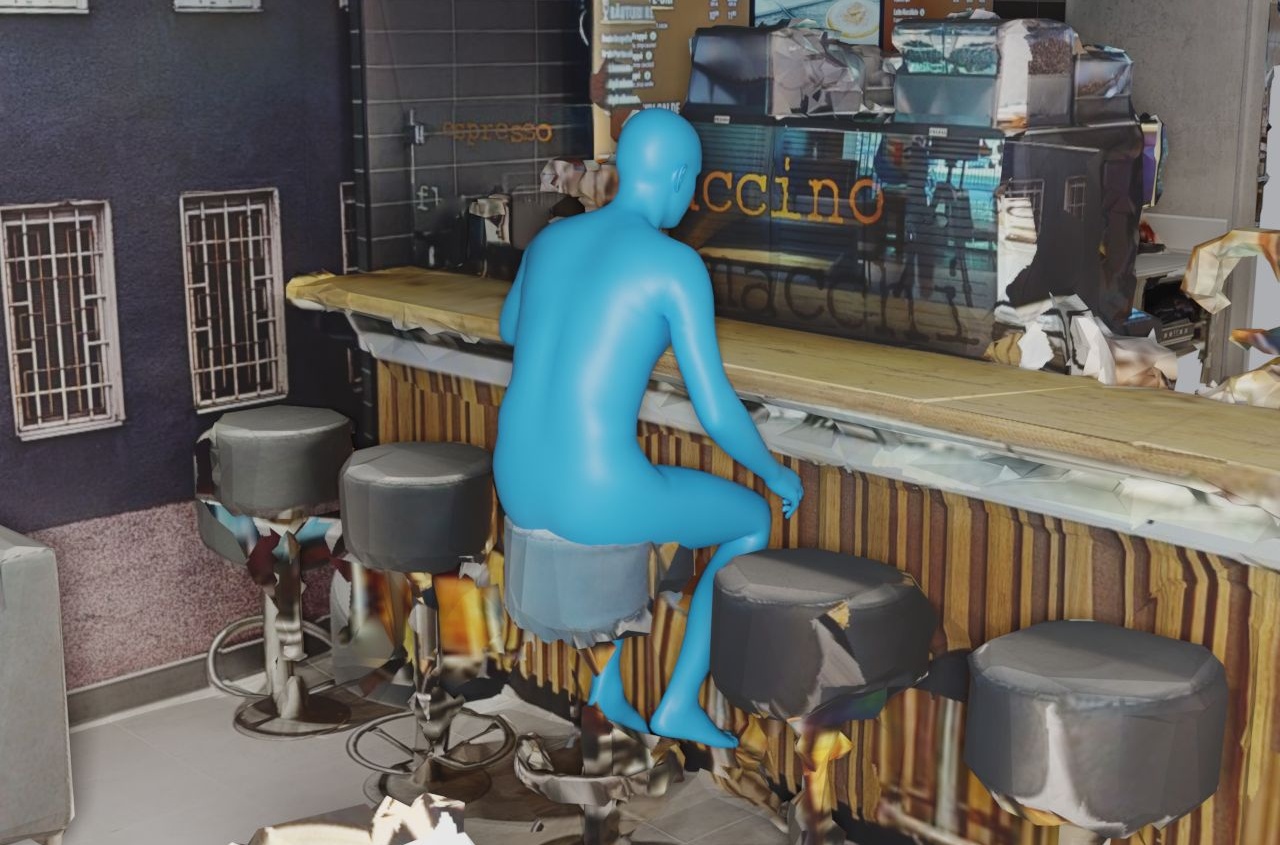} &
    \includegraphics[width=0.18\textwidth, trim=0cm 0cm 0cm 0.369cm, clip]{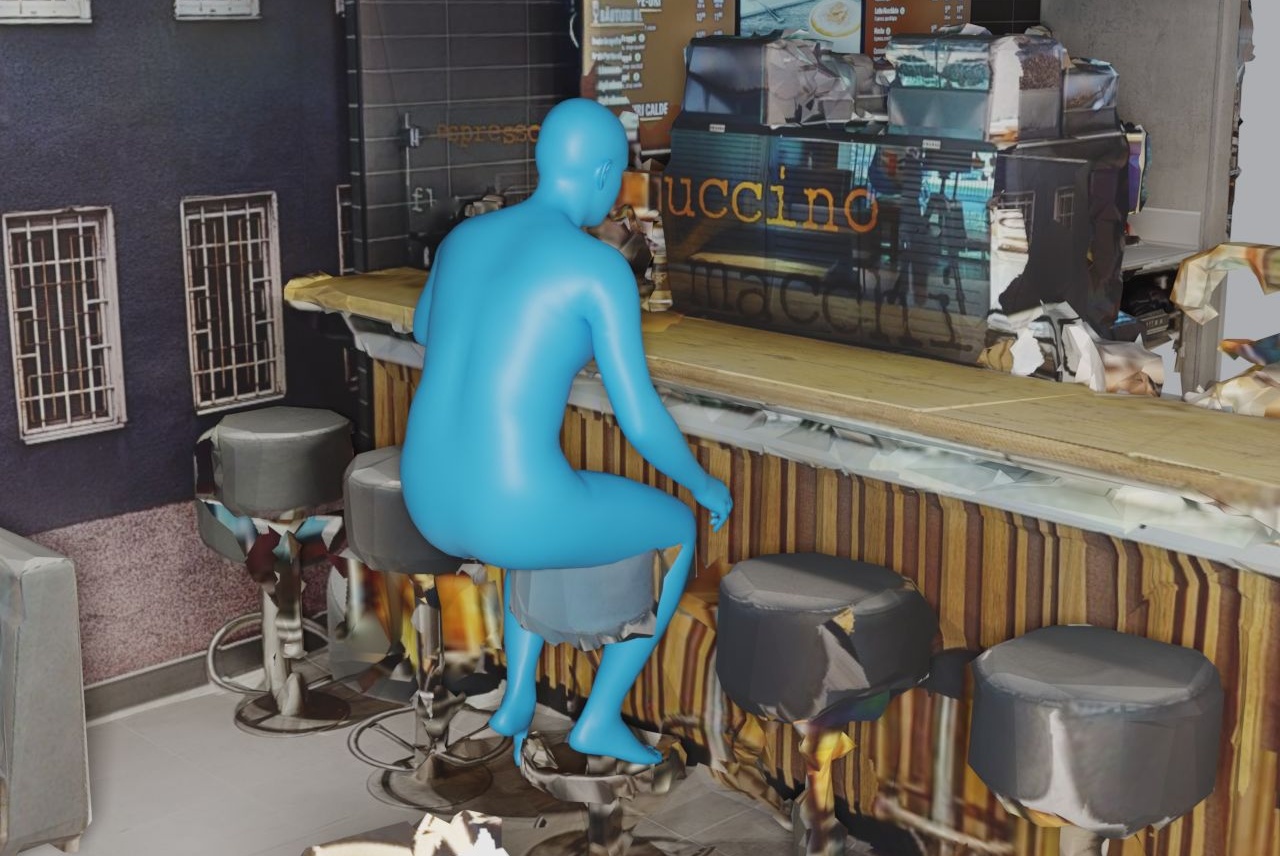}
  \\
    \includegraphics[width=0.18\textwidth]{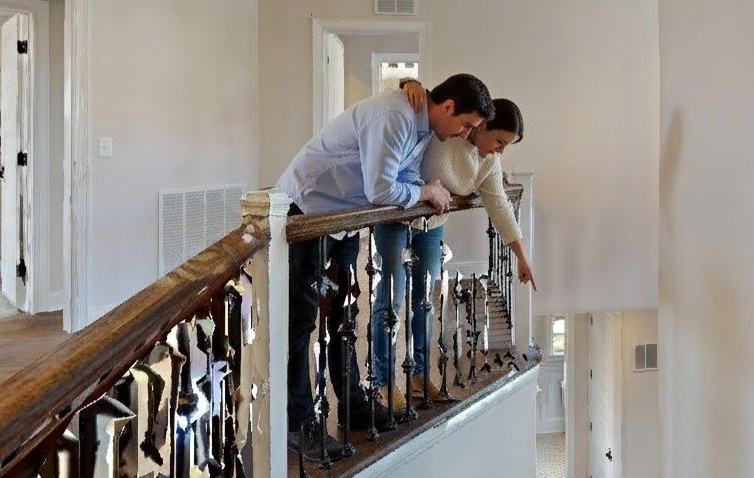} &
    \includegraphics[width=0.18\textwidth, trim=0cm 0cm 0cm 0.4cm, clip]{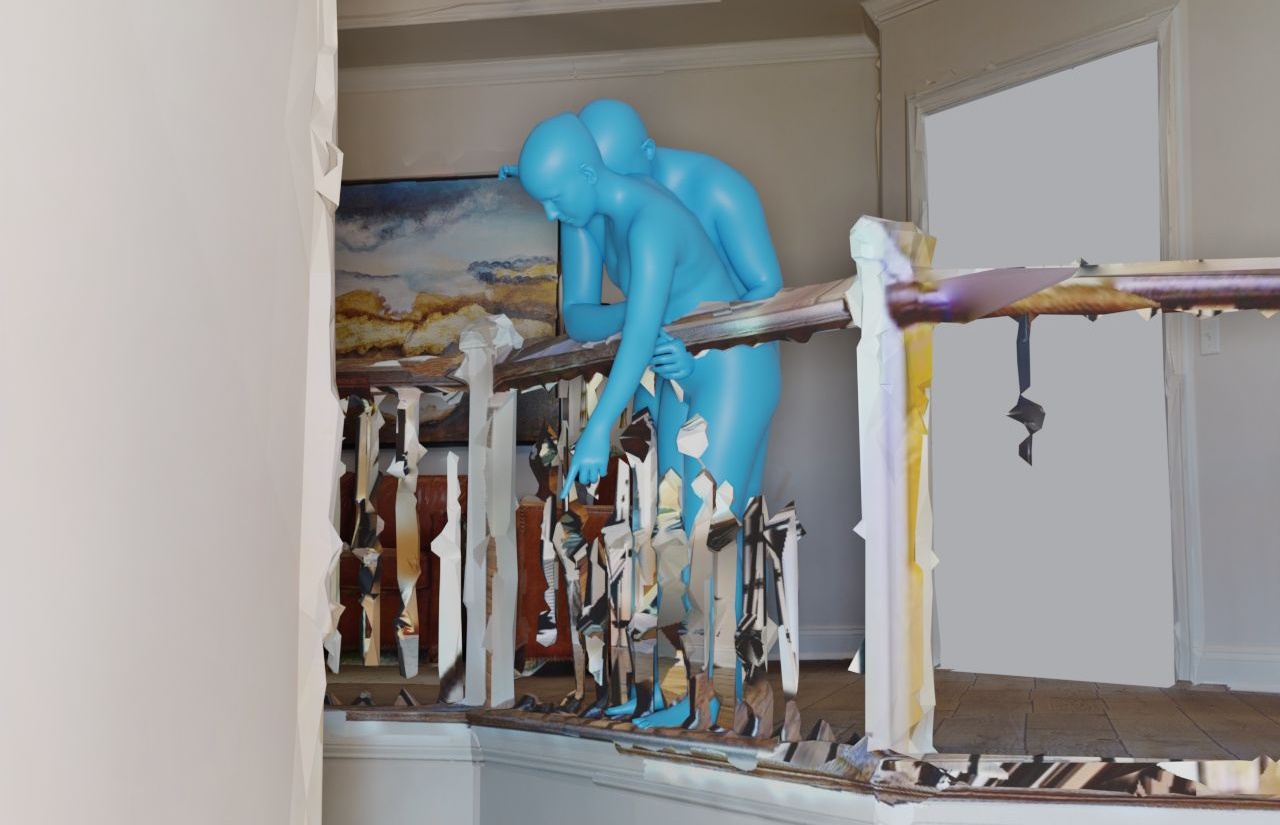} &
    \includegraphics[width=0.18\textwidth]{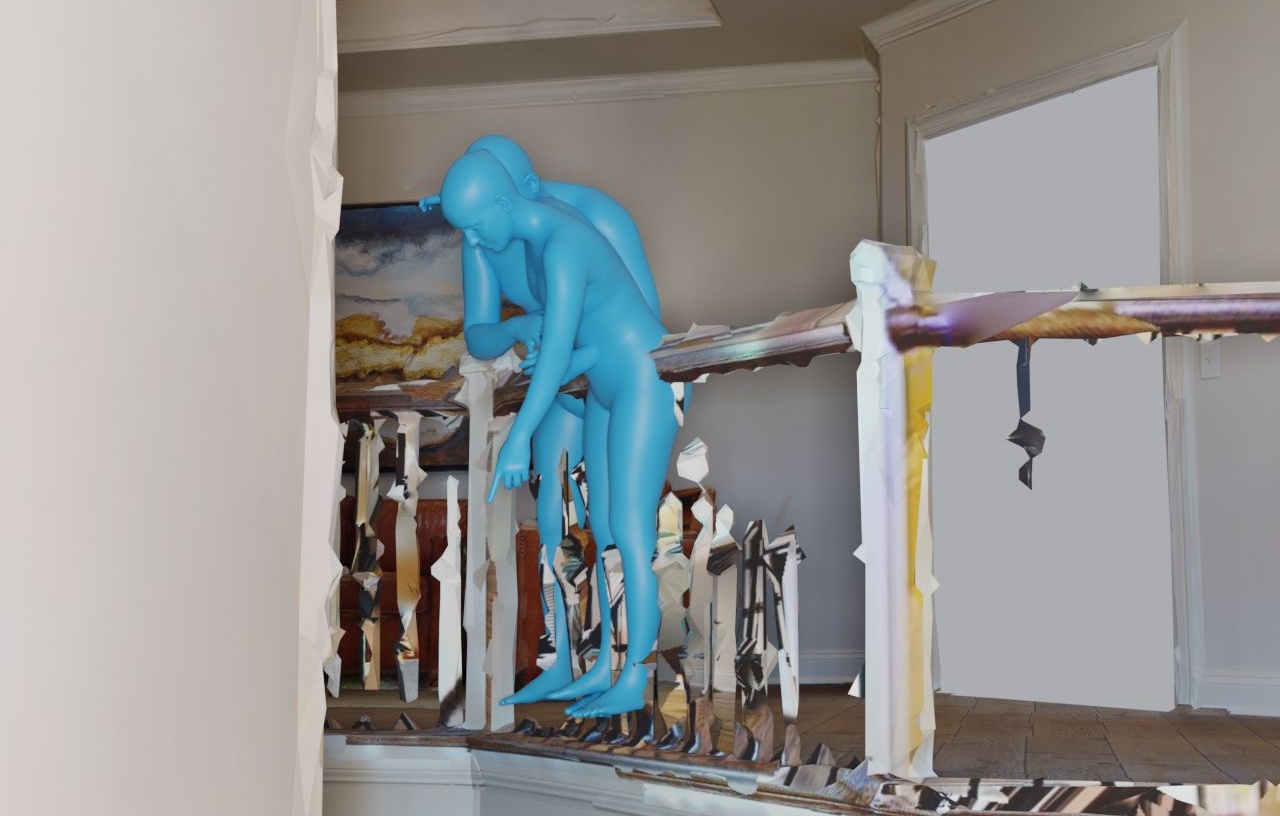}

  \\
    \includegraphics[width=0.18\textwidth]{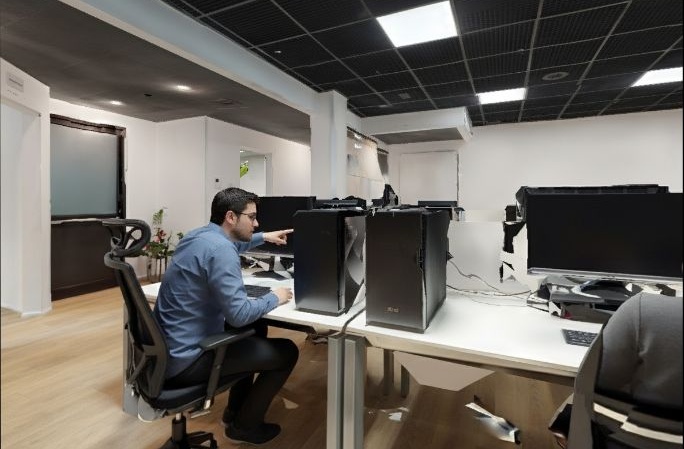} &
    \includegraphics[width=0.18\textwidth]{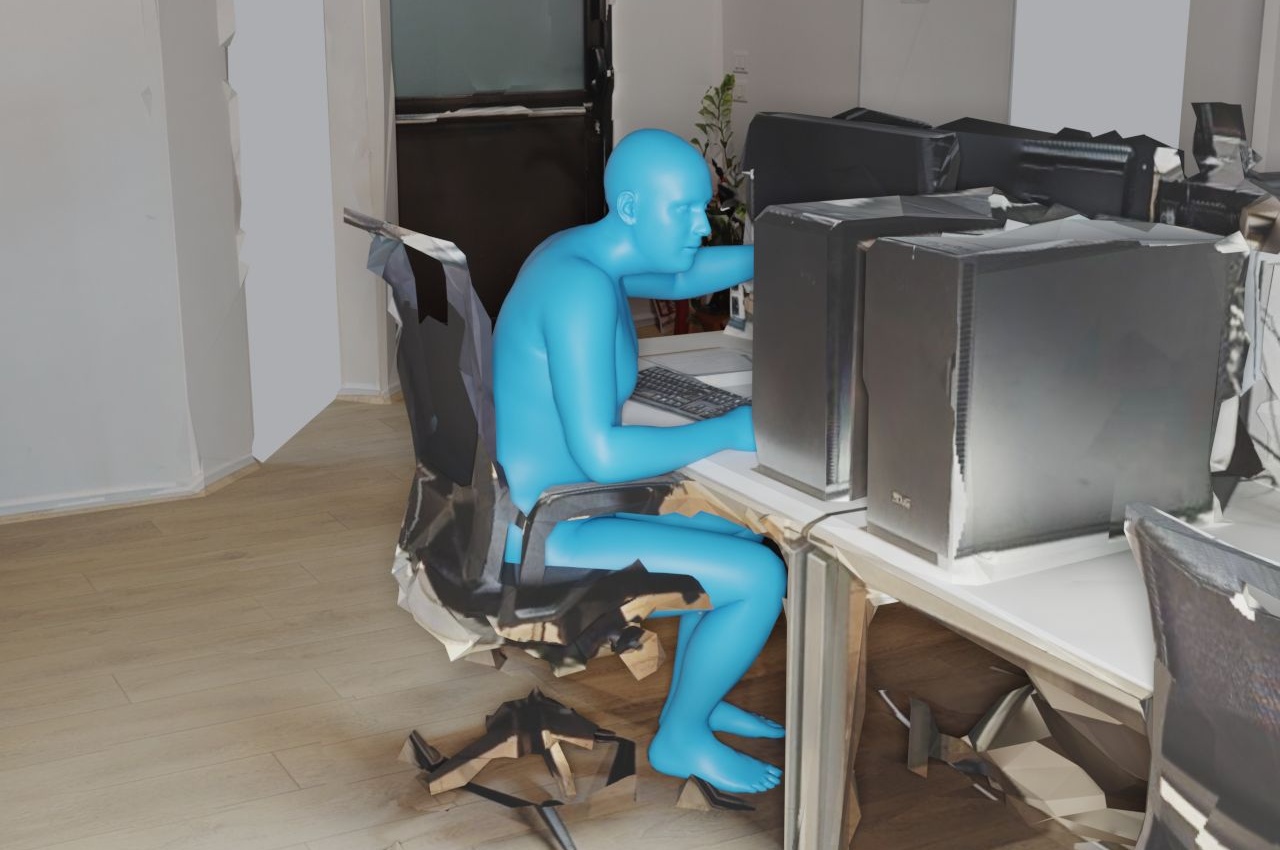} &
    \includegraphics[width=0.18\textwidth]{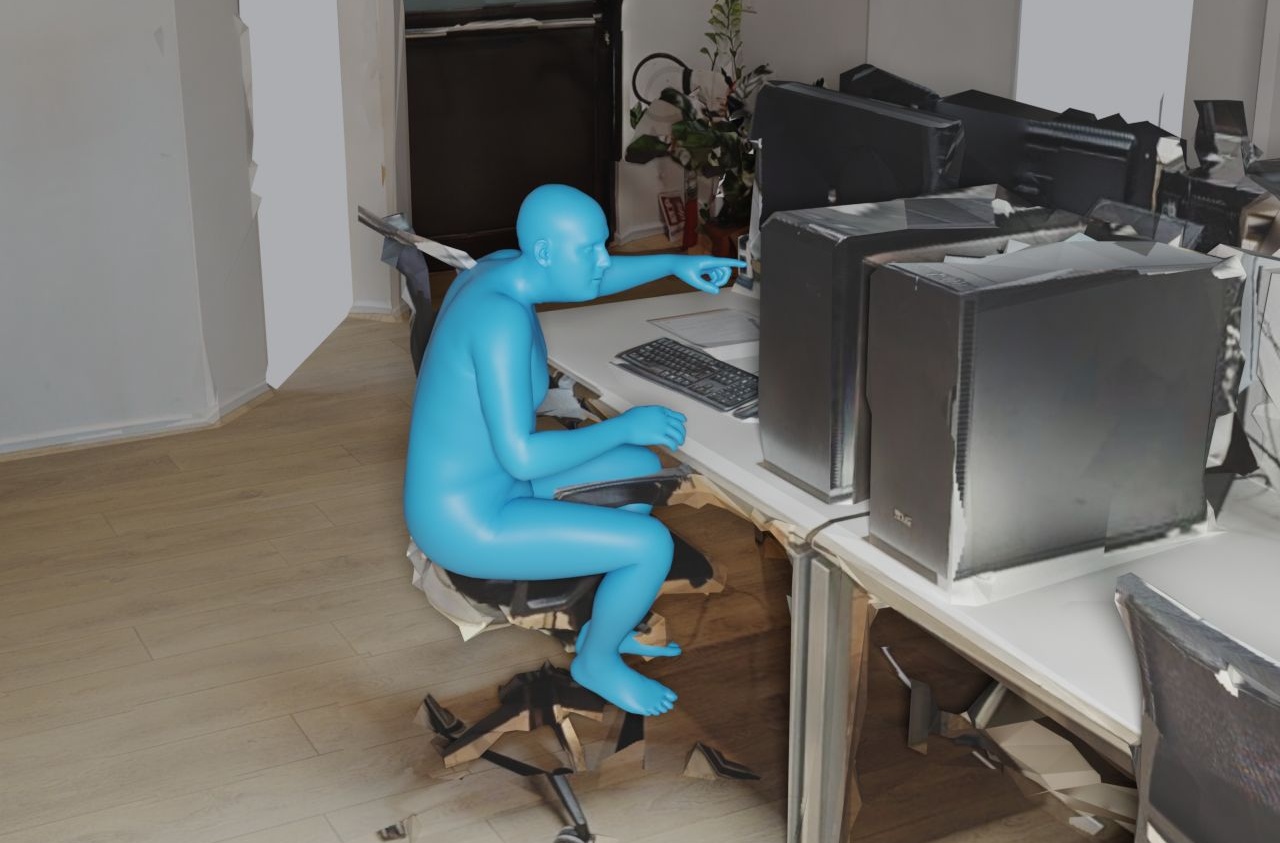}

  \\
  \end{tabular}
  \caption{Ablation of complete-scene versus partial-geometry lifting. Starting from the same edited image, we run lifting either with the complete scene mesh or with partial depth only. When hidden scene geometry is unavailable, the signed-distance field becomes ambiguous in occluded regions, leading to misplaced support and incorrect contacts. Using the full mesh provides consistent geometric constraints and yields cleaner reconstructions.}
  \label{fig:suppl:abl:scene}
\end{figure*}

We ablate the use of the complete 3D scene during the lifting step against a variant that relies only on partial depth maps. We present the results in \Cref{fig:suppl:abl:scene}.
When only partial depth is available, occluded scene surfaces are absent from the geometry used during optimization. This makes the signed-distance field incomplete exactly where it is needed most, so the contact and penetration terms can guide the body toward incorrect placements, as seen for the chairs (rows 1 and 3) and the railing (row 2).
Using the full scene mesh restores those missing constraints and lets the optimizer reason about nearby but unseen support surfaces. This reduces interpenetration and yields more faithful contact placement, especially in challenging cases such as the chair in row 3.
Quantitatively, this is reflected in \Cref{tab:abl:rec}: removing the complete scene (\emph{w/o gt scene}) drops contact F1 from 0.368 to 0.297, with corresponding decreases in precision (0.424$\rightarrow$0.354) and recall (0.393$\rightarrow$0.337), confirming that the missing geometric constraints translate directly into worse contact placement.

\section{Explicit Scale Parameter}
\label{suppl:sec:scale}

\begin{figure*}[t]
  \centering
  \setlength{\tabcolsep}{1pt}
  \renewcommand{\arraystretch}{0.8}
  \begin{tabular}{c c c}
    \textbf{Input} & \textbf{w/ scale} & \textbf{w/o scale} \\
    \includegraphics[width=0.18\textwidth,trim=0cm 0cm 0cm 0.05cm,clip]{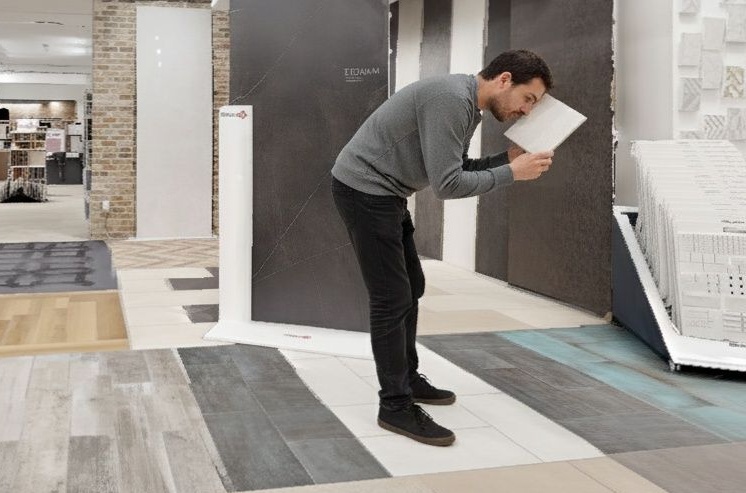} &
    \includegraphics[width=0.18\textwidth]{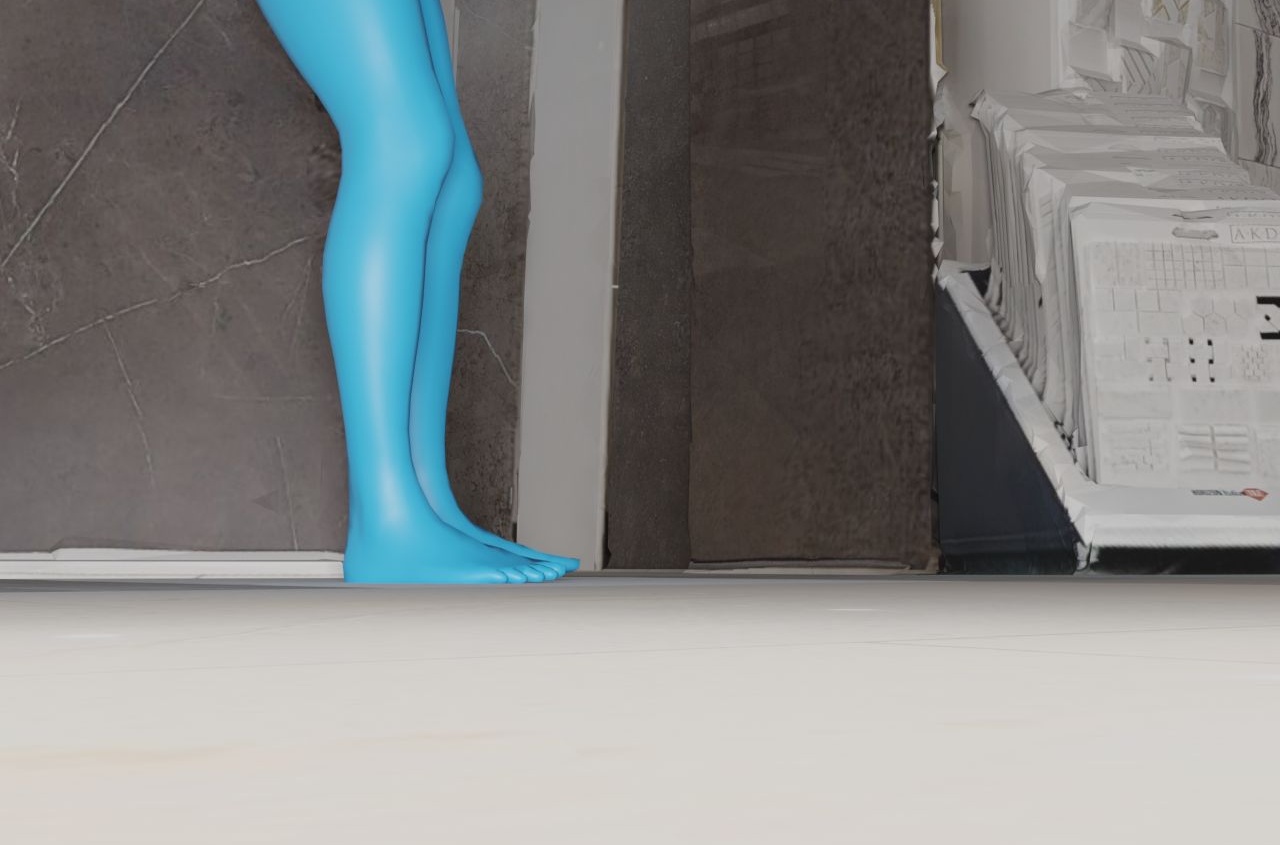} &
    \includegraphics[width=0.18\textwidth]{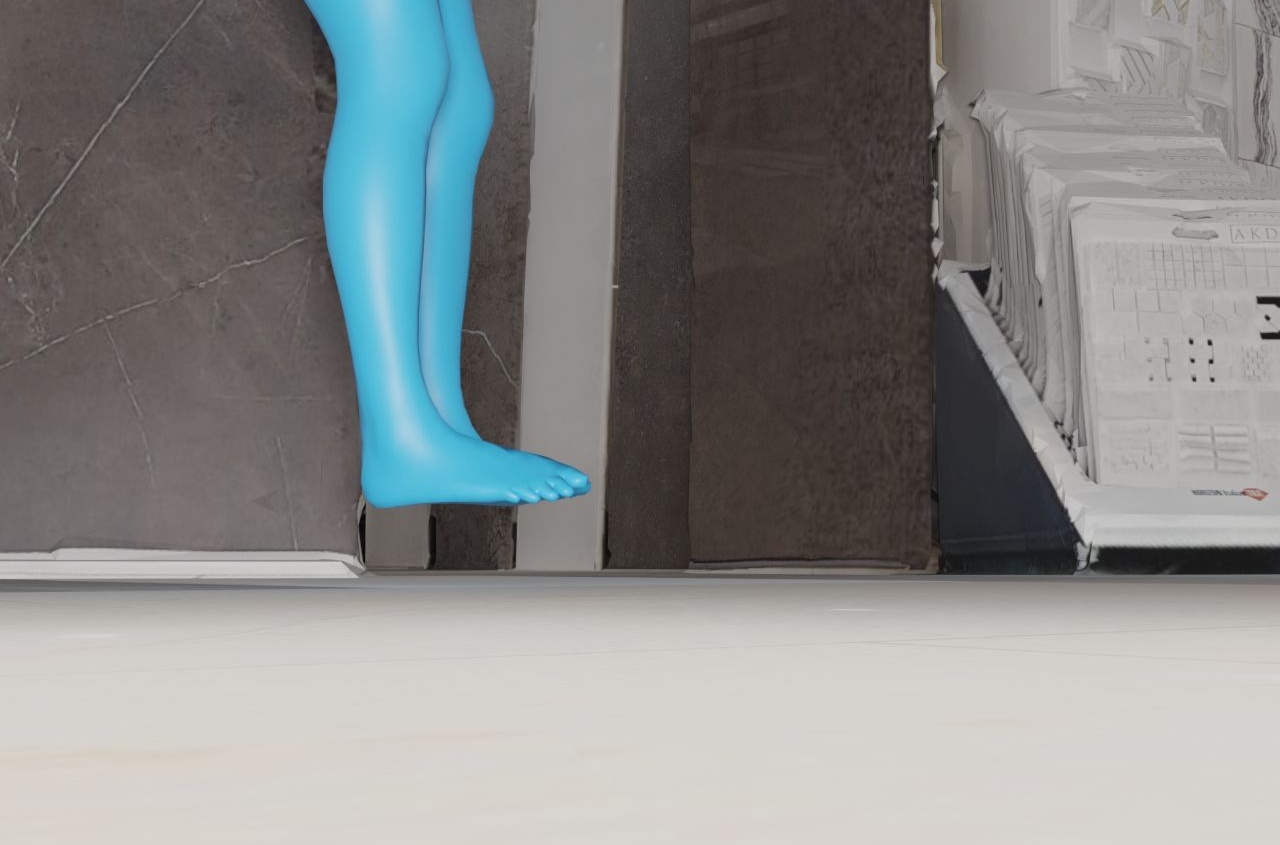}
  \\
    \includegraphics[width=0.18\textwidth]{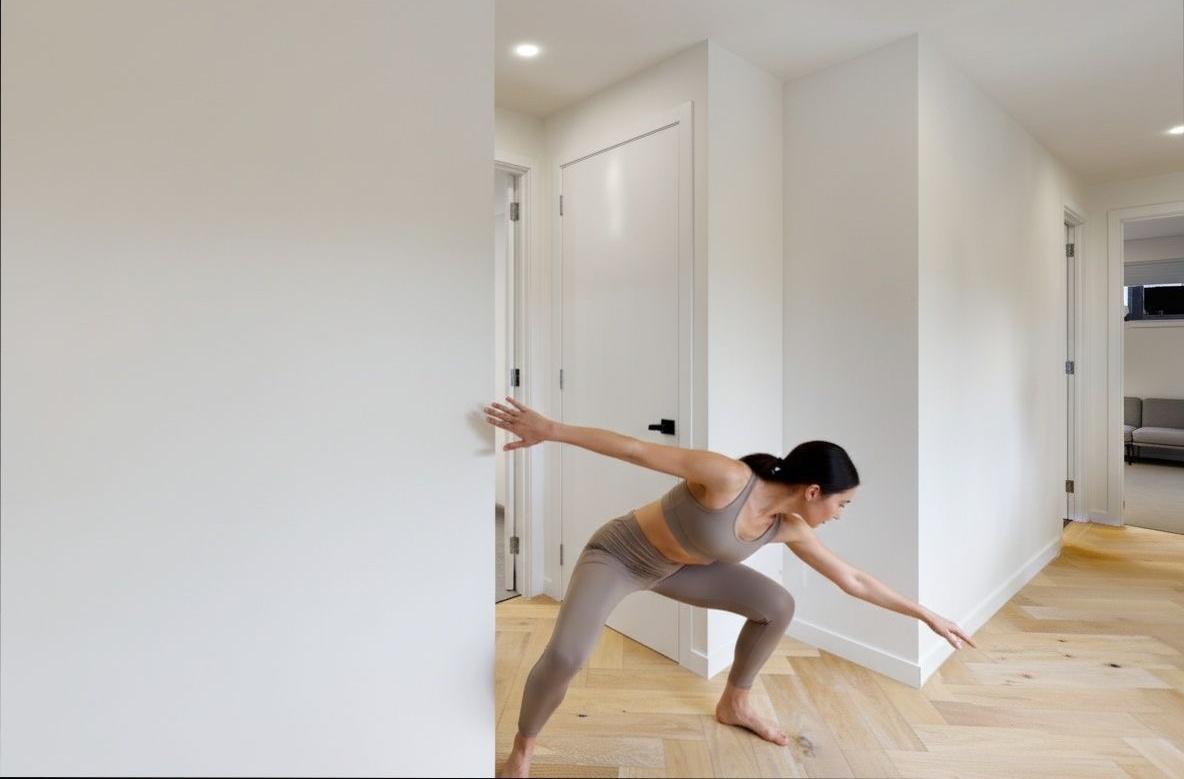} &
    \includegraphics[width=0.18\textwidth,trim=0cm 0cm 0cm 0.05cm,clip]{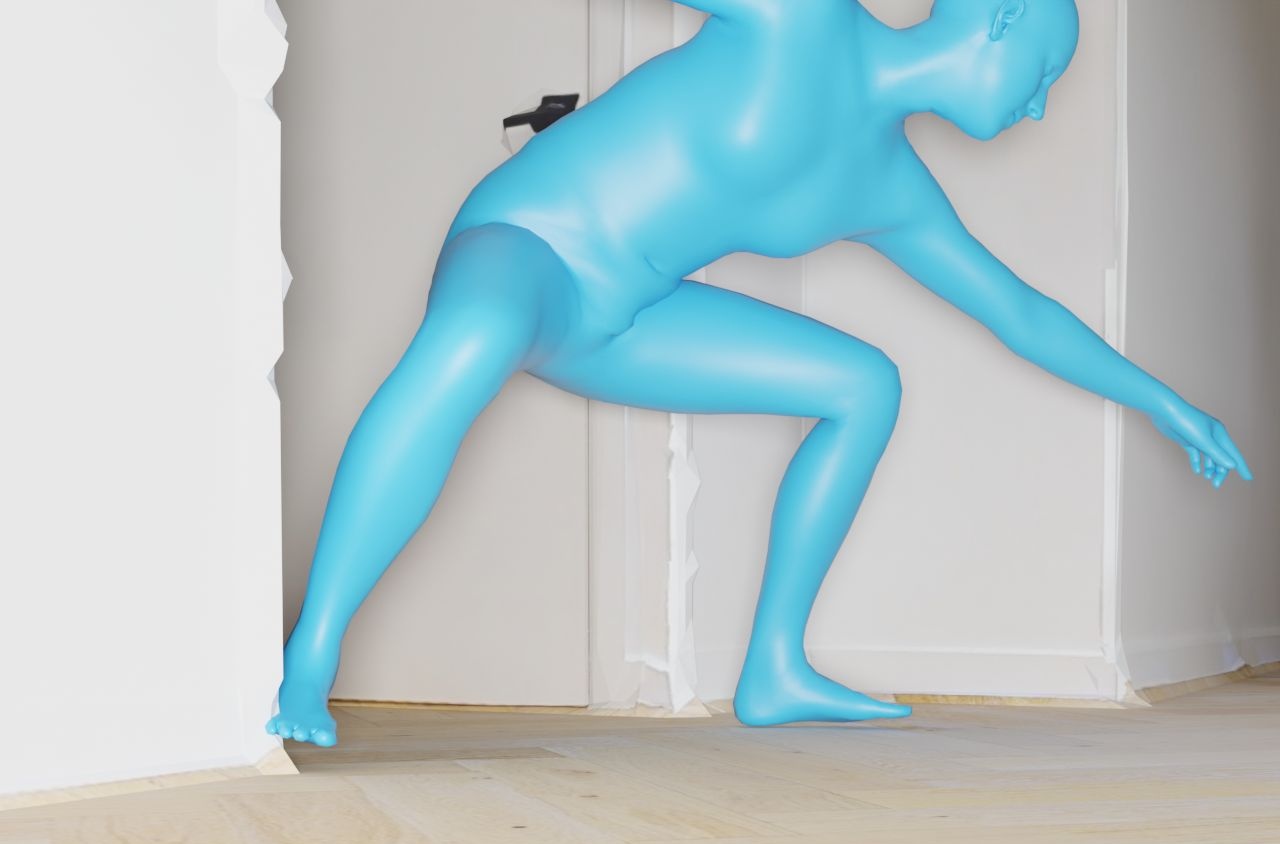} &
    \includegraphics[width=0.18\textwidth,trim=0cm 0cm 0cm 0.05cm,clip]{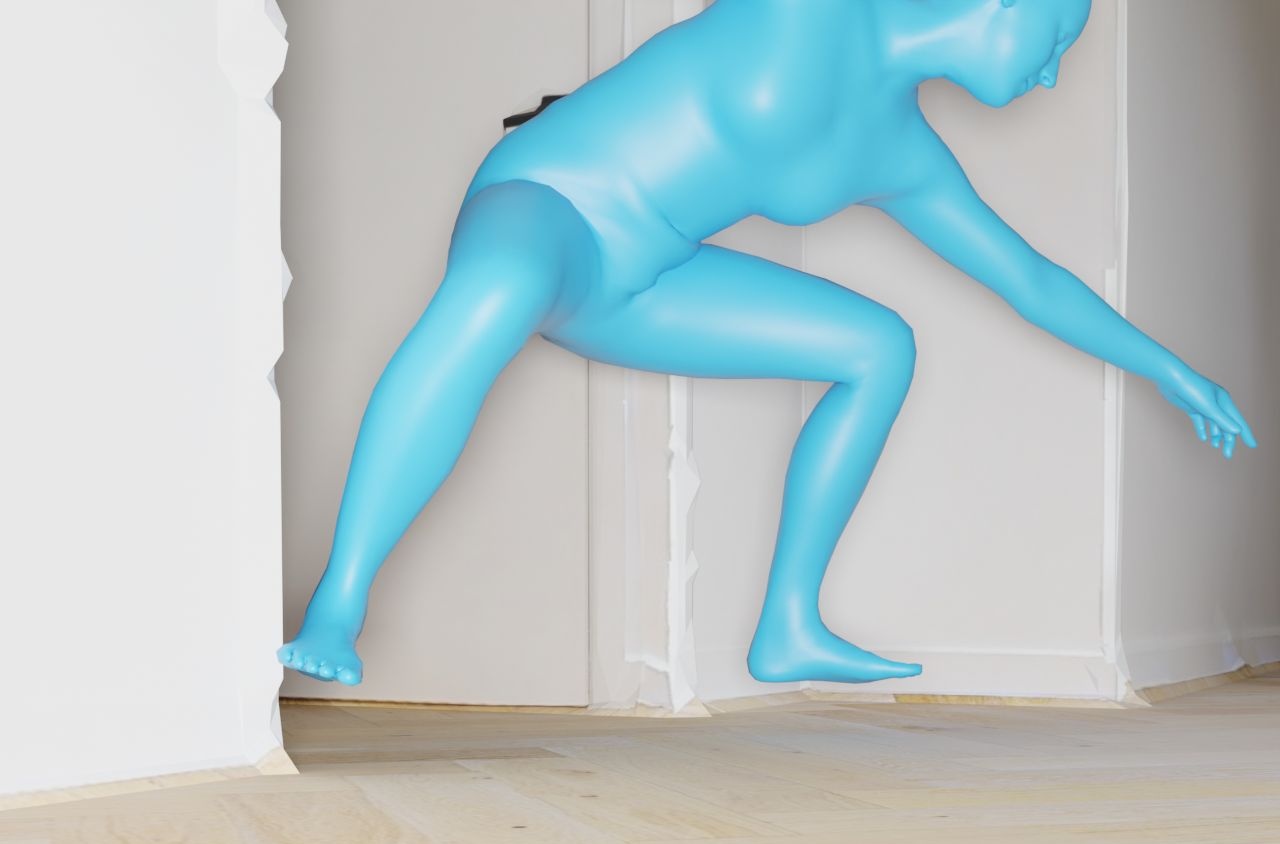}

  \\
    \includegraphics[width=0.18\textwidth]{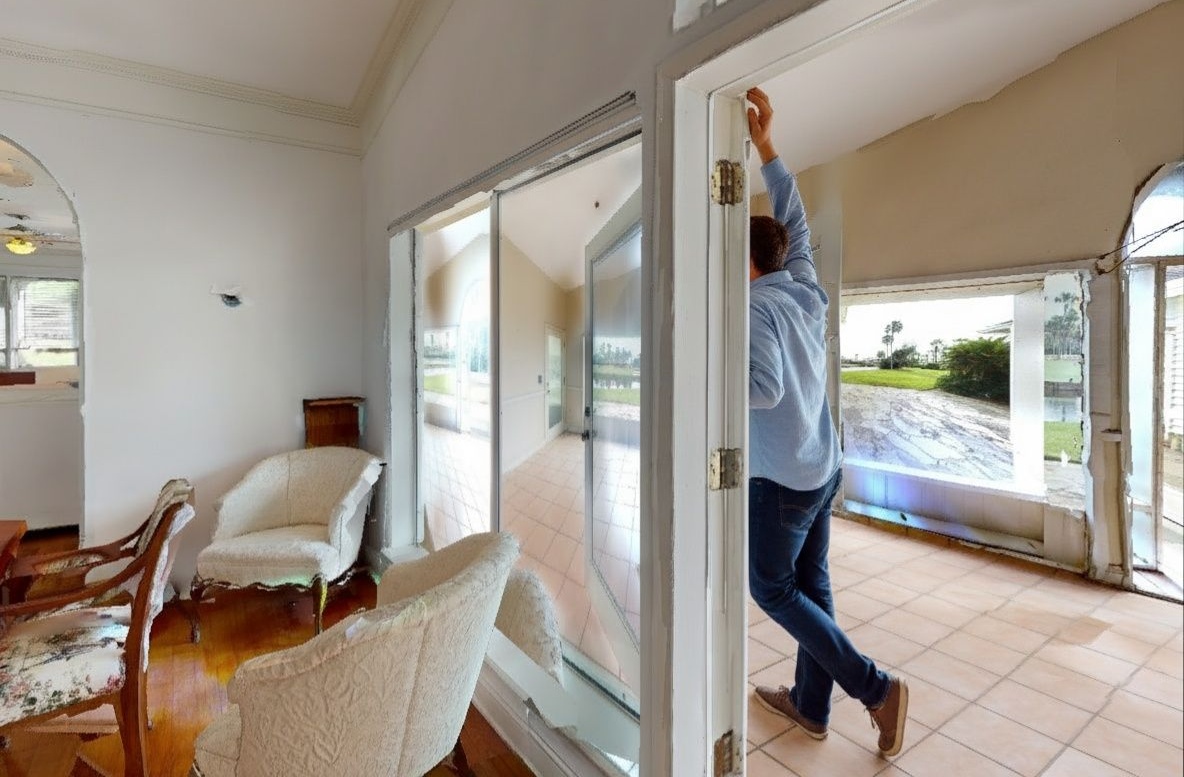} &
    \includegraphics[width=0.18\textwidth,trim=0cm 0cm 0cm 0.17cm,clip]{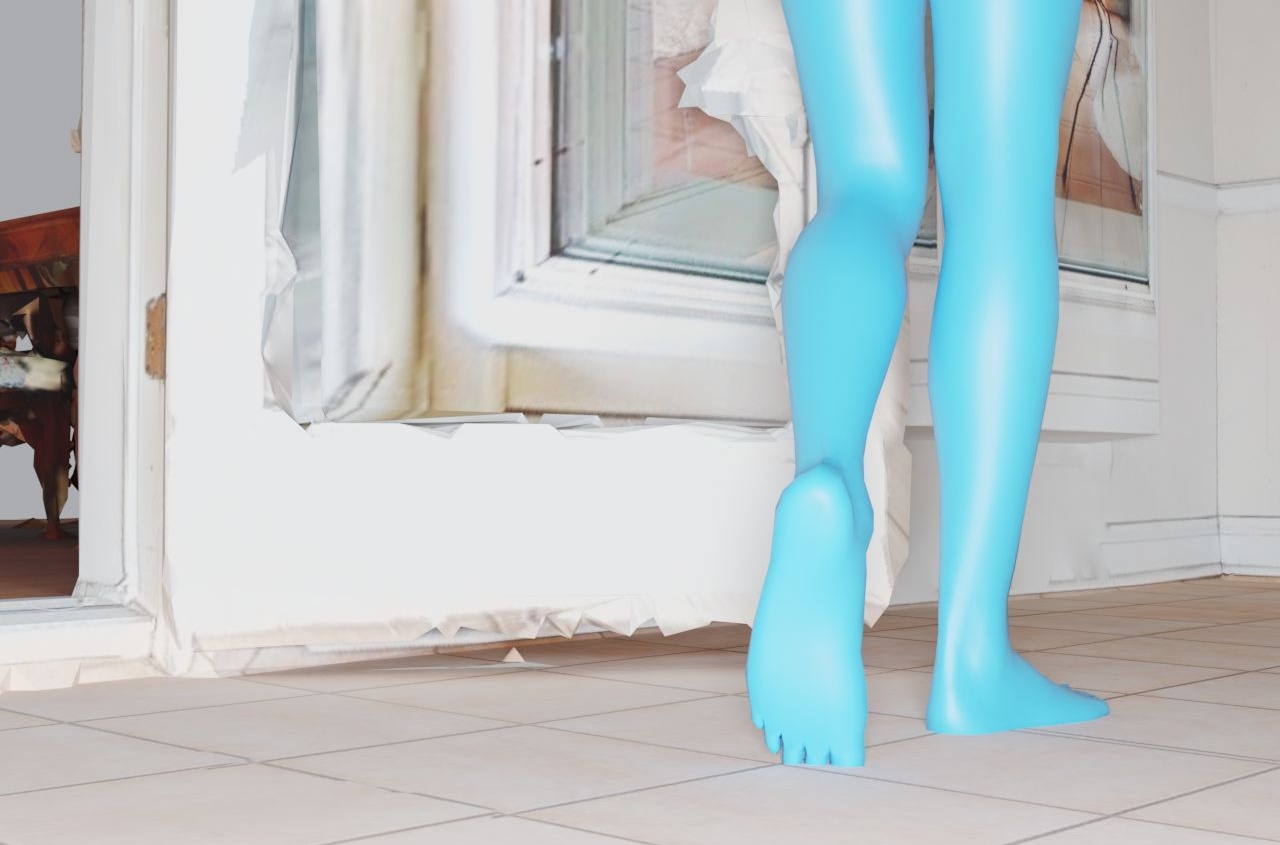} &
    \includegraphics[width=0.18\textwidth,trim=0cm 0cm 0cm 0.17cm,clip]{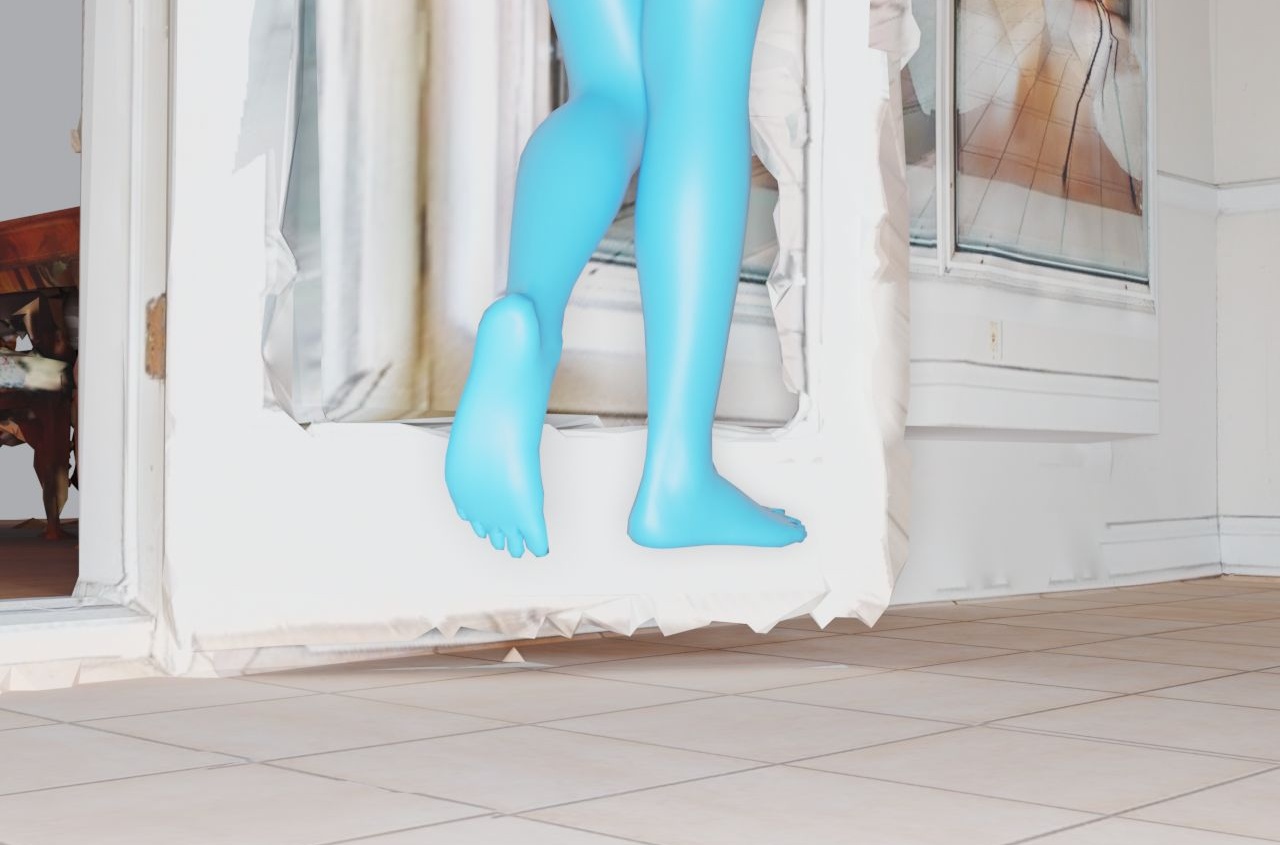}
\\
    \includegraphics[width=0.18\textwidth,trim=0cm 0cm 0cm 0.1cm,clip]{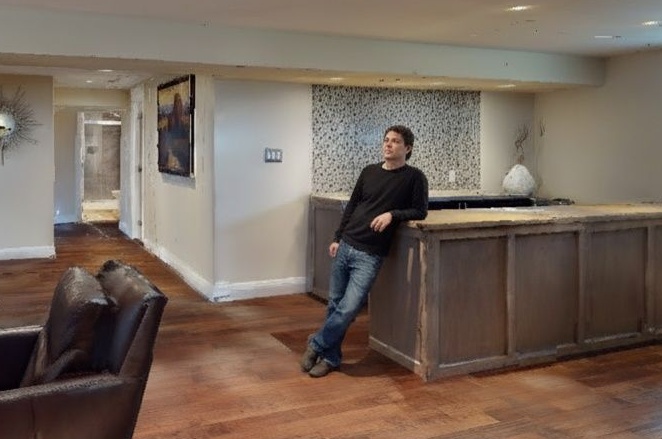} &
    \includegraphics[width=0.18\textwidth]{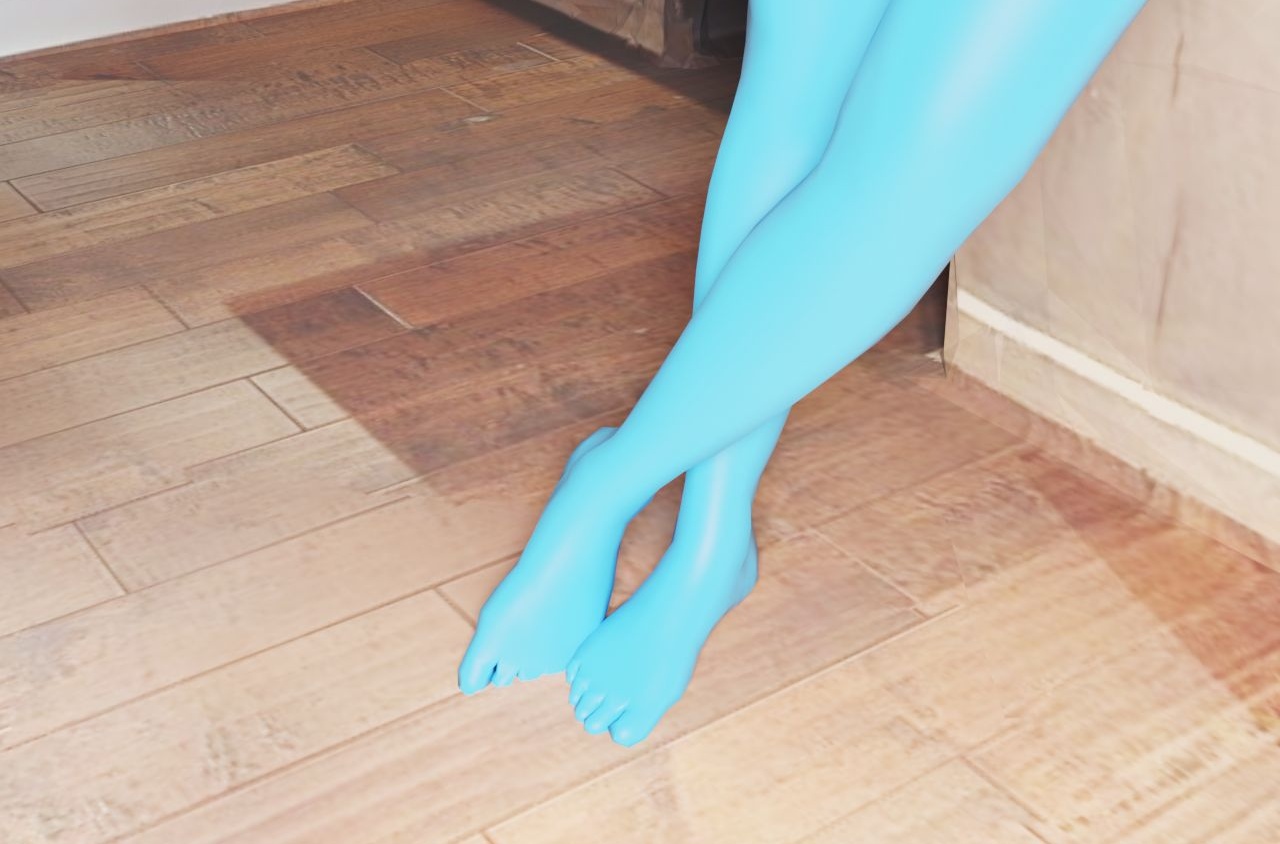} &
    \includegraphics[width=0.18\textwidth]{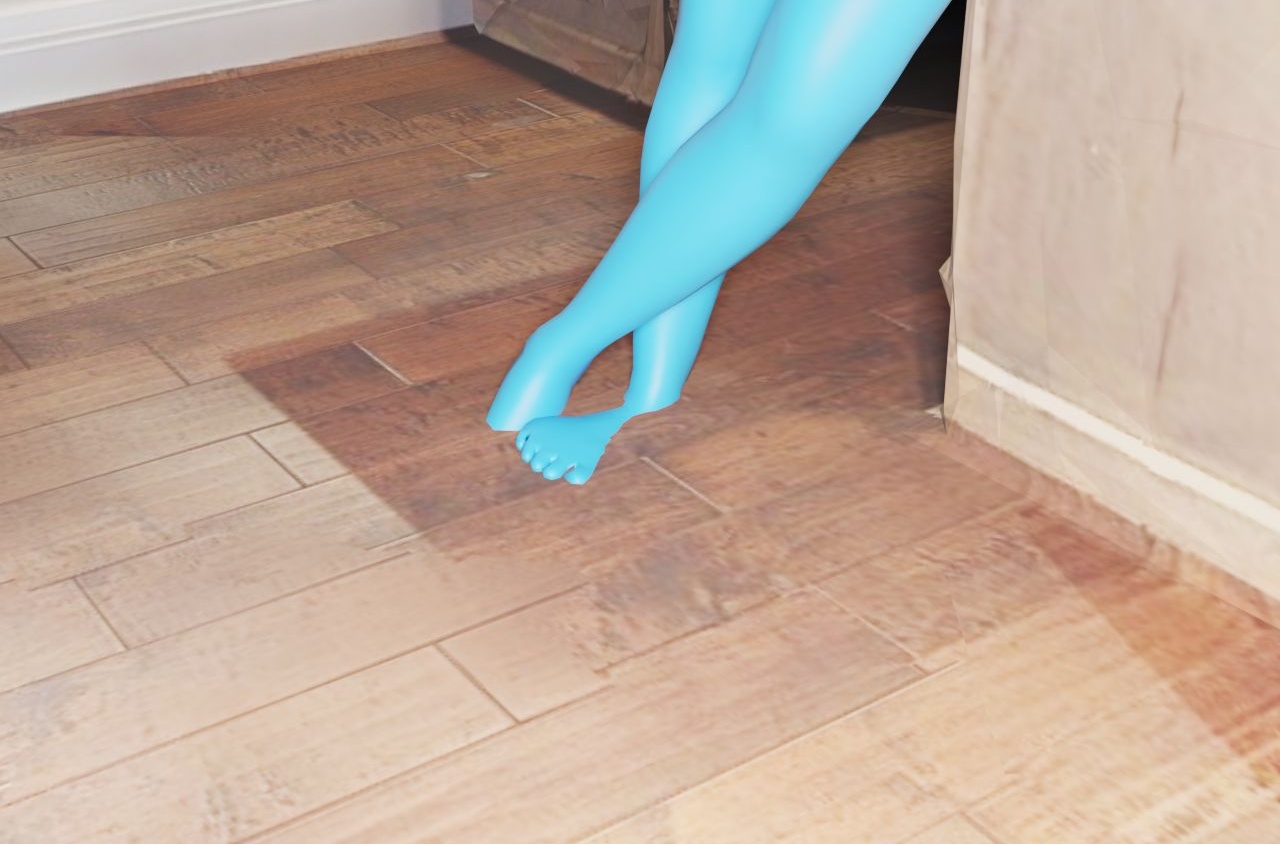}

  \\
  \end{tabular}
  \caption{Ablation of the scale variable used during lifting. We apply the same procedure to the edited image, once optimizing the scale term $s$ and once keeping scale fixed. Without a dedicated scale variable, metric mismatch must be absorbed indirectly through coupled changes in body shape and position while still matching the image silhouette. Optimizing $s$ provides a direct correction for that mismatch and improves grounding, support, and contact with the scene.}
  \label{fig:suppl:abl:scale}
\end{figure*}
We ablate the explicit scale parameter introduced in the lifting stage and compare it to a variant that keeps scale fixed. We present the results in \Cref{fig:suppl:abl:scale}.
Without a dedicated scale variable, any mismatch between the pose estimator's scale and the scene must be compensated by the remaining optimization variables. In practice this is awkward: the SMPL-X shape coefficients $\betas$ affect the body nonlinearly and change its outline, while translation has to move in tandem to preserve reprojection agreement. That coupling makes the correct metric size hard to recover.
Allowing the optimizer to adjust $s$ directly separates scene-level sizing from body shape, making the fit more stable. Qualitatively, this leads to better floor support, more accurate placement relative to nearby surfaces, and fewer floating or interpenetrating bodies.
This effect is also visible in \Cref{tab:abl:rec}: removing the scale parameter (\emph{w/o human scale}) drops contact F1 from 0.368 to 0.334 (precision 0.424$\rightarrow$0.396, recall 0.393$\rightarrow$0.359), showing that the indirect compensation through $\betas$ and translation is unable to recover the same quality of contact placement.

\section{Ablation of Interaction and Scene Diversity}\label{sec:what-makes-useful}
\begin{table}[t]
  \centering
    \caption{\textbf{Ablation of interaction and scene diversity.} GRAFT~\citep{graft} trained on size-matched subsets of \ourdata{} ($\sim$22k images each; scene-view subsets with 2,110, 4,219, and 8,438 views) that differ in a single factor, evaluated on PROX Quant.}
  \setlength{\tabcolsep}{7pt}
  \renewcommand{\arraystretch}{1.1}
  \begin{tabular}{lcccccc}
    \toprule
    & \multicolumn{2}{c}{Interactions} & \multicolumn{3}{c}{Scene views} & Captured \\
    \cmidrule(lr){2-3} \cmidrule(lr){4-6} \cmidrule(lr){7-7}
    & feet-only & all & 25\% & 50\% & 100\% & PROX \\
    \midrule
    PROX F1 $\uparrow$ & 0.468 & \textbf{0.577} & 0.546 & 0.551 & \textbf{0.578} & 0.492 \\
    \bottomrule
  \end{tabular}

  \label{tab:data-ablation}
\end{table}

To understand where the downstream gains come from, we train GRAFT on subsets of \ourdata{} that are matched in size ($\sim$22k images each) and differ in a single factor: interaction diversity (feet-only contacts vs.\ all interactions) or scene diversity (25\%, 50\%, or 100\% of all views).
As shown in \Cref{tab:data-ablation}, interaction diversity is the dominant factor: at equal size, F1 on PROX rises from 0.468 to 0.577. Scene diversity contributes a smaller but consistent improvement (0.546 $\rightarrow$ 0.578 from 25\% to 100\%).
Note that even our smallest subset spans 2,110 scene views, an order of magnitude beyond captured datasets, so these gains are measured in a regime of scene coverage that capture cannot reach.
Notably, the feet-only subset underperforms training on captured PROX data (0.492), while the diverse subset clearly surpasses it, despite containing only about a quarter as many training images ($\sim$22k vs.\ $\sim$88k). This indicates that the benefit of \ourdata{} stems mainly from the breadth of interactions it contains. The same trend holds against TRUMANS (\Cref{tab:damon-contact}), which contains far more frames than \ourdata{} but narrower interactions. Compared with \Cref{table:hsr}, GRAFT trained on BEDLAM ($0.396$) or on our feet-only subset ($0.468$) performs worse than GRAFT trained on PROX ($0.492$), whereas training on our subset with all interactions exceeds it ($0.577$).

\section{Physical Stability of \ourdata{}}
\label{suppl:sec:stability}
\begin{table}[h]
  \centering
\caption{\textbf{Physical stability of dataset poses.} Percentage of poses whose center of mass projects inside the base of support, following IPMAN~\citep{tripathi2023ipman}.}
  \setlength{\tabcolsep}{8pt}
  \renewcommand{\arraystretch}{1.1}
  \begin{tabular}{lc}
    \toprule
    Dataset & Stable poses (\%) $\uparrow$ \\
    \midrule
    PROX-D & 65.40 \\
    TRUMANS & 75.88 \\
    \ourdata{} (Ours) & \textbf{83.21} \\
    \bottomrule
  \end{tabular}

  \label{tab:suppl:stability}
\end{table}

Our lifting stage (\Cref{sec:lifting}) does not explicitly enforce physical balance, so we verify that the resulting poses are physically stable.
Following IPMAN~\citep{tripathi2023ipman}, we consider a pose stable if the projection of its center of mass onto the ground plane lies within its base of support (BoS). The BoS is the convex hull of the body's contact points, projected onto the ground plane, where contact points are obtained with a simple distance threshold to the scene geometry.
As shown in \Cref{tab:suppl:stability}, 83.21\% of \ourdata{} poses are stable, compared to 75.88\% for TRUMANS and 65.40\% for PROX-D.

\section{Additional Qualitative Results}
\label{suppl:sec:data}
\begin{figure*}[t]
  \centering
  \includegraphics[width=\linewidth]{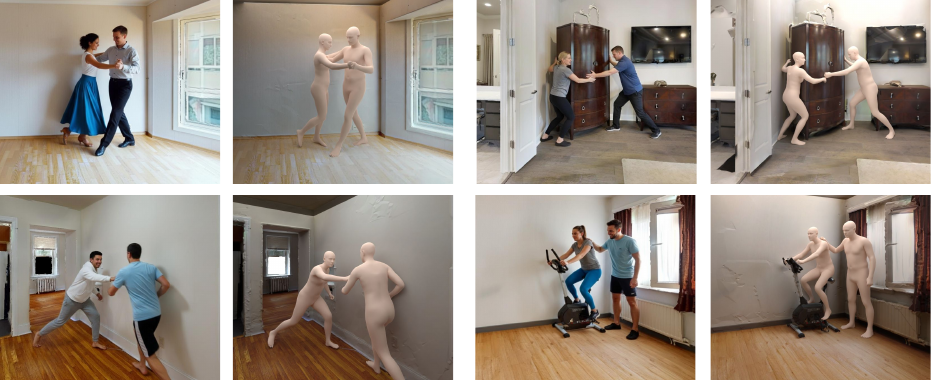}
  \caption{\ours{} can naturally generate interactions involving multiple people, which was not supported by prior work.
 }
  \label{fig:multi:person}
\end{figure*}

\newcommand{\suppgenzi}[1]{\includegraphics[width=0.17\textwidth]{assets/suppl_ours_vs_baselines/#1__human_01_Genzi_suppl.png}}
\newcommand{\suppposa}[1]{\includegraphics[width=0.17\textwidth]{assets/suppl_ours_vs_baselines/#1__human_03_POSA_suppl.png}}
\newcommand{\suppours}[1]{\includegraphics[width=0.17\textwidth]{assets/suppl_ours_vs_baselines/#1__human_02_Ours_suppl.png}}

\newcommand{\dcell}[1]{\parbox[t]{0.17\textwidth}{\centering #1}}
\newcommand{\descrow}[5]{%
  & \dcell{#1} & \dcell{#2} & \dcell{#3} & \dcell{#4} & \dcell{#5} \\
}

\newcommand{\descriptionsA}{}
\renewcommand{\descriptionsA}{\descrow{Taking something \\out of the freezer}{Reaching for a book\\ in the top row}{Touching the door knob}{Opening the door}{Reaching for the plant}}

\newcommand{\descriptionsB}{}
\renewcommand{\descriptionsB}{\descrow{Sitting on the stool \\in the closet}{Taking something \\ out of the fireplace}{Sitting on the sofa \\and leaning back}{Relaxing on a bed }{Sitting casually \\ on a platform}}

\begin{figure*}[t]
  \centering
  \setlength{\tabcolsep}{3pt}
  \renewcommand{\arraystretch}{0.95}
  \begin{tabular}{cccccc}
    \rotatebox{90}{\hspace{1.2cm}GenZI} &
    \suppgenzi{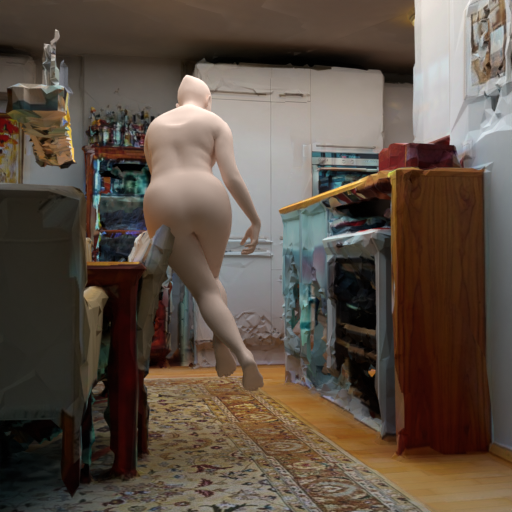} &
    \suppgenzi{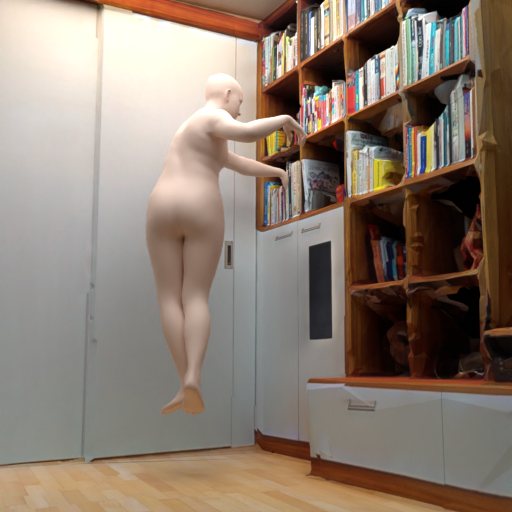} &
    \suppgenzi{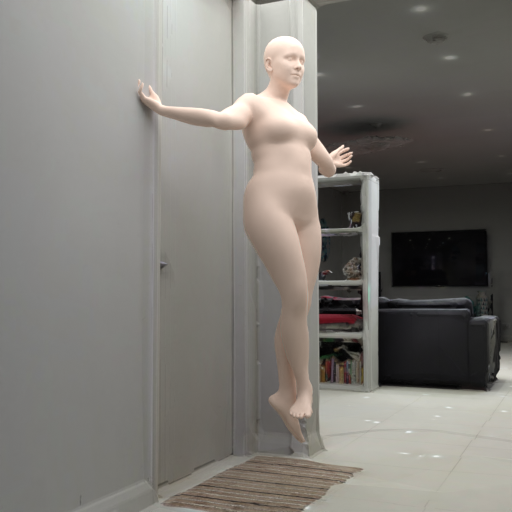} &
    \suppgenzi{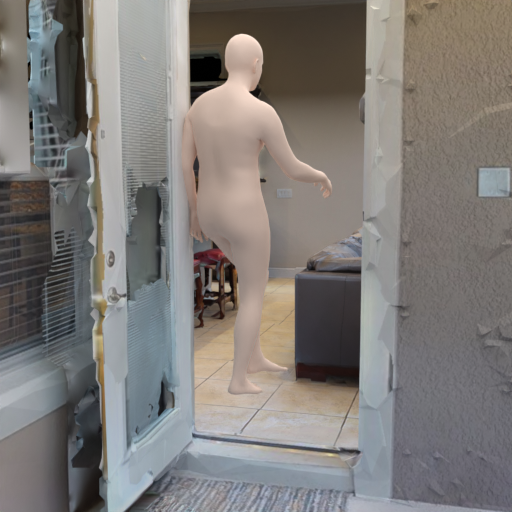} &
    \suppgenzi{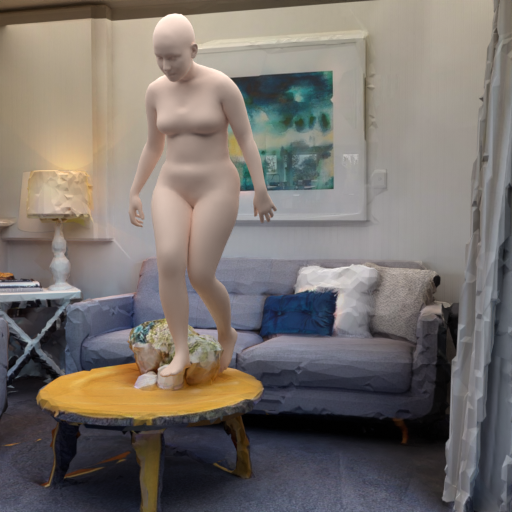} \\

    \rotatebox{90}{\hspace{1.15cm}POSA++} &
    \suppposa{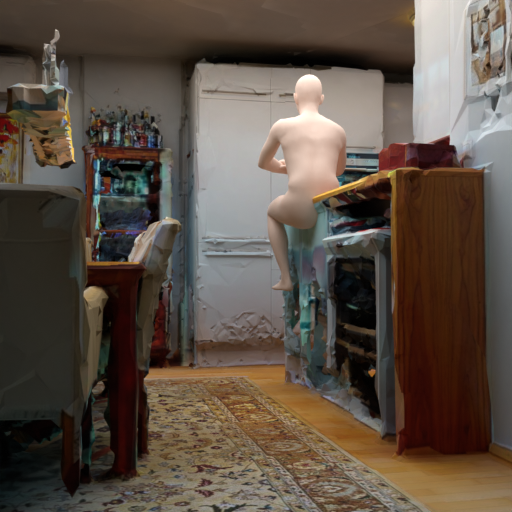} &
    \suppposa{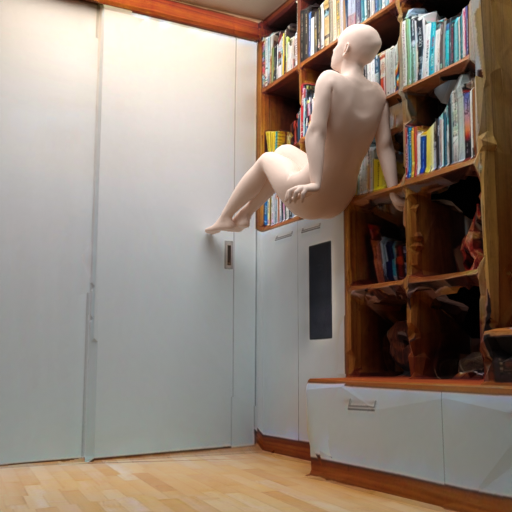} &
    \suppposa{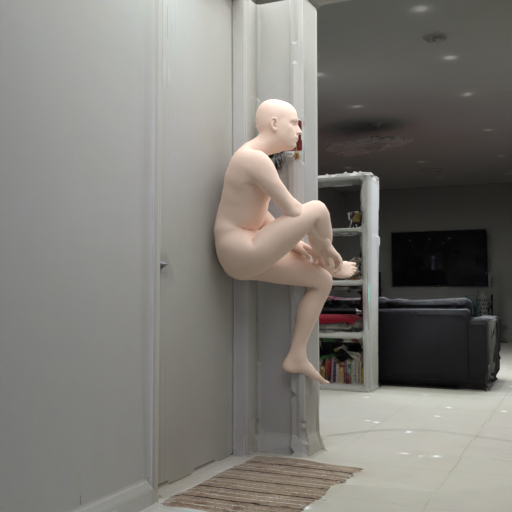} &
    \suppposa{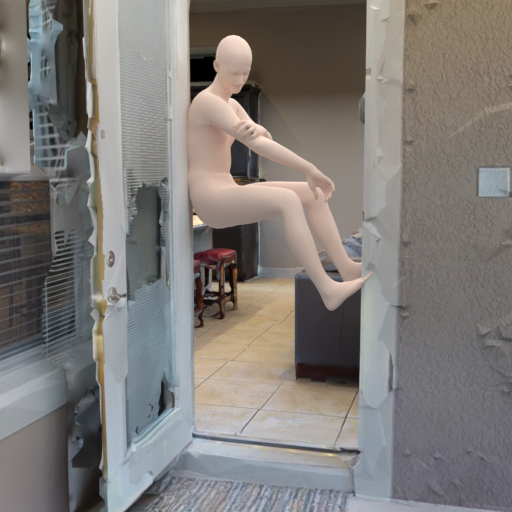} &
    \suppposa{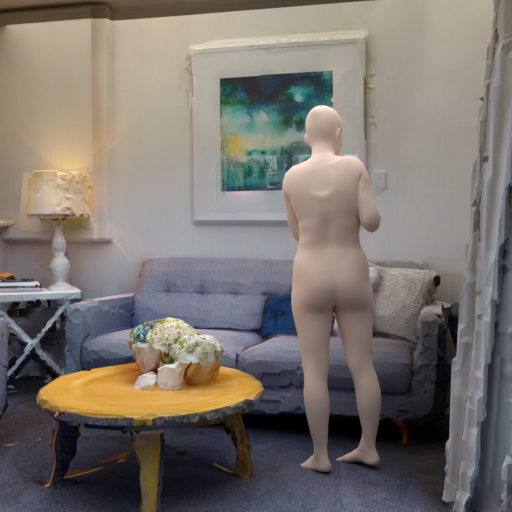} \\

    \rotatebox{90}{\hspace{0.7cm}\ours{} (Ours)} &
    \suppours{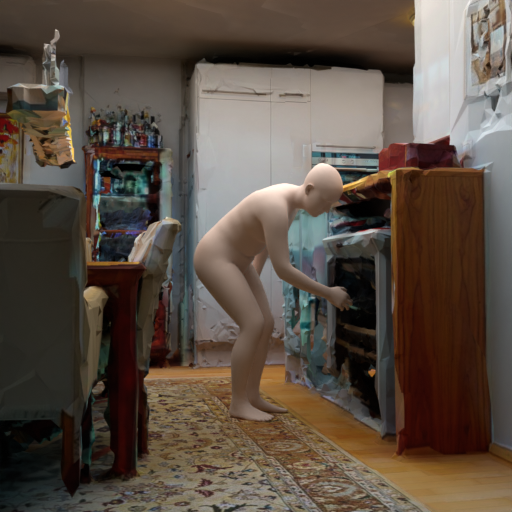} &
    \suppours{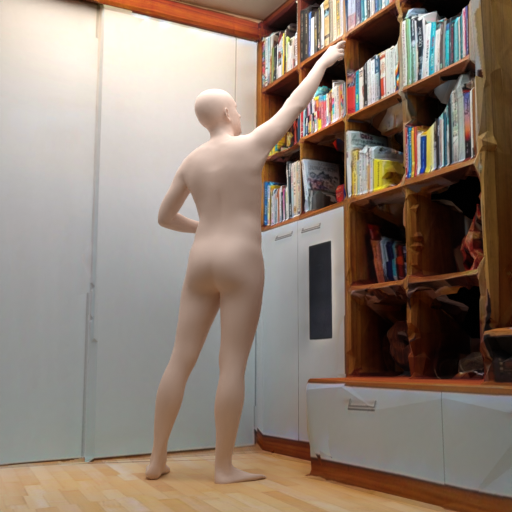} &
    \suppours{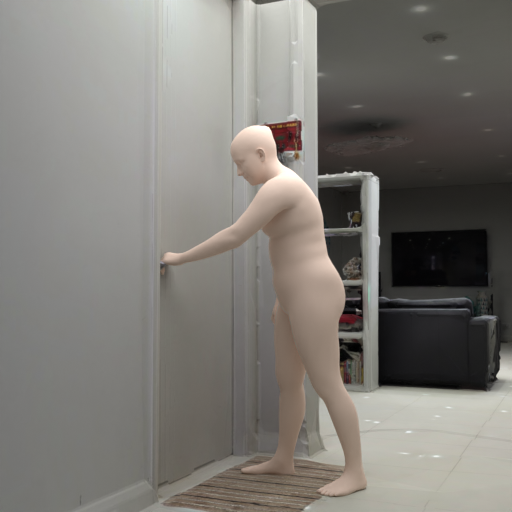} &
    \suppours{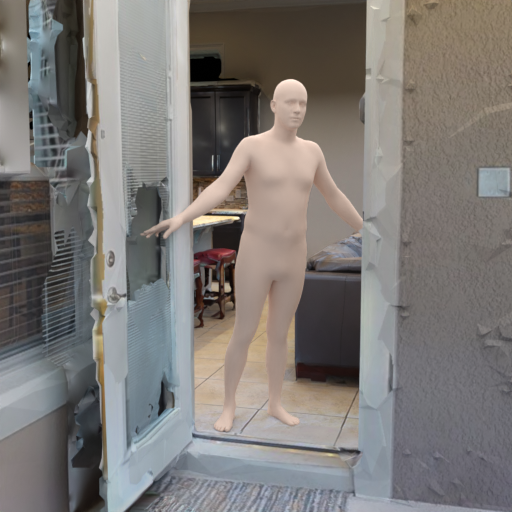} &
    \suppours{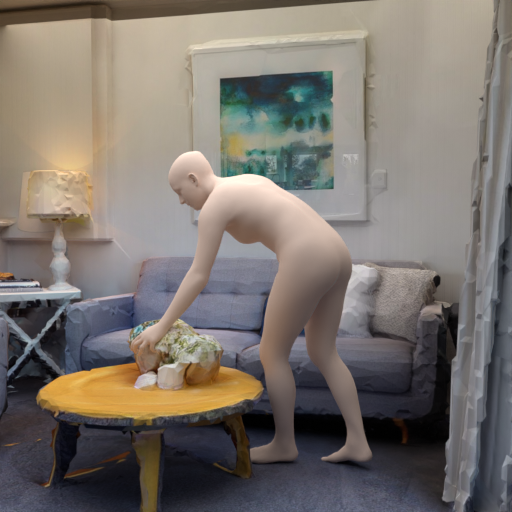} \\
    \descriptionsA
  \end{tabular}

  \vspace{0.5em}

  \begin{tabular}{cccccc}
    \rotatebox{90}{\hspace{1.2cm}GenZI} &
    \suppgenzi{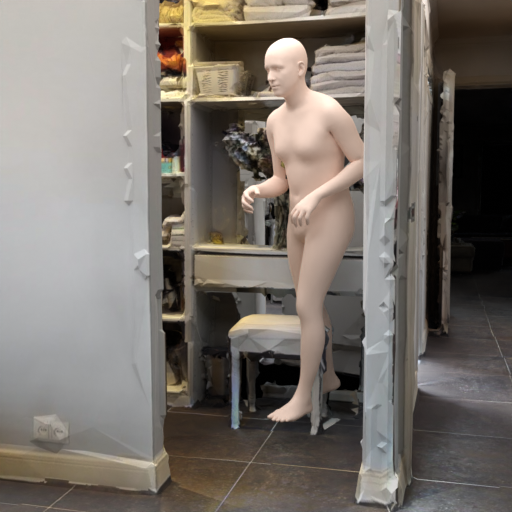} &
    \suppgenzi{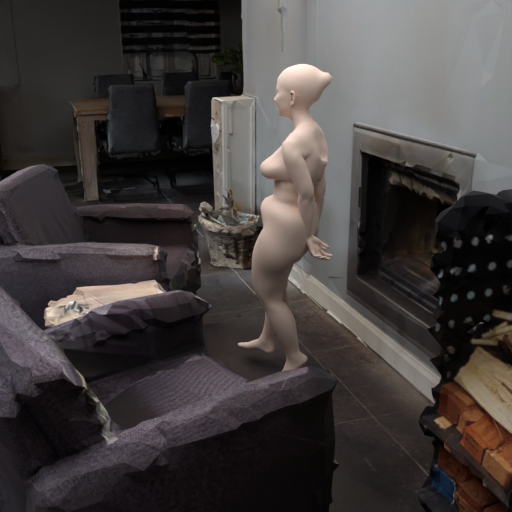} &
    \includegraphics[width=0.17\textwidth]{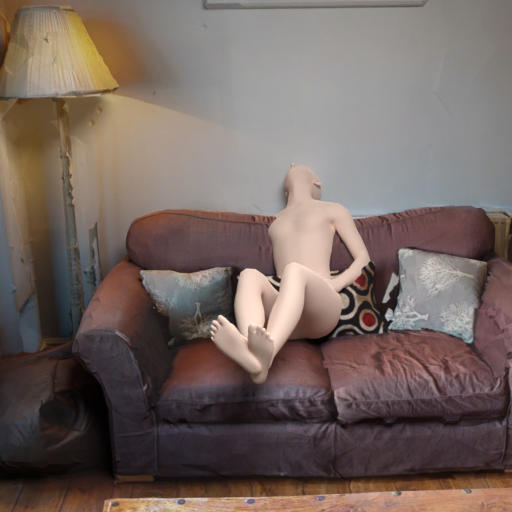} &
    \suppgenzi{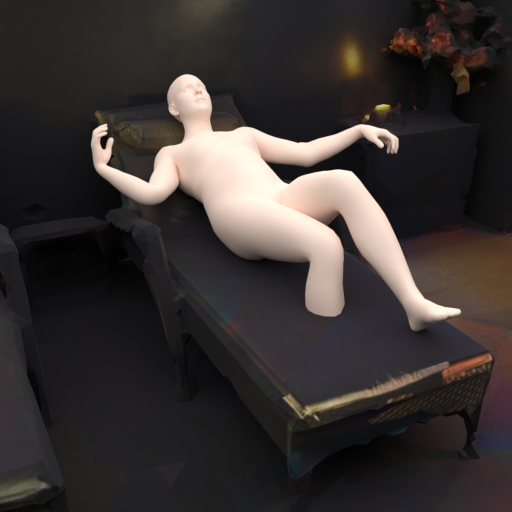} &
    \suppgenzi{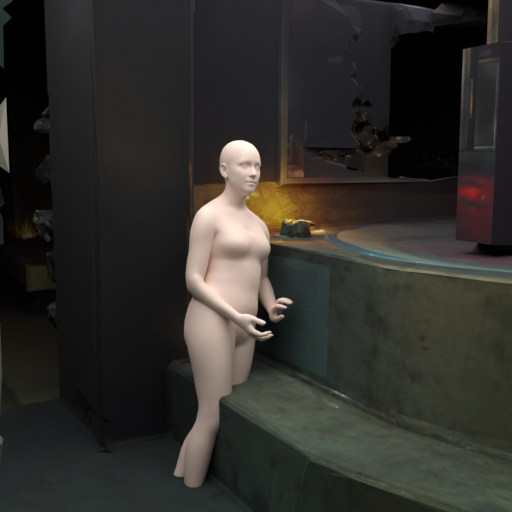} \\

    \rotatebox{90}{\hspace{1.15cm}POSA++} &
    \suppposa{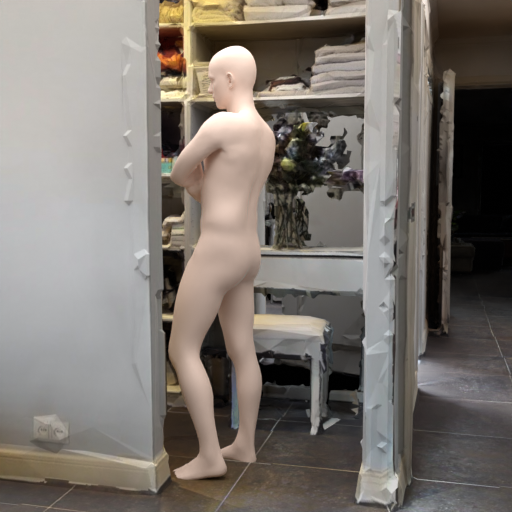} &
    \suppposa{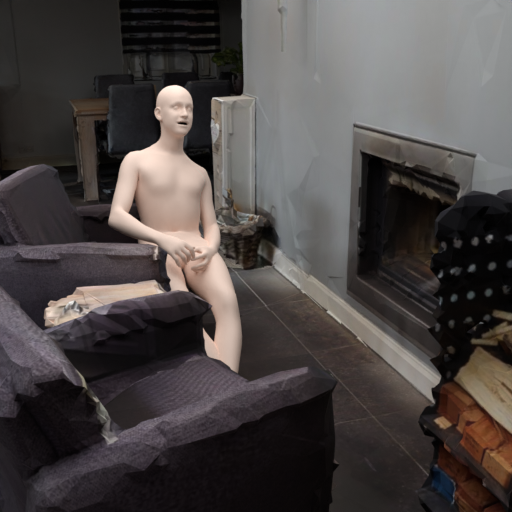} &
    \includegraphics[width=0.17\textwidth]{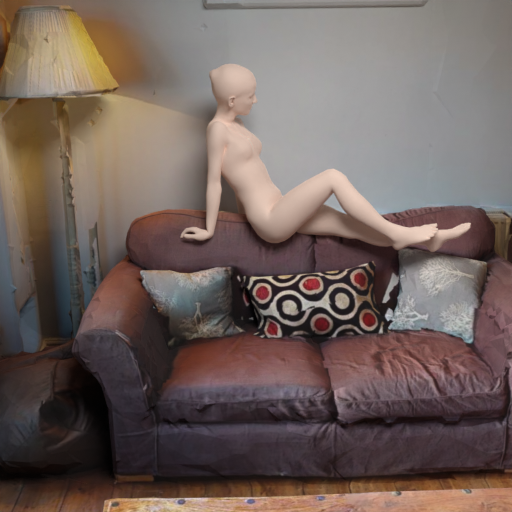} &
    \suppposa{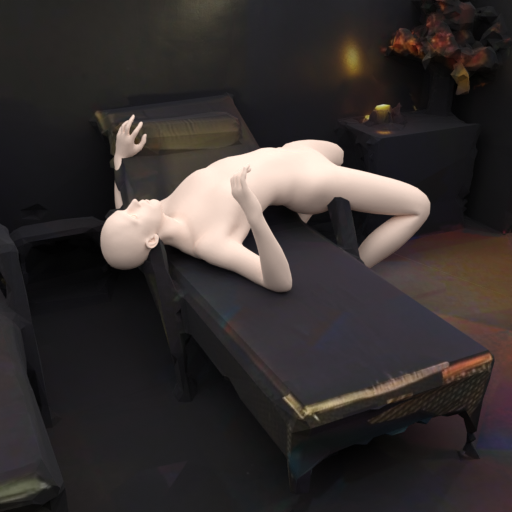} &
    \suppposa{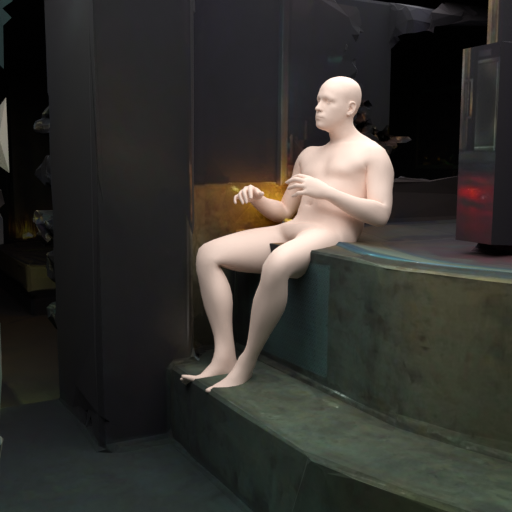} \\

    \rotatebox{90}{\hspace{0.7cm}\ours{} (Ours)} &
    \suppours{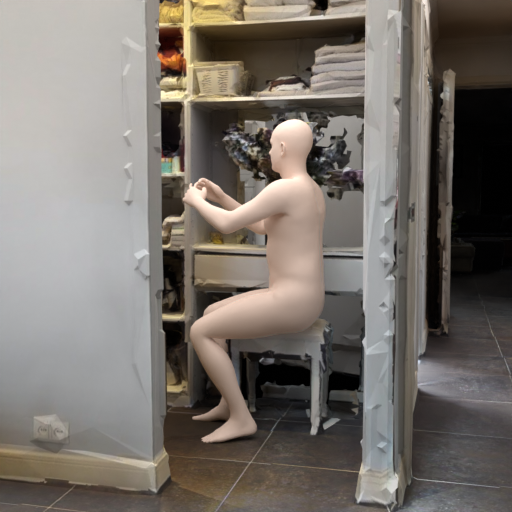} &
    \suppours{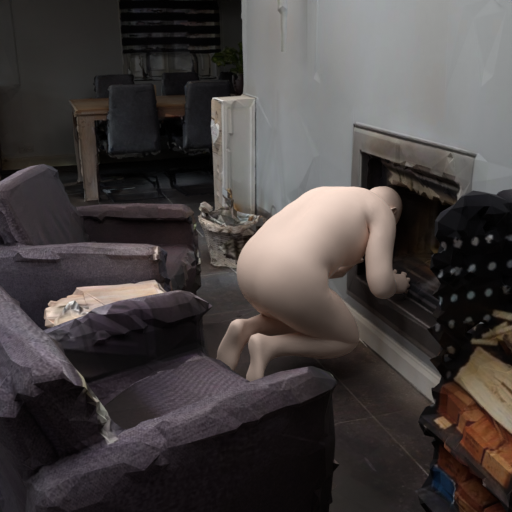} &
    \includegraphics[width=0.17\textwidth]{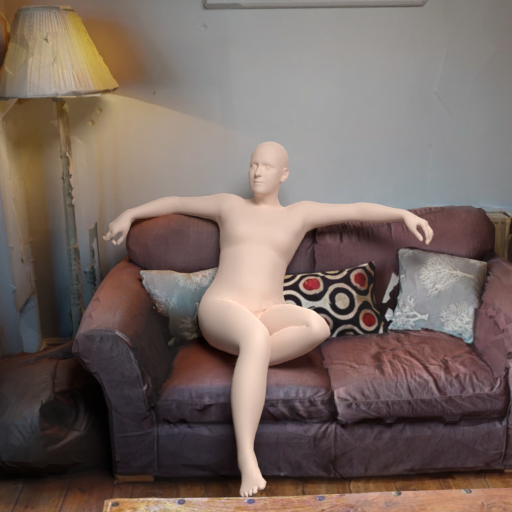} &
    \suppours{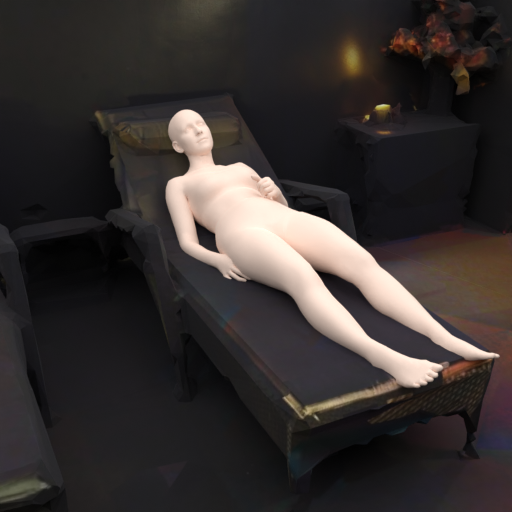} &
    \suppours{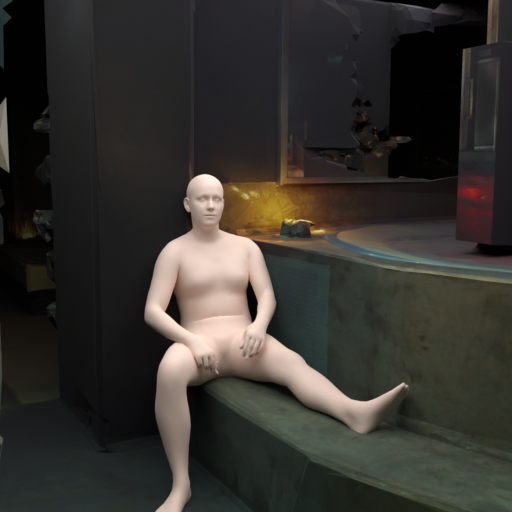} \\
    \descriptionsB
  \end{tabular}

  \caption{Additional qualitative comparisons of \ours{} (Ours) with GenZI~\citep{genzi} and \posa~\citep{posa}. Both baselines fail at physical plausibility (meshes stick to walls, feet do not touch the floor) and meaningful interaction (e.g., reaching for a book, touching a door handle, reaching for flowers), while \ours{} places people meaningfully without deep scene penetrations or floating bodies.}
  \label{fig:more:baselines}
\end{figure*}

\Cref{fig:multi:person} shows interactions involving several people in the same scene, which \ours{} produces from a single prompt to the image-editing model.
\Cref{fig:more:baselines} compares \ours{} with GenZI and \posa{} on further scenes and prompts.

We show more examples from our dataset \ourdata{} in \Cref{fig:suppl:data:1,fig:suppl:data:2,fig:suppl:data:3}.
Our data contains diverse scene interactions such as riding a gym bike (\Cref{fig:suppl:data:3}), looking for a clothing item (\Cref{fig:suppl:data:1}), measuring distances on a map (\Cref{fig:suppl:data:2}) and playing pinball (\Cref{fig:suppl:data:2}).

\section{Failure Cases}\label{suppl:sec:failures}
\begin{figure*}[t]
  \centering
  \def\imgdir{figures/failures/img}
  \setlength{\tabcolsep}{1pt}
  \renewcommand{\arraystretch}{0.5}
  \begin{tabular}{c c c c}
    \rotatebox{90}{\makebox[0.197\linewidth][c]{\scriptsize (a) Implausible edit}} &
    \includegraphics[width=0.27\linewidth]{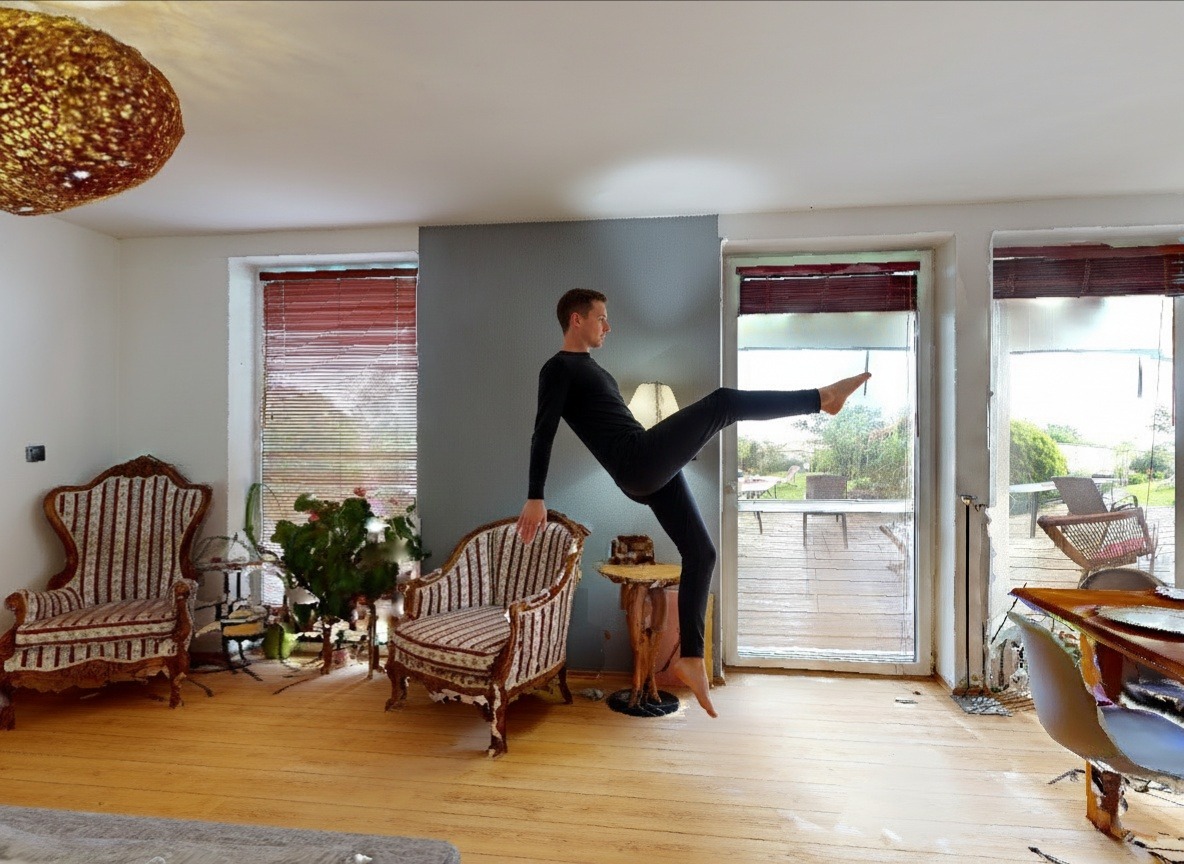} &
    \includegraphics[width=0.27\linewidth]{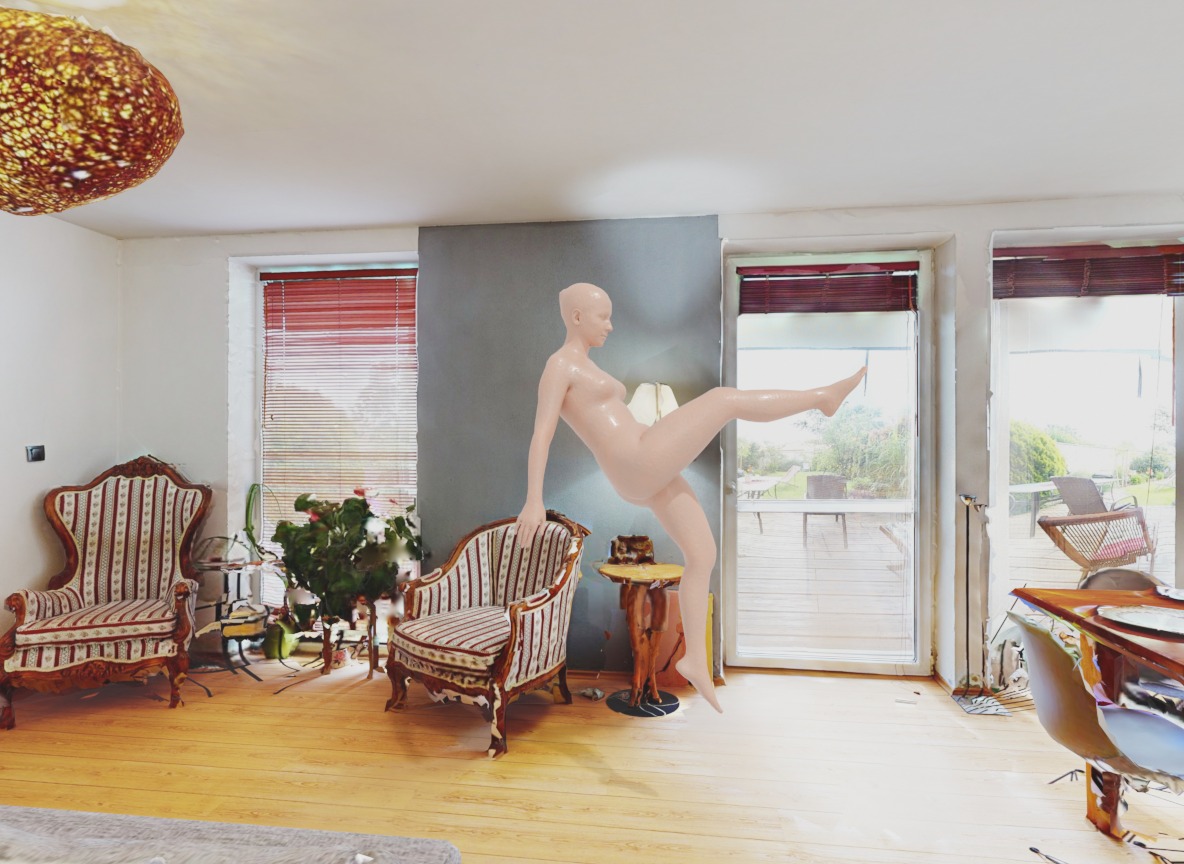} &
    \includegraphics[width=0.27\linewidth]{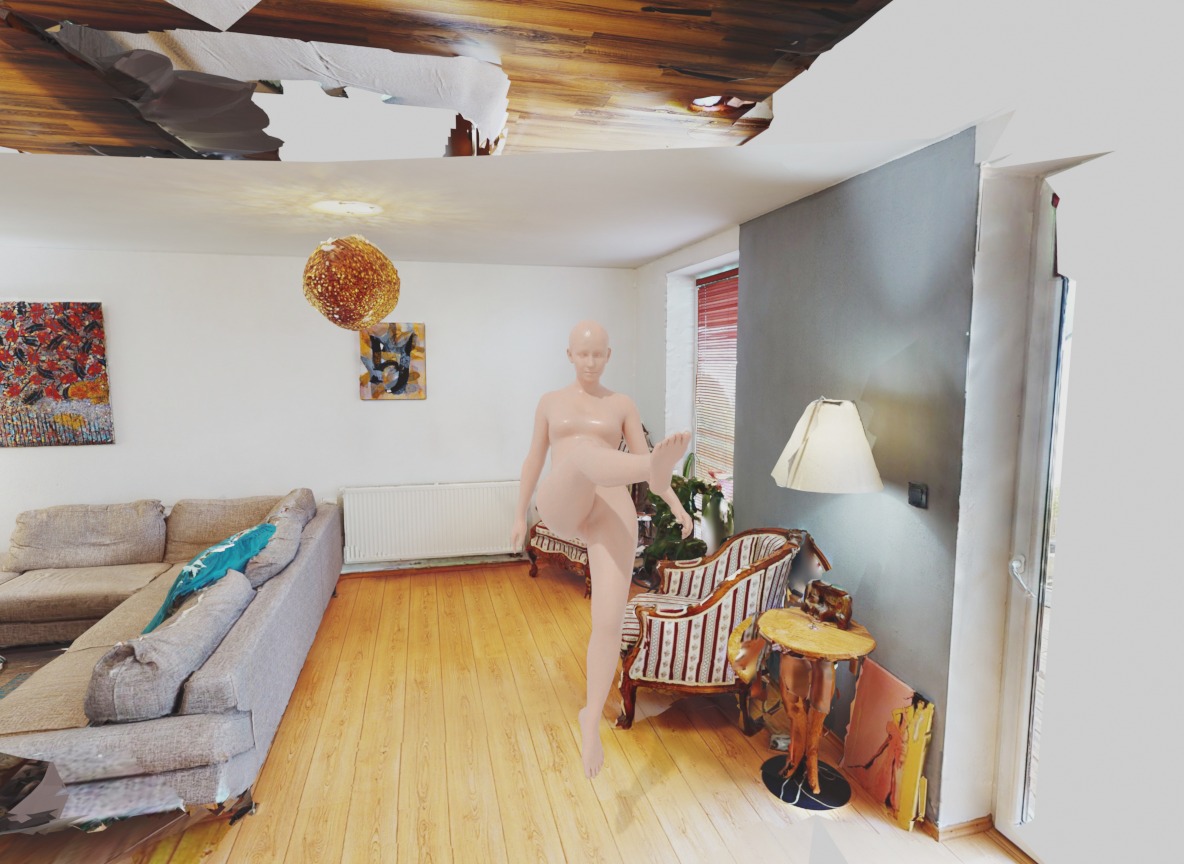} \\
    & \scriptsize RGB Image & \scriptsize 3D Annotation & \scriptsize Side view \\[4pt]

    \rotatebox{90}{\makebox[0.197\linewidth][c]{\scriptsize (b) Occlusion}} &
    \includegraphics[width=0.27\linewidth]{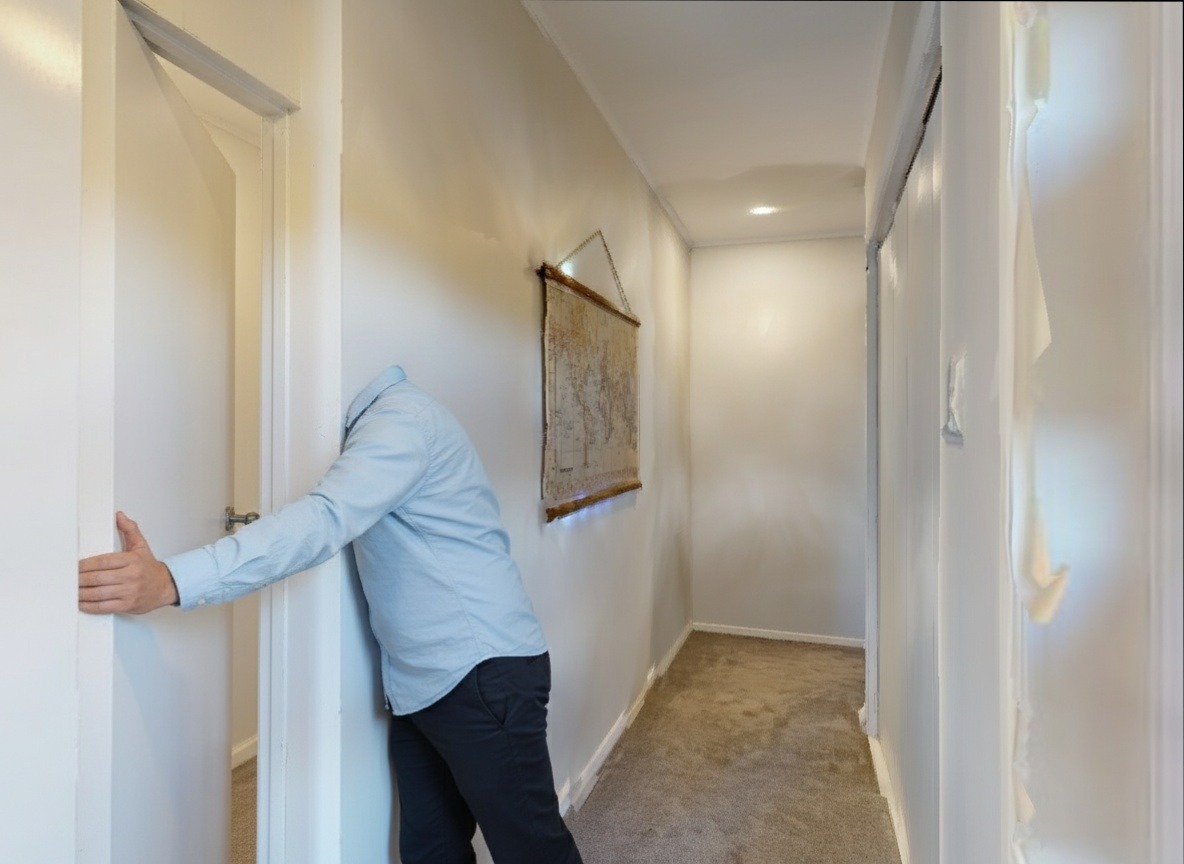} &
    \includegraphics[width=0.27\linewidth]{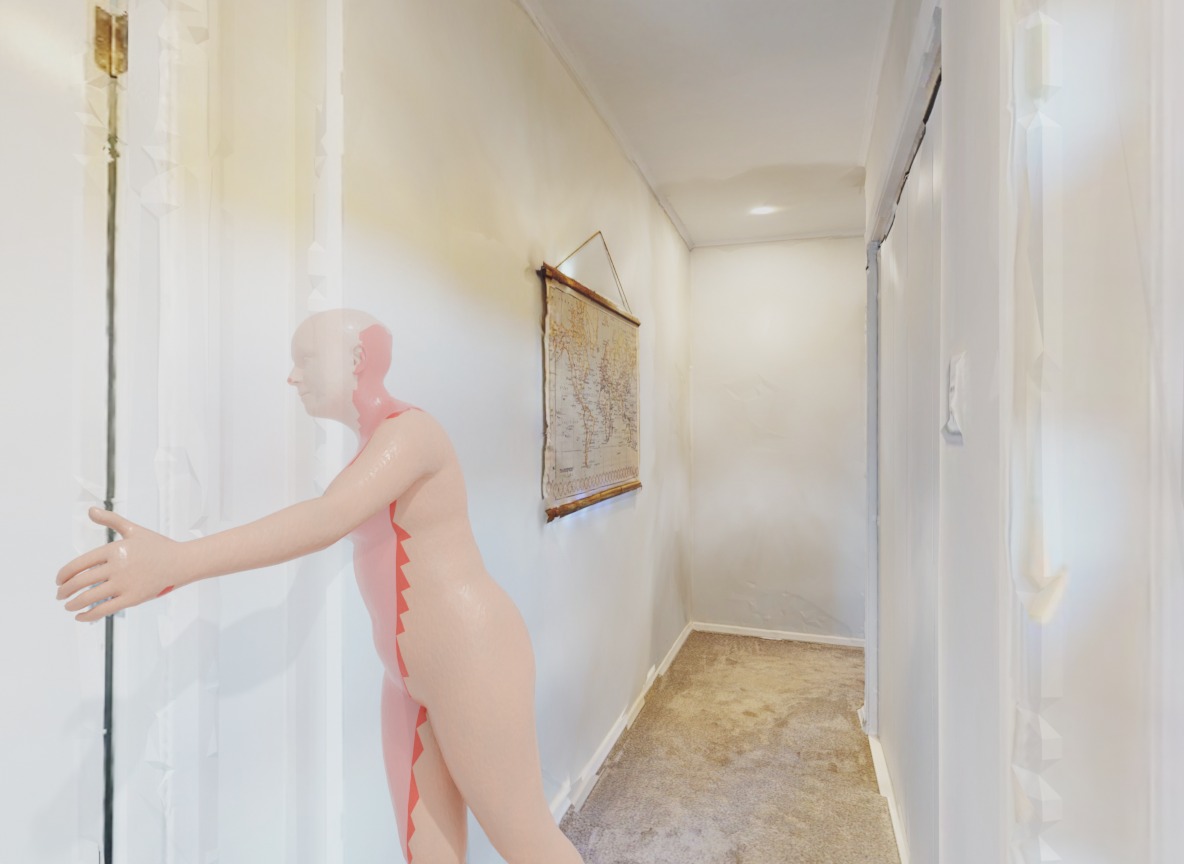} &
    \includegraphics[width=0.27\linewidth]{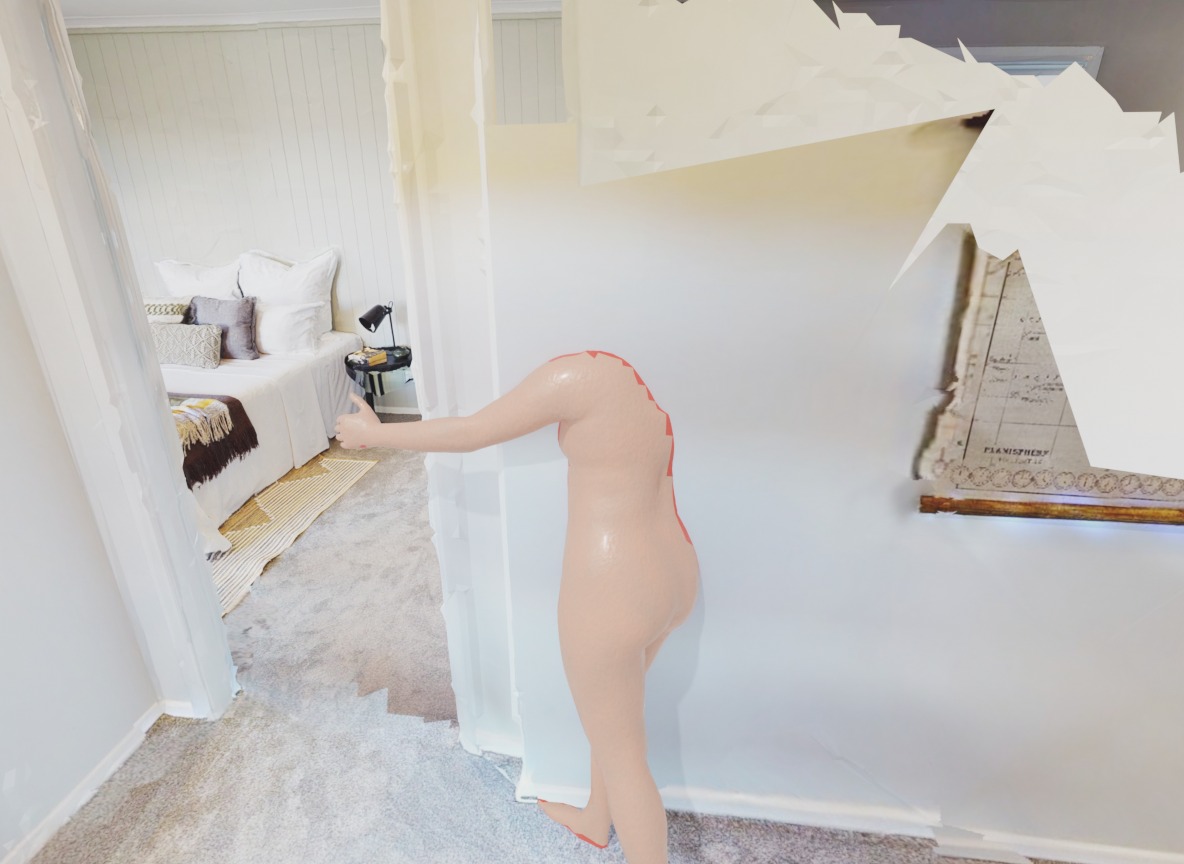} \\
    & \scriptsize RGB Image & \scriptsize 3D Annotation & \scriptsize Side view \\[4pt]

    \rotatebox{90}{\makebox[0.197\linewidth][c]{\scriptsize (c) Truncation}} &
    \includegraphics[width=0.27\linewidth]{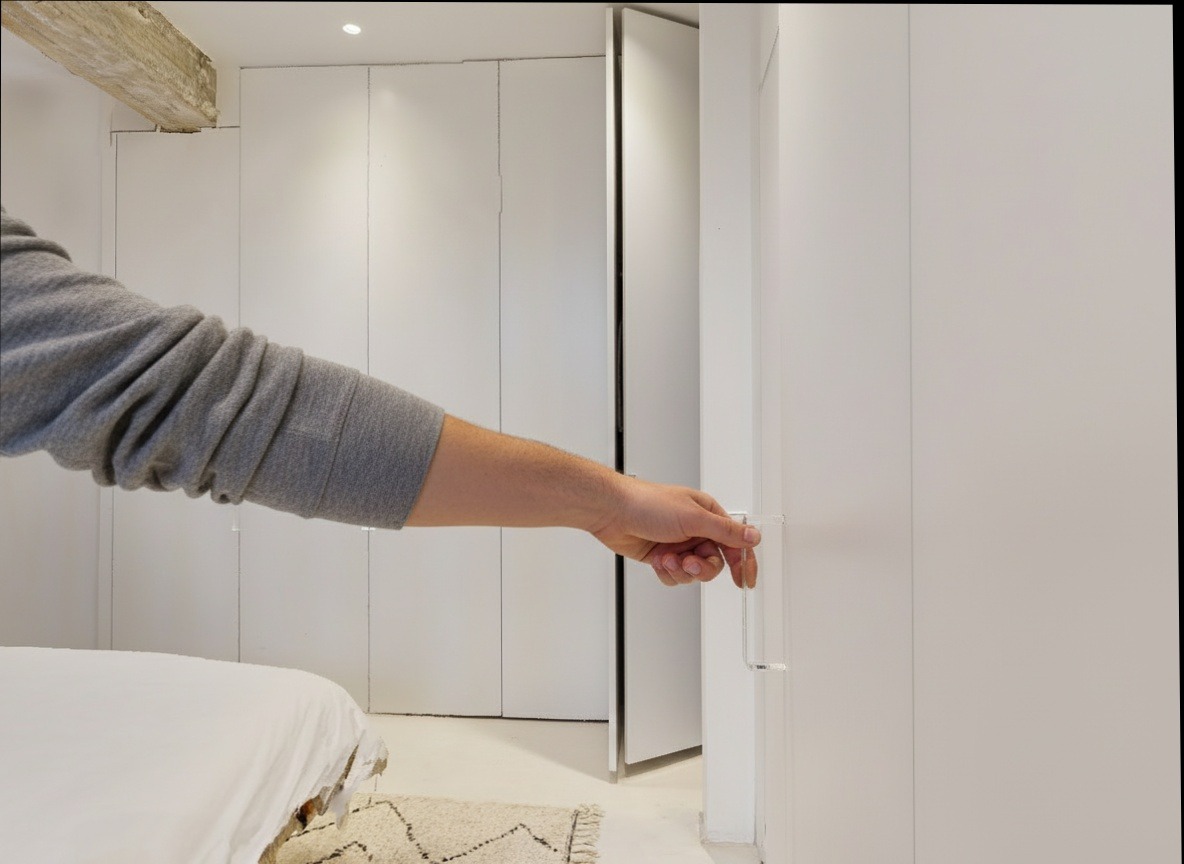} &
    \includegraphics[width=0.27\linewidth]{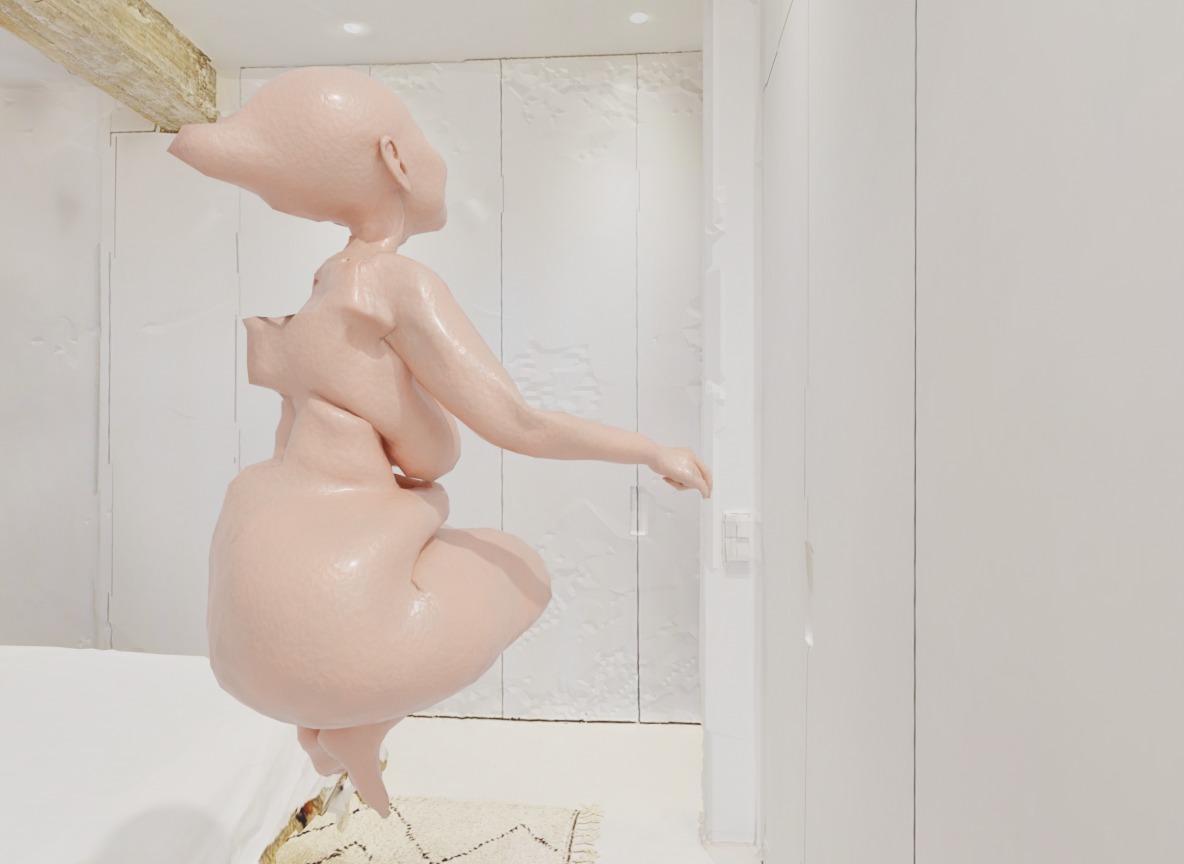} &
    \includegraphics[width=0.27\linewidth]{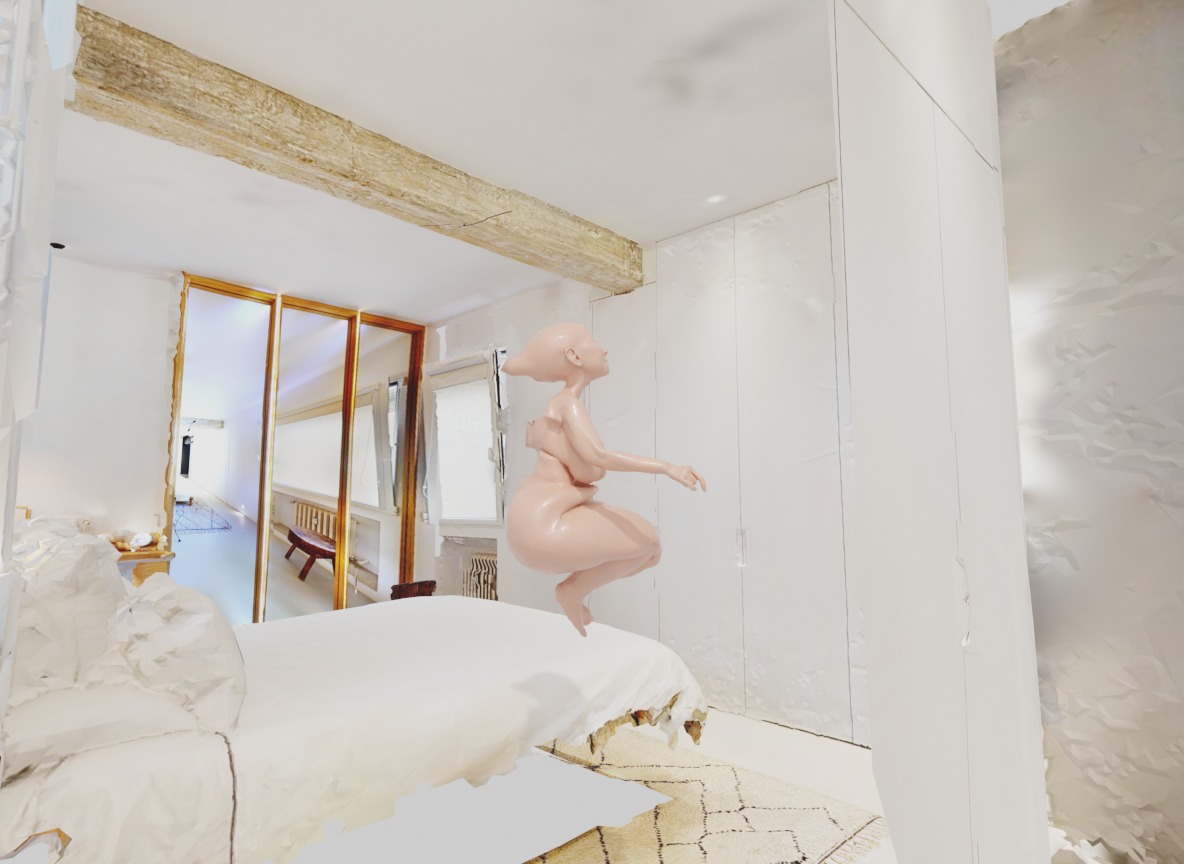} \\
    & \scriptsize RGB Image & \scriptsize 3D Annotation & \scriptsize Side view \\[4pt]

    \rotatebox{90}{\makebox[0.197\linewidth][c]{\scriptsize (d) Scene edit}} &
    \includegraphics[width=0.27\linewidth]{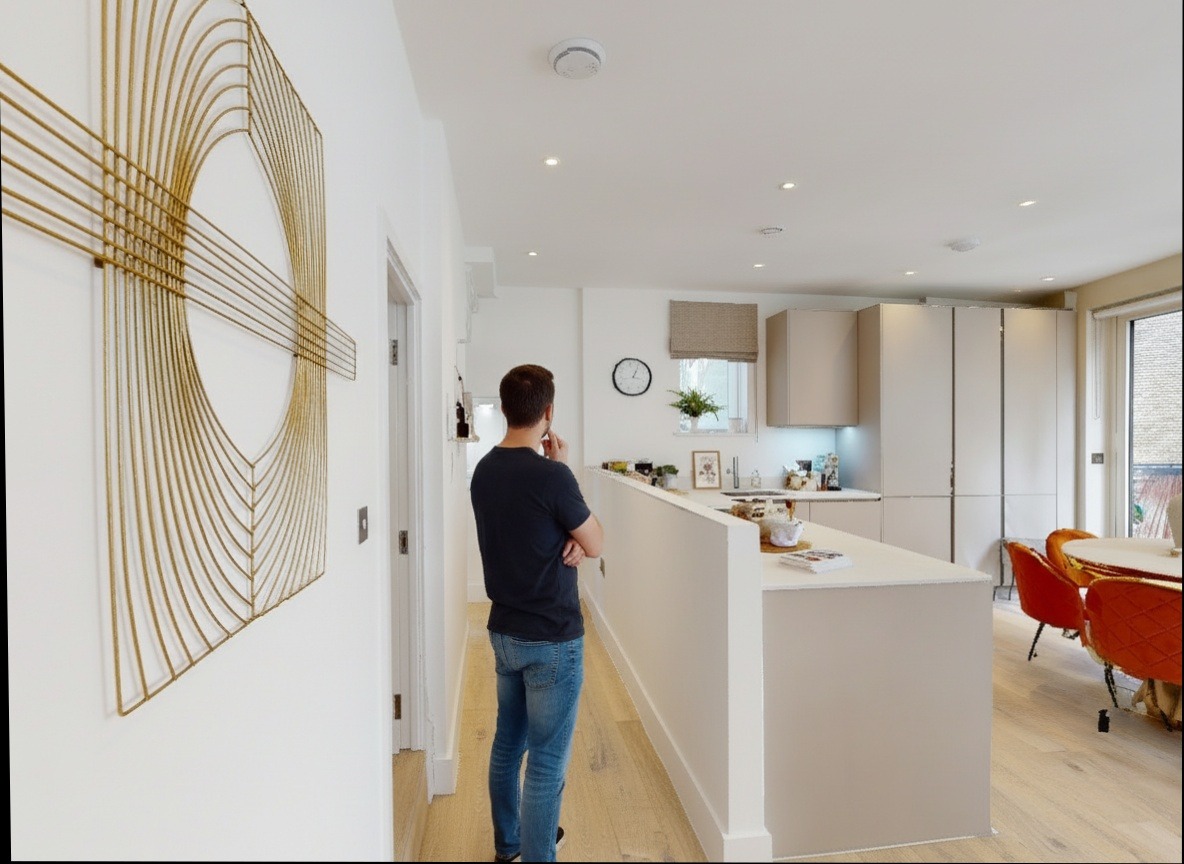} &
    \includegraphics[width=0.27\linewidth]{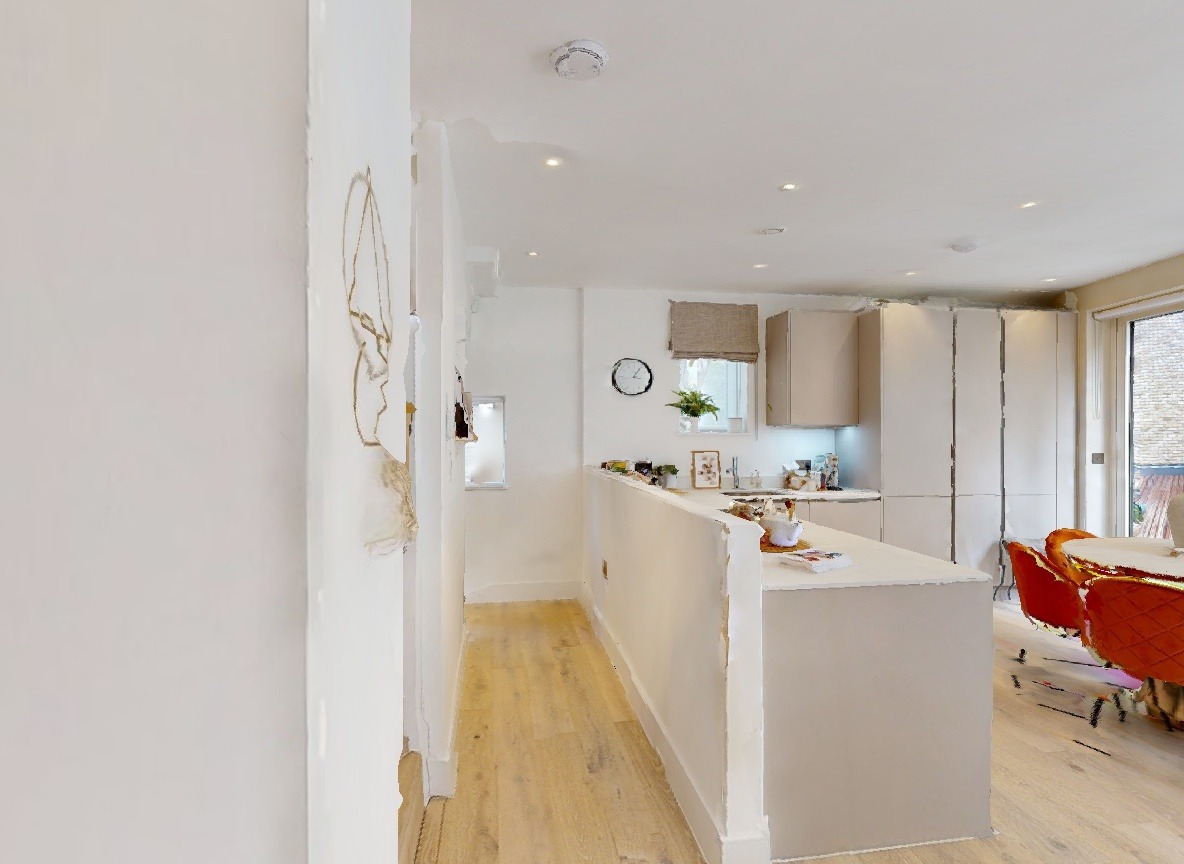} &
    \includegraphics[width=0.27\linewidth]{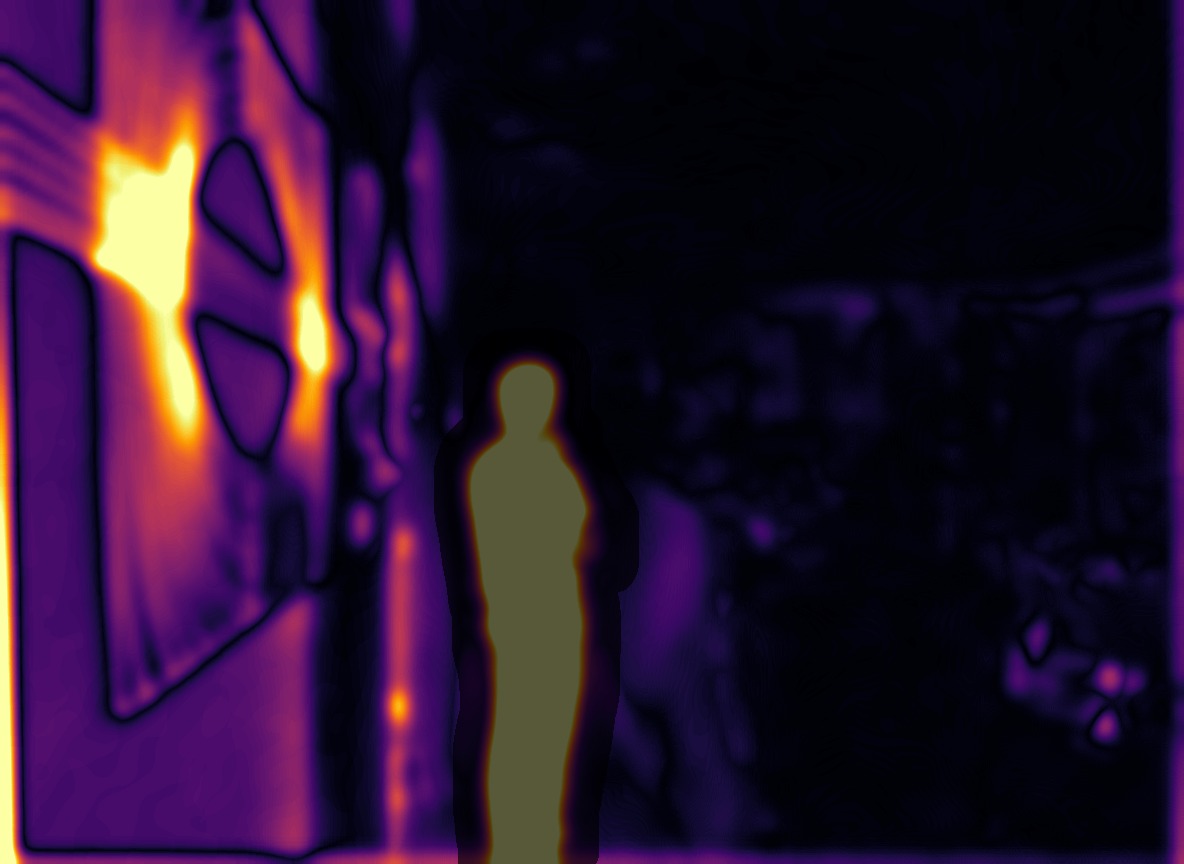} \\
    & \scriptsize Edited image & \scriptsize Rendered scene & \scriptsize Change outside the person \\[4pt]

    \rotatebox{90}{\makebox[0.197\linewidth][c]{\scriptsize (e) Scan holes}} &
    \includegraphics[width=0.27\linewidth]{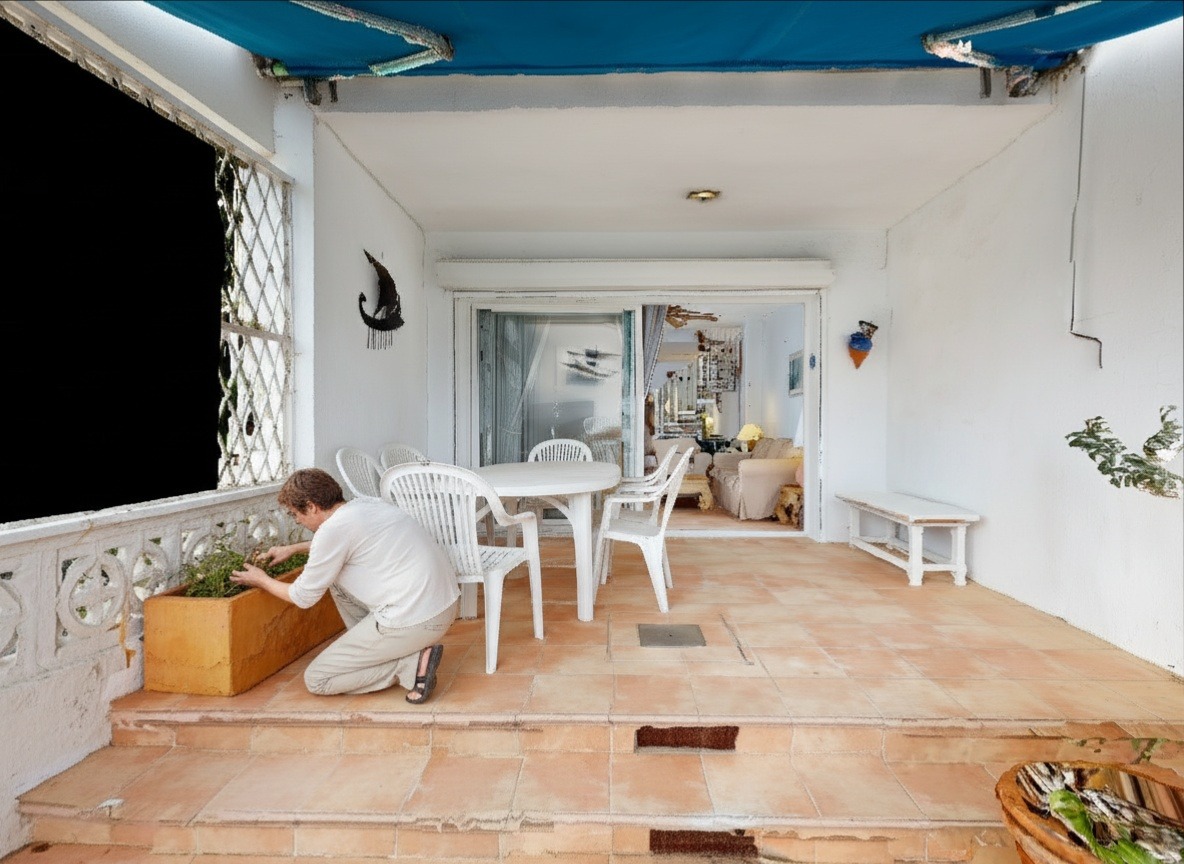} &
    \includegraphics[width=0.27\linewidth]{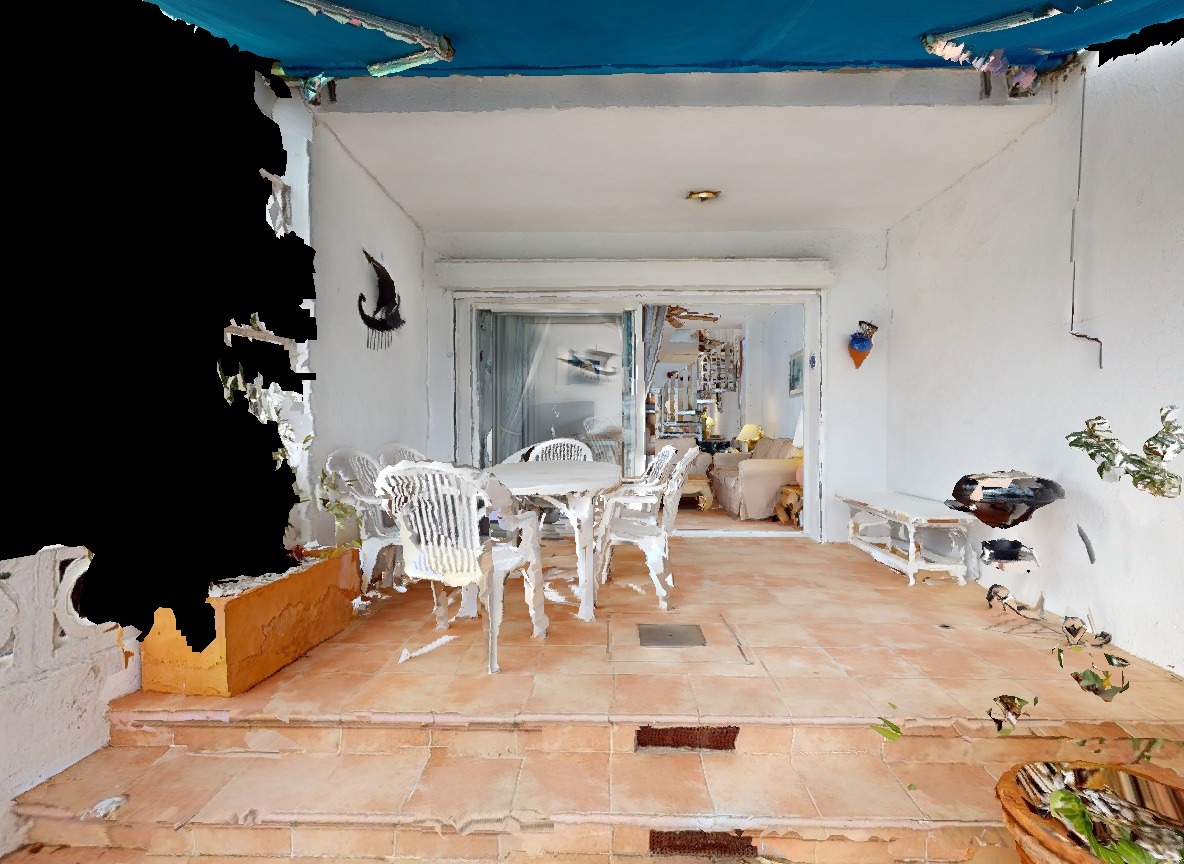} &
    \includegraphics[width=0.27\linewidth]{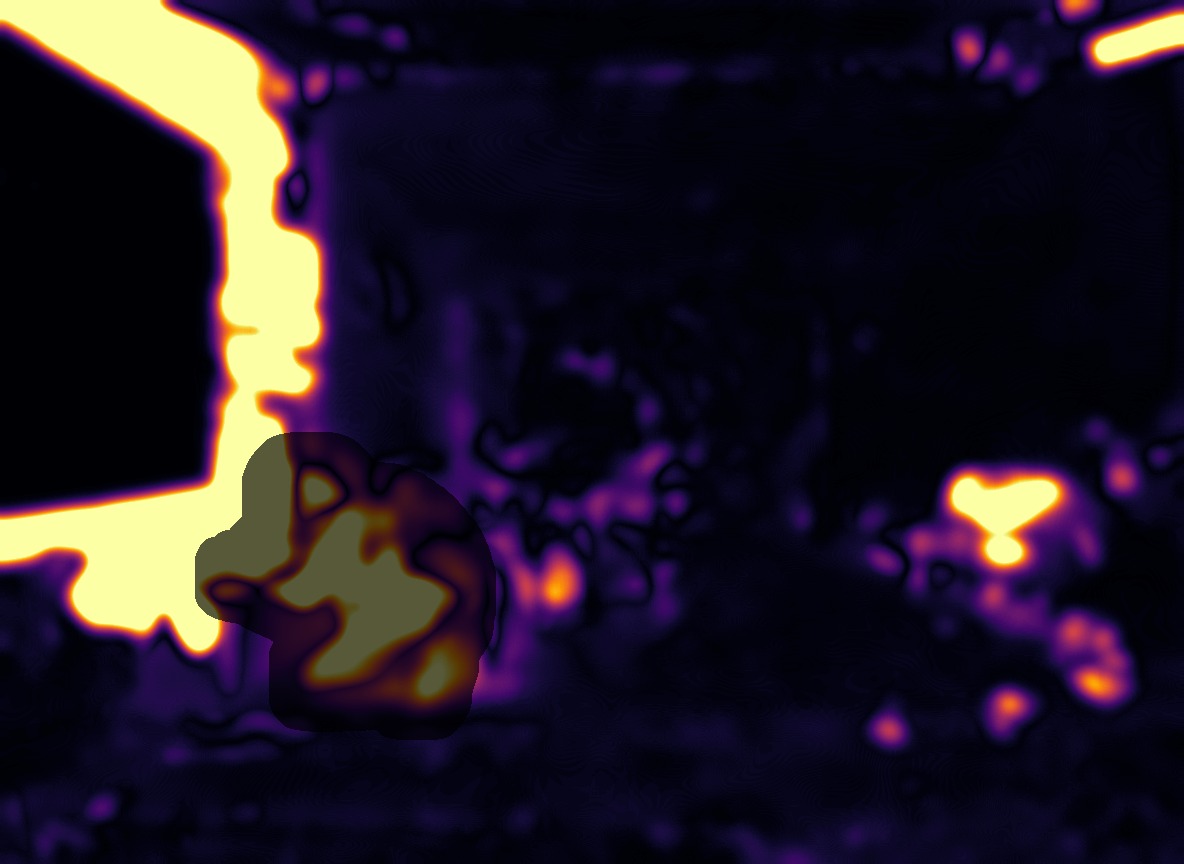} \\
    & \scriptsize Edited image & \scriptsize Rendered scene (holes in black) & \scriptsize Change outside the person \\
  \end{tabular}
  \caption{\textbf{Failure cases.} Many of these failure modes originate in the image editor and are expected to diminish as editing models improve.
    (a)~The editor produces a person without physical support; lifting faithfully reproduces the floating pose.
    (b)~In the edit the head is hidden behind a door frame; without evidence for it, the lifted body is embedded $\sim$7\,cm in the wall (input view: occluded body parts shown semi-transparent, penetration in red).
    (c)~Only an arm is visible; the lifted body is a crouched figure floating in front of the camera. In rare cases such as this one, a truncated insertion passes the visibility filter; more aggressive filtering can remove it.
    (d)~The editor alters the scene far from the person; geometry changes near the person are mostly detected by our depth-based filter.
    (e)~The editor paints over holes in the scanned mesh; a more complete scene dataset mitigates this.}
  \label{fig:failures}
\end{figure*}

\Cref{fig:failures} shows representative failure cases of \ours{}. Many originate in the image editor: lifting faithfully reconstructs what the edited image shows, including a person without physical support (a). The other cases are local, and most are handled by the quality control of \Cref{suppl:sec:qc}. The visibility filter removes most strongly truncated insertions; rare cases such as (c) pass it, and more aggressive filtering can remove them. Scene changes far from the person (d) can be removed by compositing only the edited patch into the rendered scene, while changes close to the person are mostly caught by the depth-based filter. When the edit hides a body part (b), the lifted body can penetrate the scene by a few centimeters, and holes in the scanned meshes (e) stem from the source scans. Since \ours{} is modular, its quality will improve with advances in reconstruction and image generation models.

\newlength{\suppS}\setlength{\suppS}{62.6pt}
\newlength{\suppG}\setlength{\suppG}{6.9pt}
\newcommand{\suppdatapair}[1]{%
  \includegraphics[width=\suppS]{assets/suppl_ours_data/#1_scene_crop.jpg}\hspace{1pt}%
  \includegraphics[width=\suppS]{assets/suppl_ours_data/#1_crop.jpg}%
}
\newcommand{\suppdatarow}[3]{%
  \suppdatapair{#1}\hspace{\suppG}\suppdatapair{#2}\hspace{\suppG}\suppdatapair{#3}\par
}

\begin{figure*}[p]
  \centering
  \begin{minipage}{\textwidth}
    \offinterlineskip
    \suppdatarow{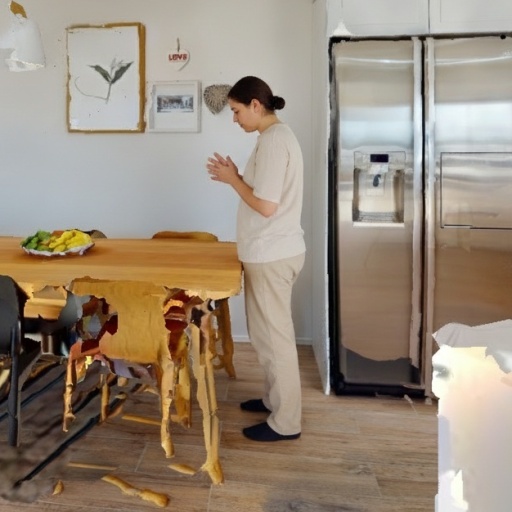}{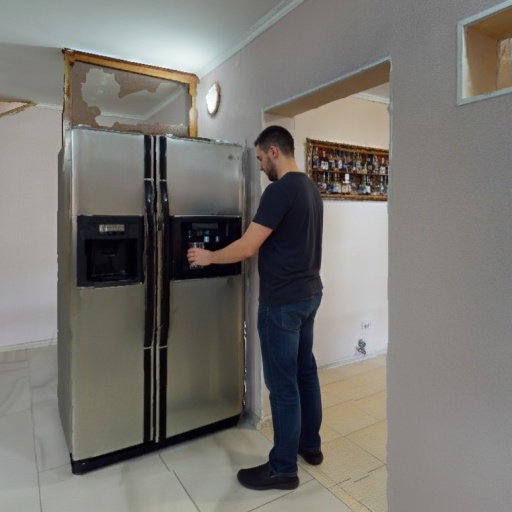}{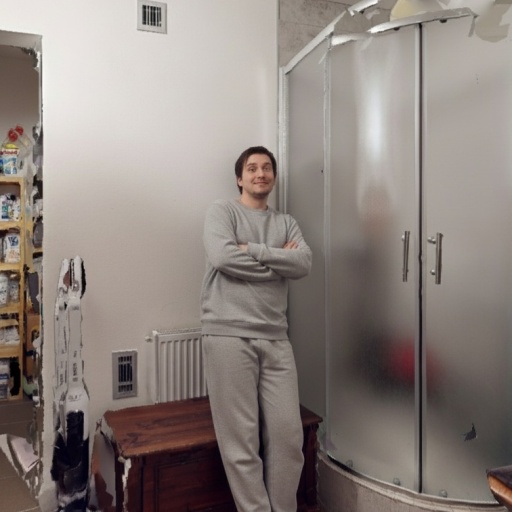}
    \vspace{\suppG}
    \suppdatarow{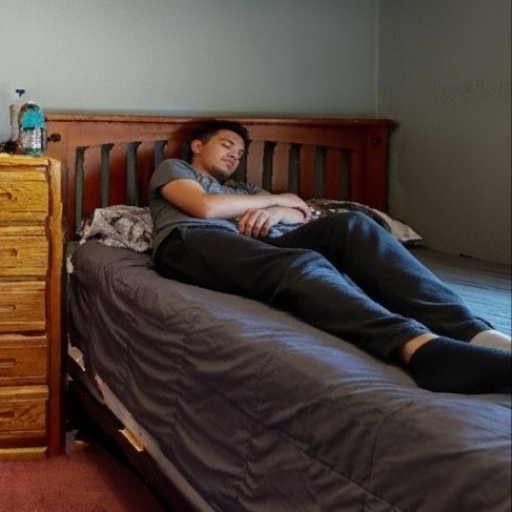}{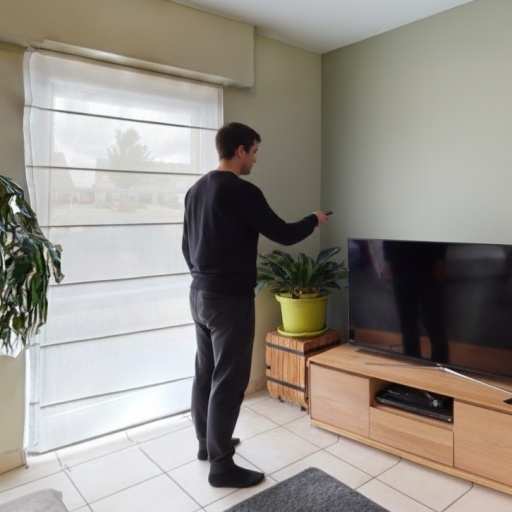}{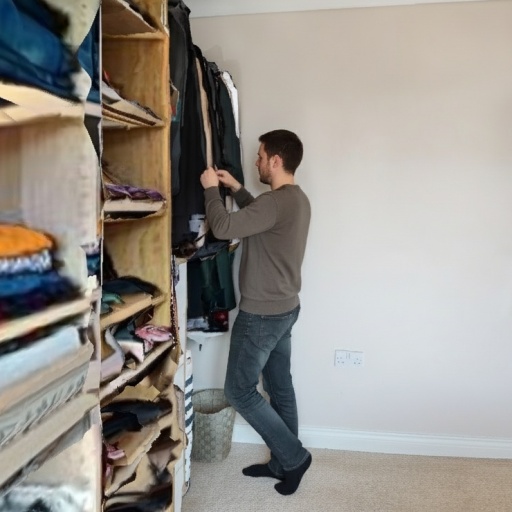}
    \vspace{\suppG}
    \suppdatarow{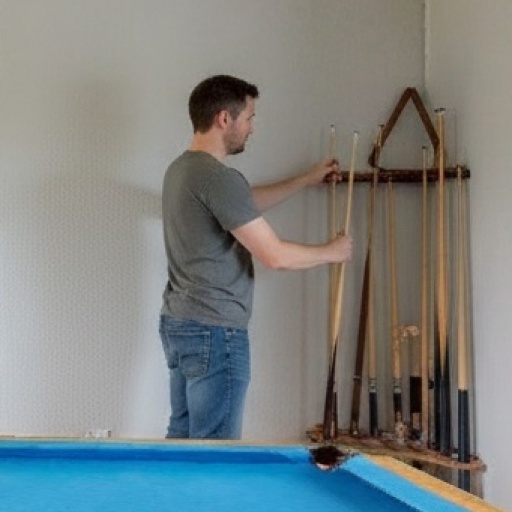}{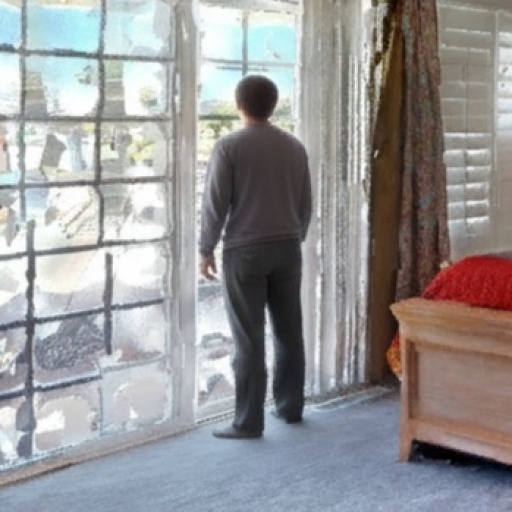}{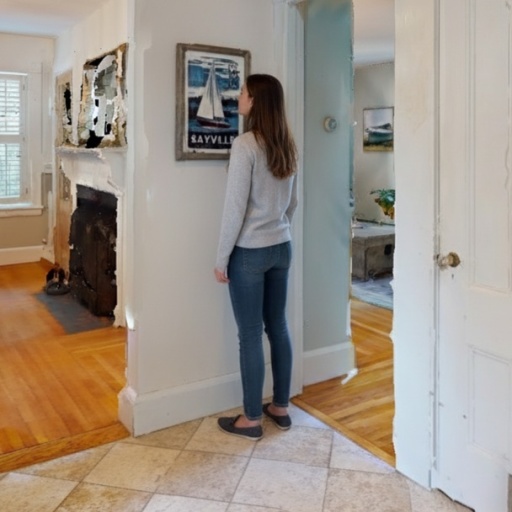}
    \vspace{\suppG}
    \suppdatarow{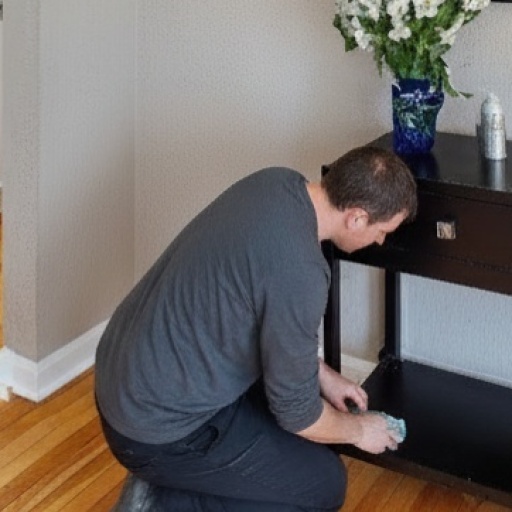}{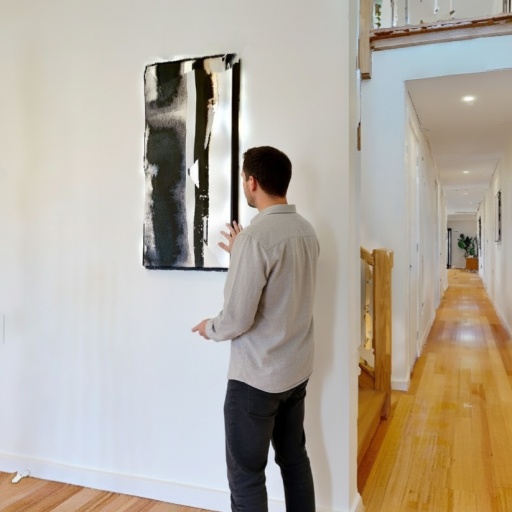}{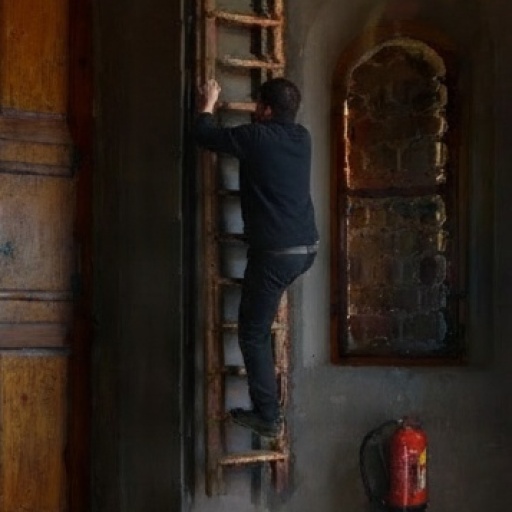}
    \vspace{\suppG}
    \suppdatarow{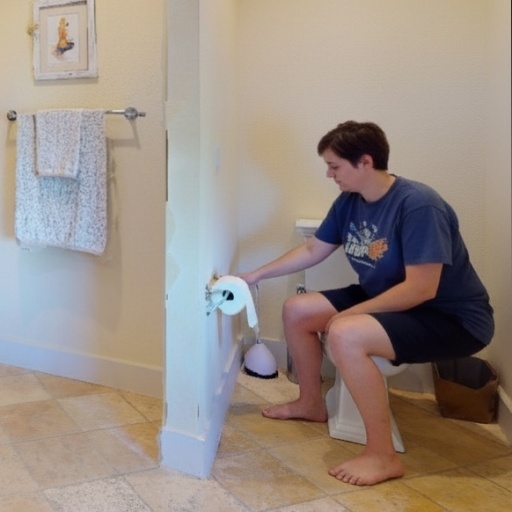}{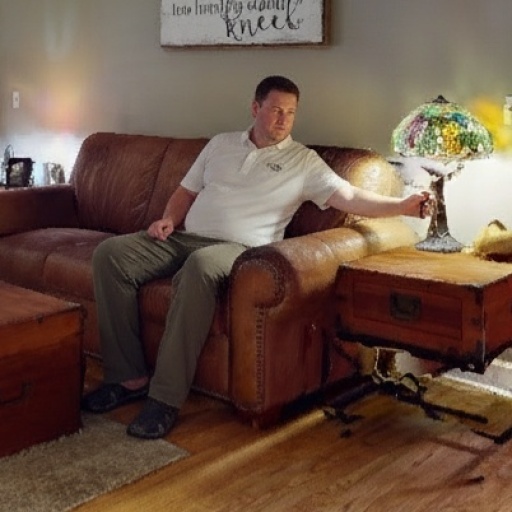}{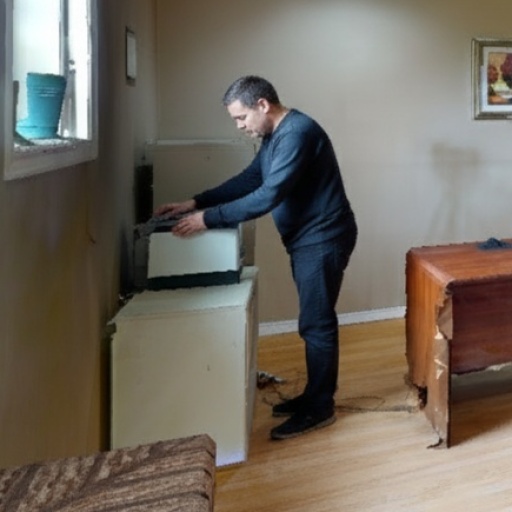}
    \vspace{\suppG}
    \suppdatarow{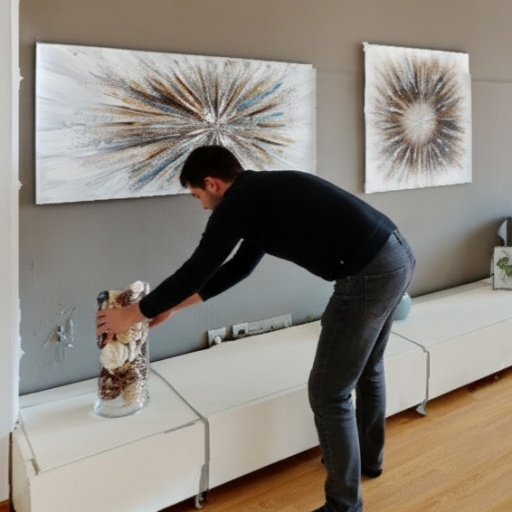}{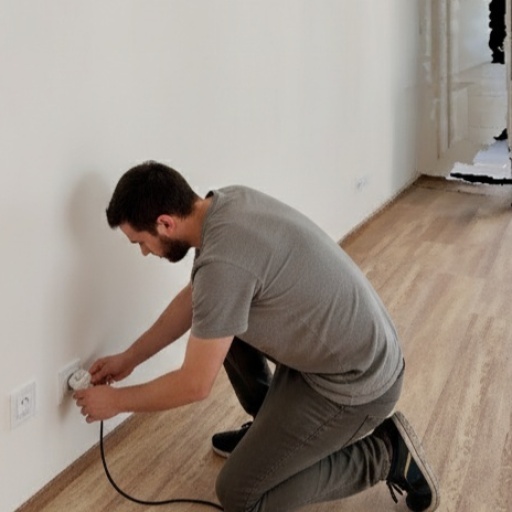}{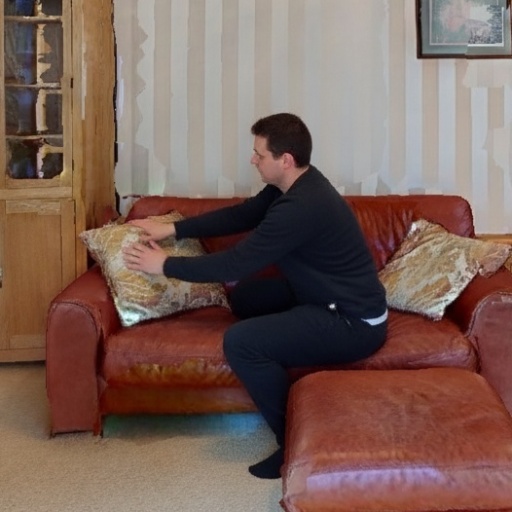}
    \vspace{\suppG}
    \suppdatarow{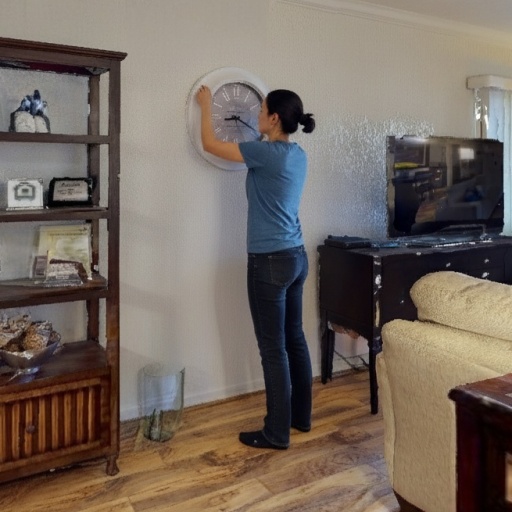}{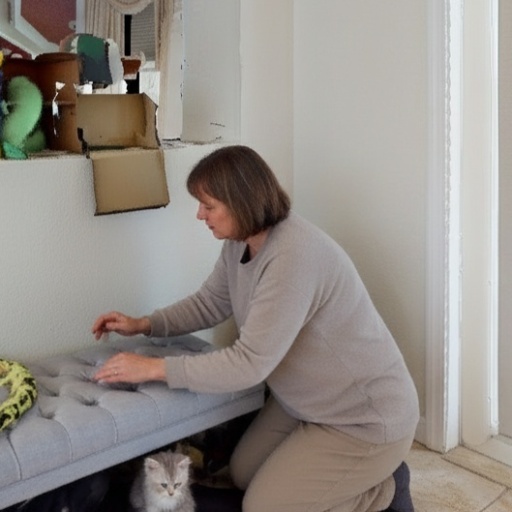}{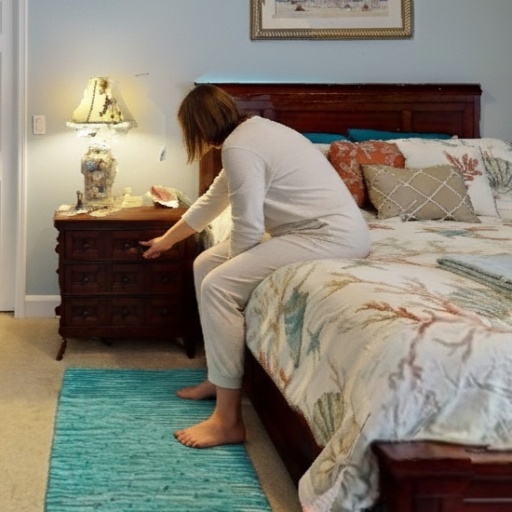}
    \vspace{\suppG}
    \suppdatarow{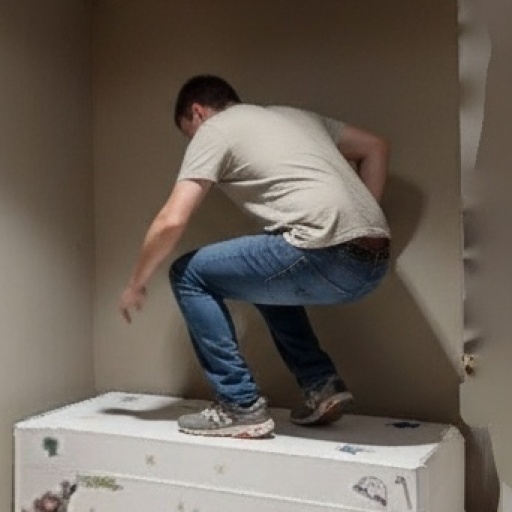}{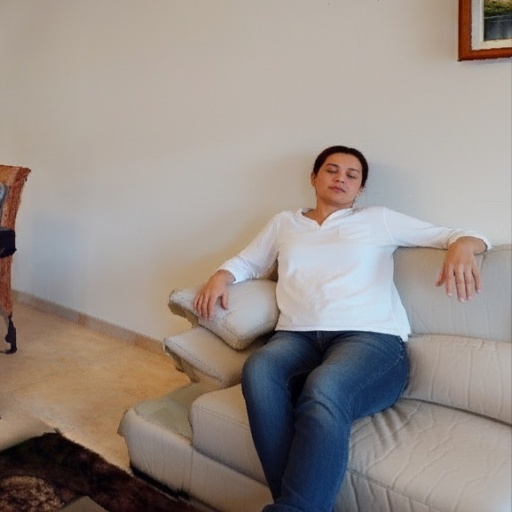}{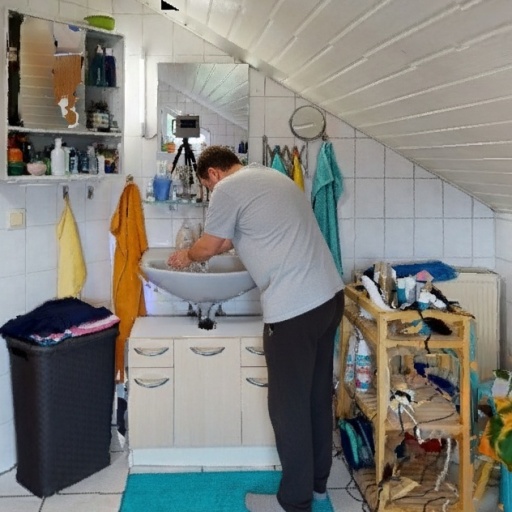}
    \vspace{\suppG}
    \suppdatarow{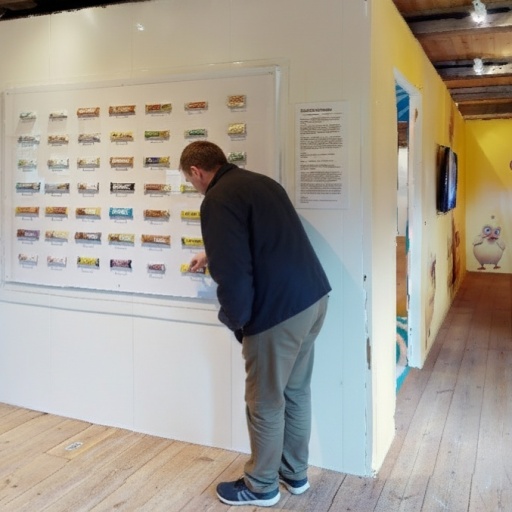}{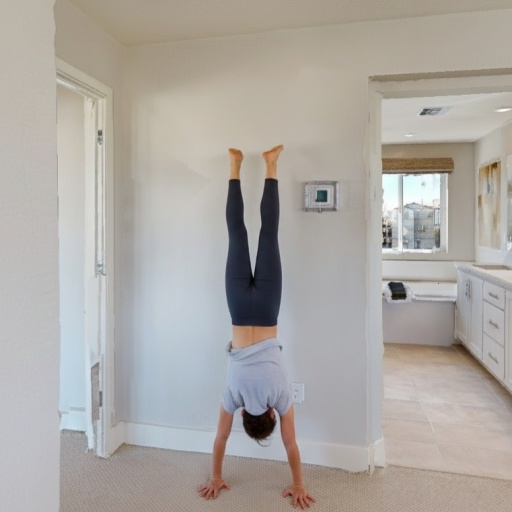}{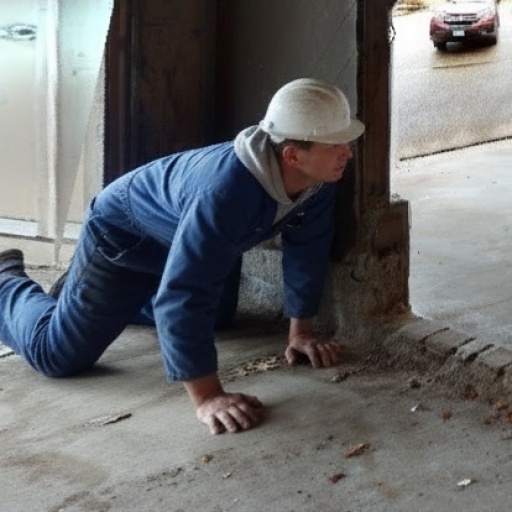}
  \end{minipage}
  \caption{Additional samples from \ourdata{}. Each pair shows the edited RGB image (left) and the corresponding lifted 3D human in the 3D scene (right).}
  \label{fig:suppl:data:1}
\end{figure*}

\begin{figure*}[p]
  \centering
  \begin{minipage}{\textwidth}
    \offinterlineskip
    \suppdatarow{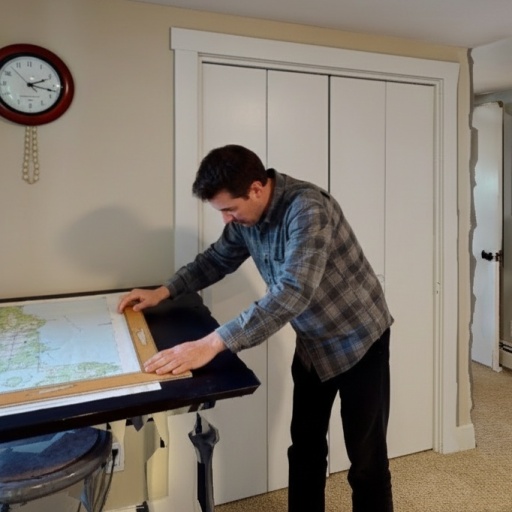}{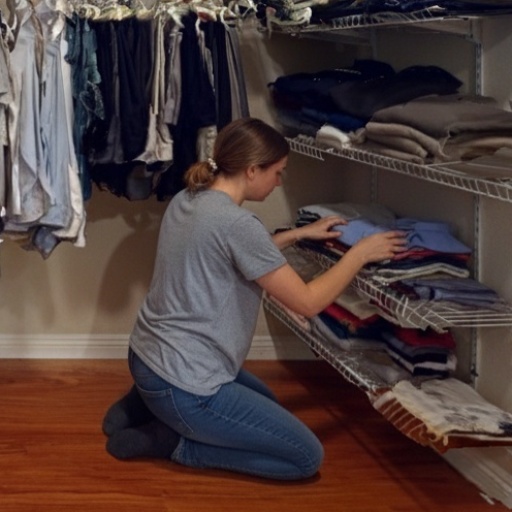}{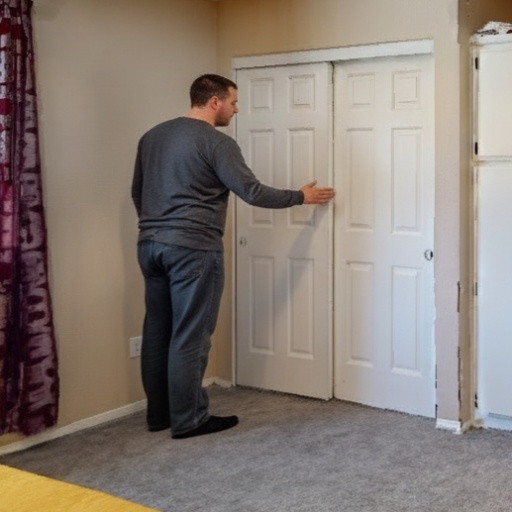}
    \vspace{\suppG}
    \suppdatarow{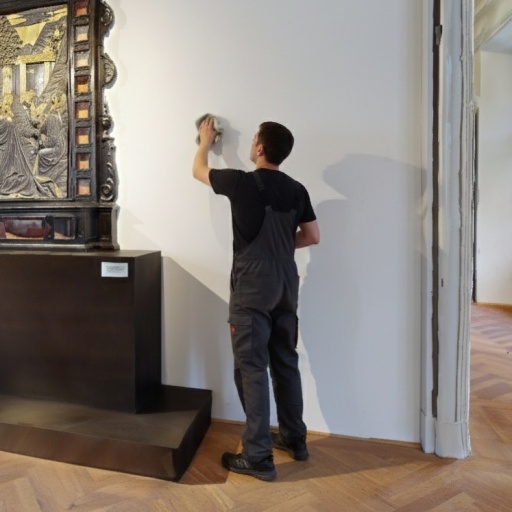}{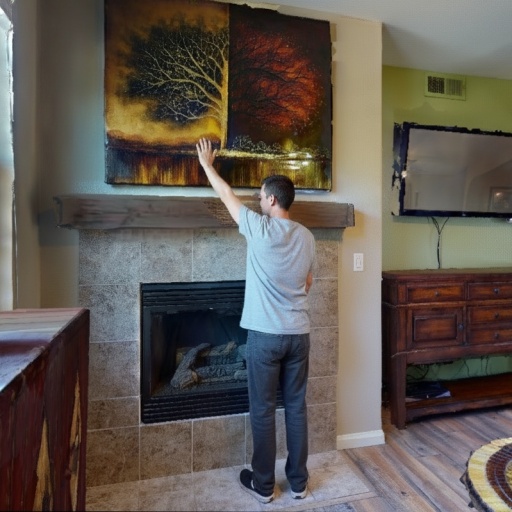}{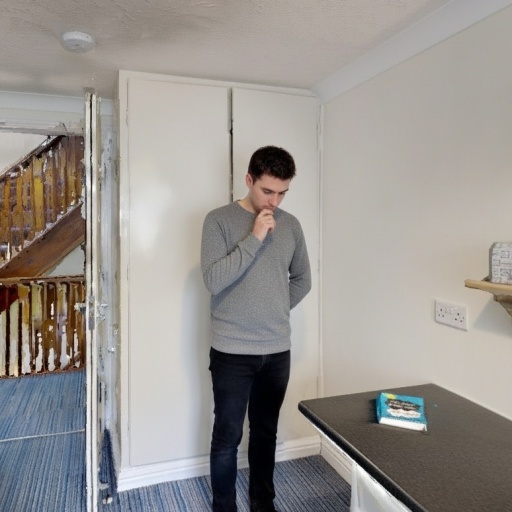}
    \vspace{\suppG}
    \suppdatarow{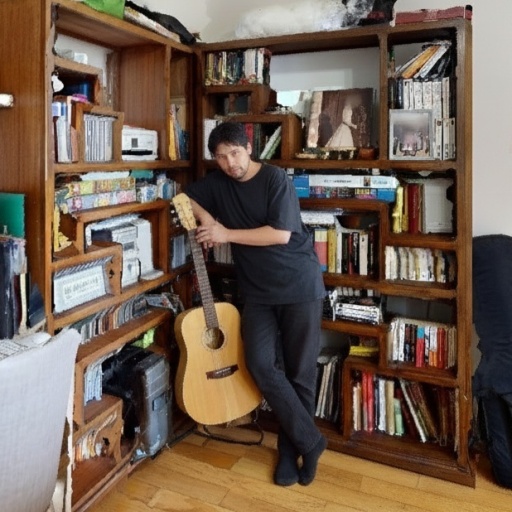}{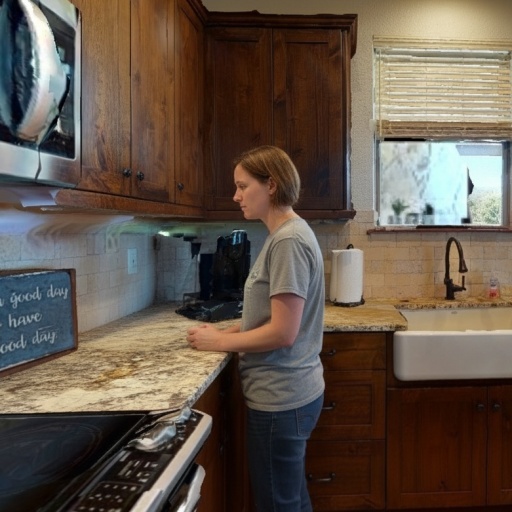}{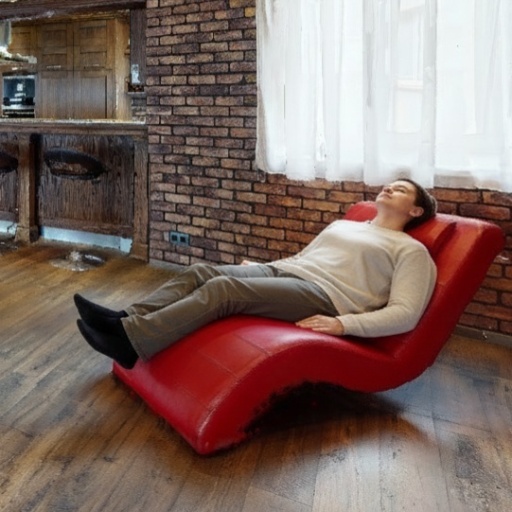}
    \vspace{\suppG}
    \suppdatarow{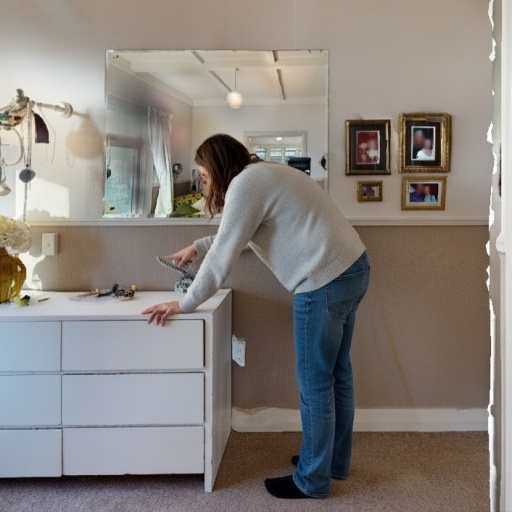}{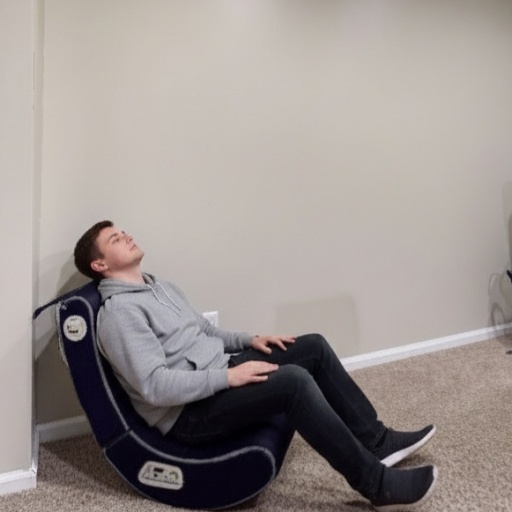}{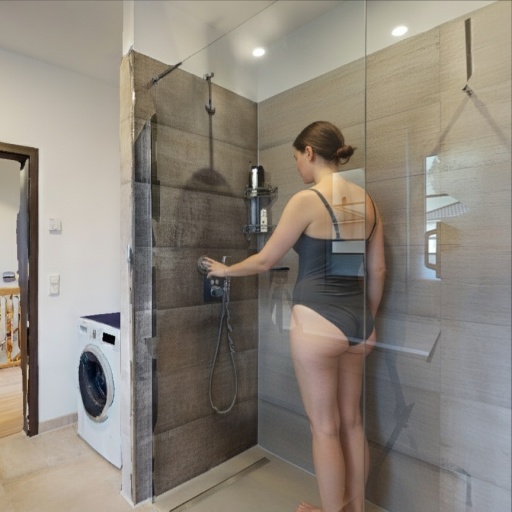}
    \vspace{\suppG}
    \suppdatarow{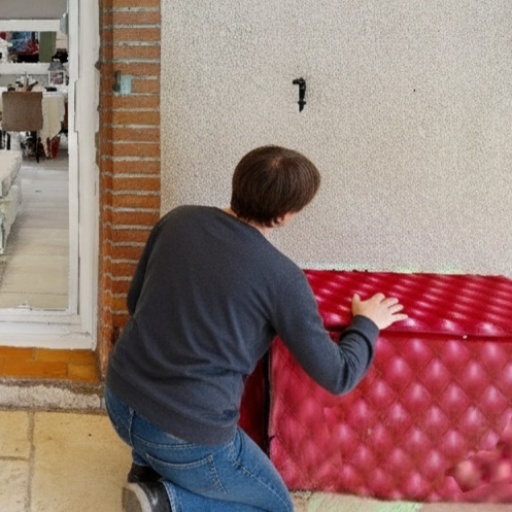}{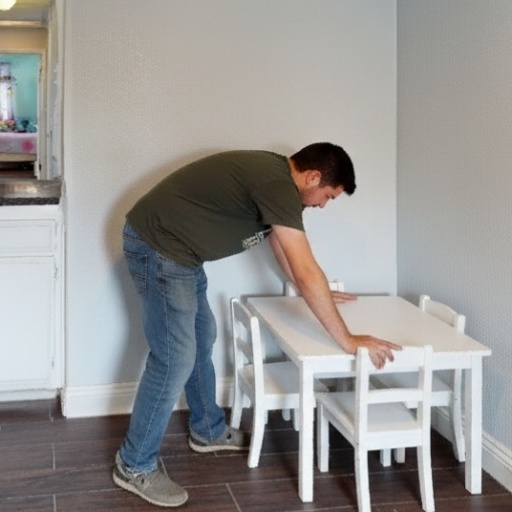}{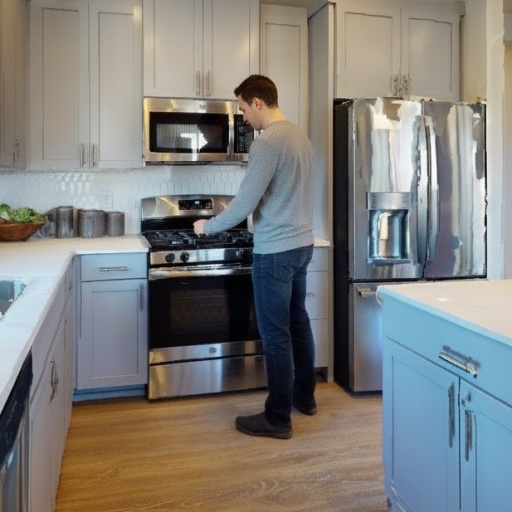}
    \vspace{\suppG}
    \suppdatarow{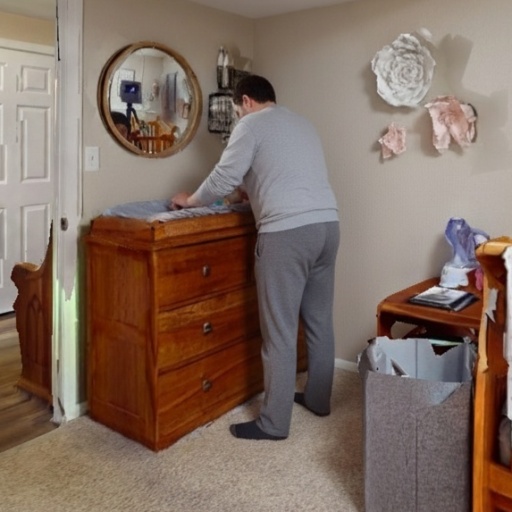}{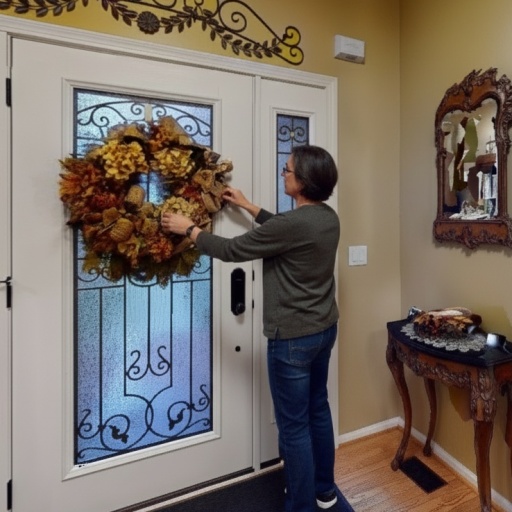}{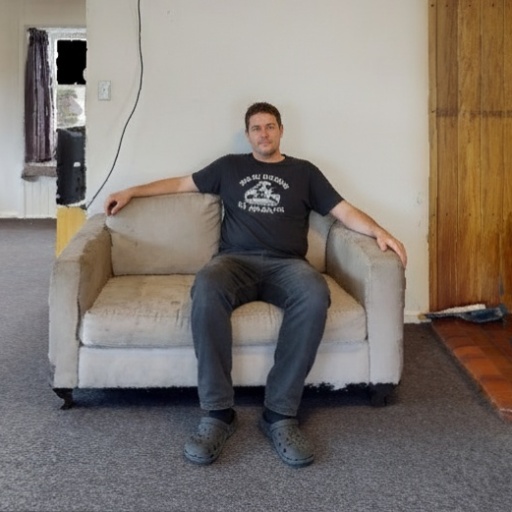}
    \vspace{\suppG}
    \suppdatarow{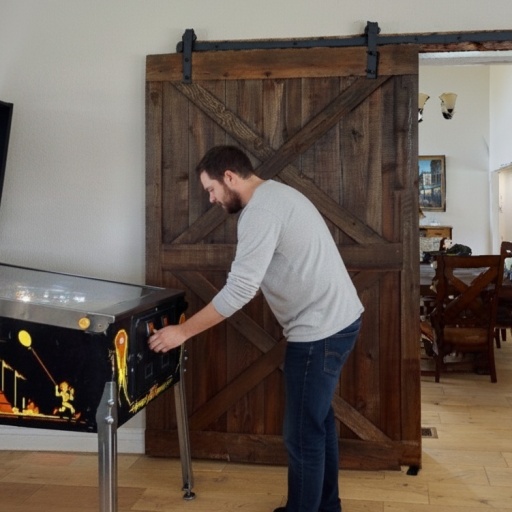}{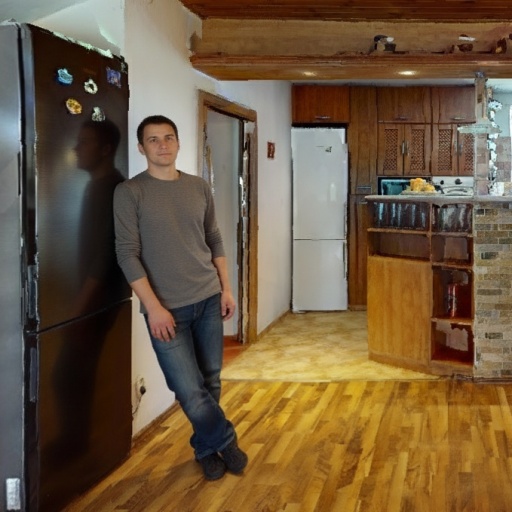}{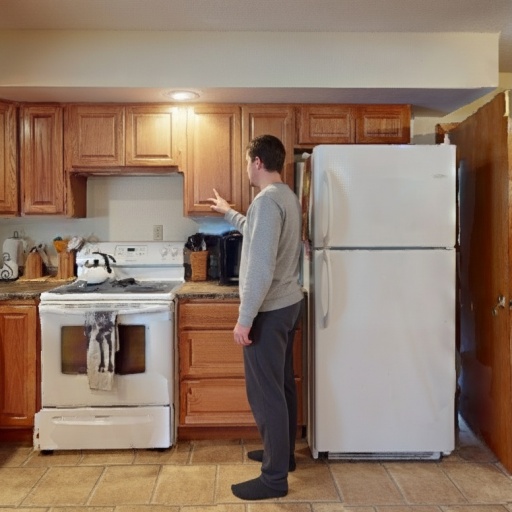}
    \vspace{\suppG}
    \suppdatarow{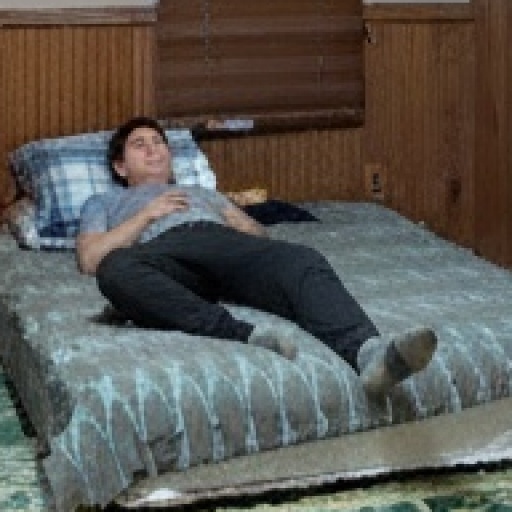}{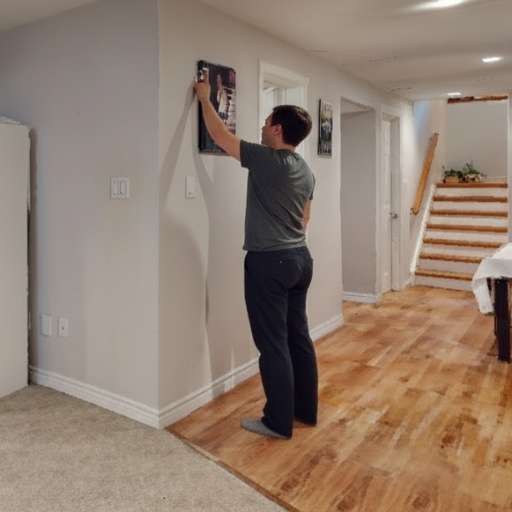}{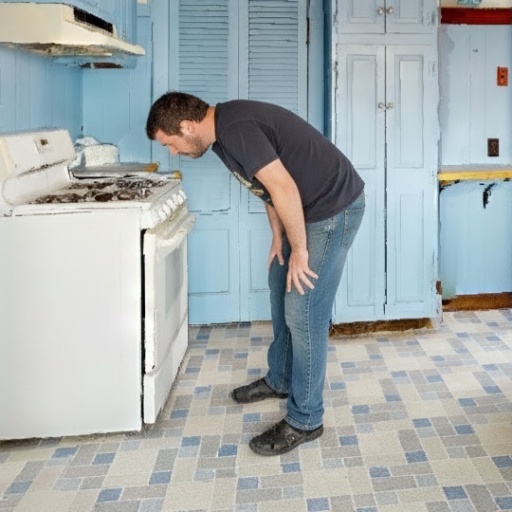}
    \vspace{\suppG}
    \suppdatarow{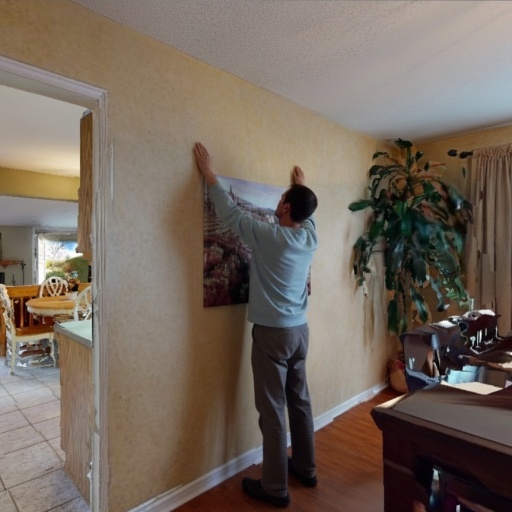}{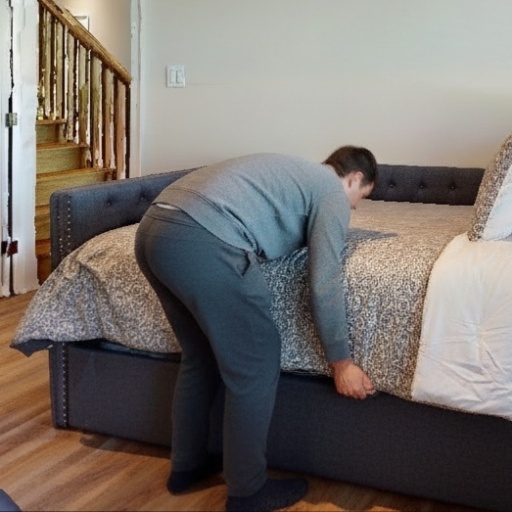}{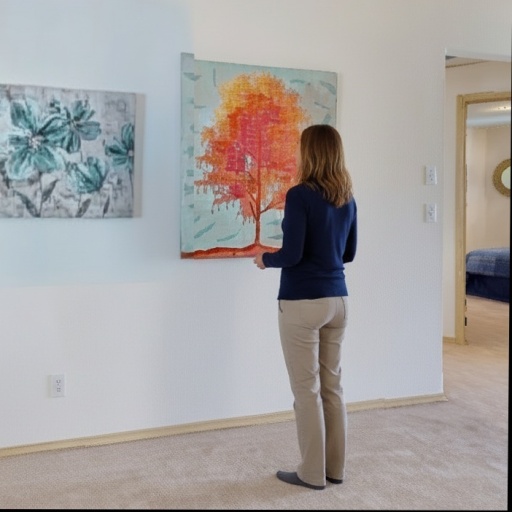}
  \end{minipage}
  \caption{Additional samples from \ourdata{}. Each pair shows the edited RGB image (left) and the corresponding lifted 3D human in the 3D scene (right).}
  \label{fig:suppl:data:2}
\end{figure*}

\begin{figure*}[p]
  \centering
  \begin{minipage}{\textwidth}
    \offinterlineskip
    \suppdatarow{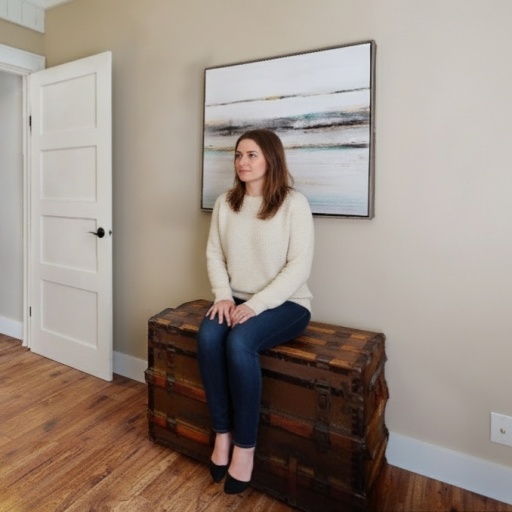}{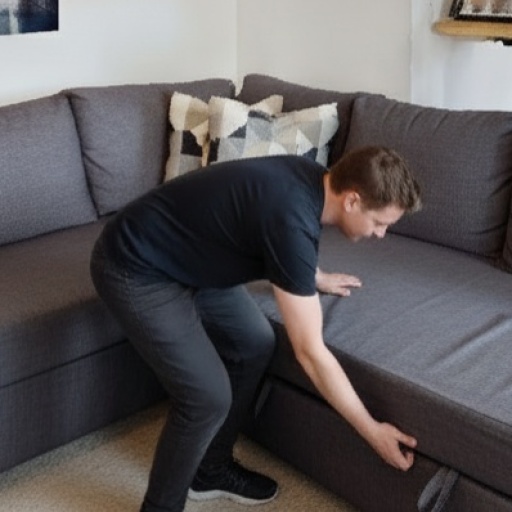}{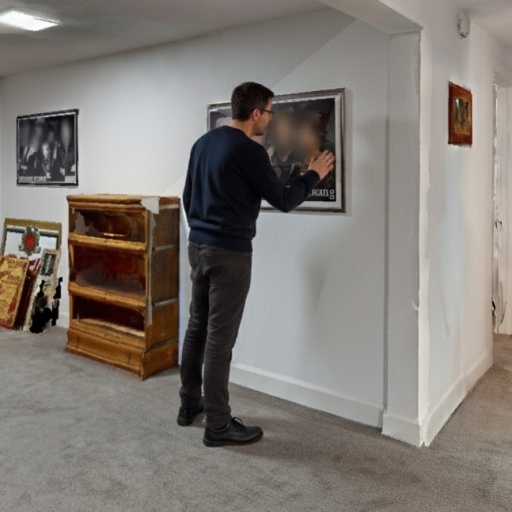}
    \vspace{\suppG}
    \suppdatarow{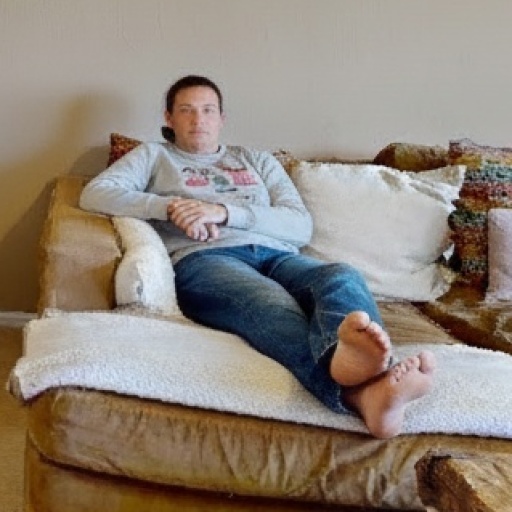}{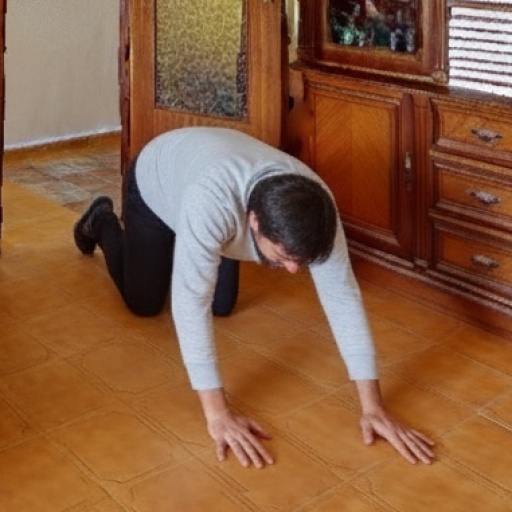}{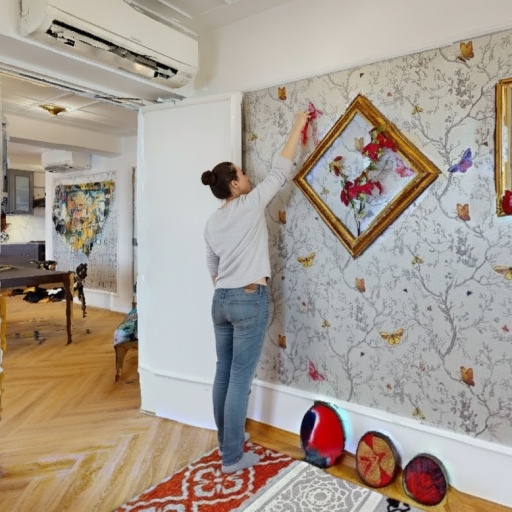}
    \vspace{\suppG}
    \suppdatarow{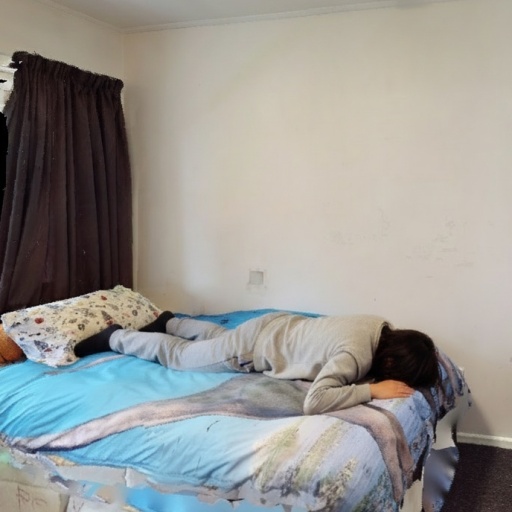}{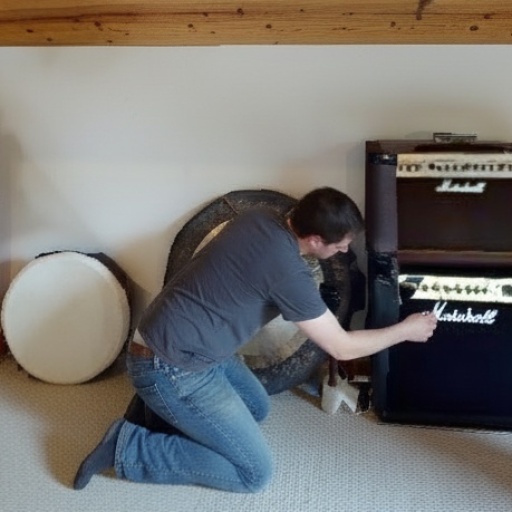}{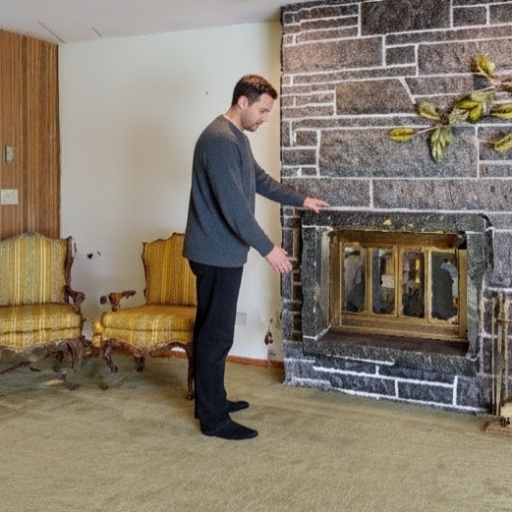}
    \vspace{\suppG}
    \suppdatarow{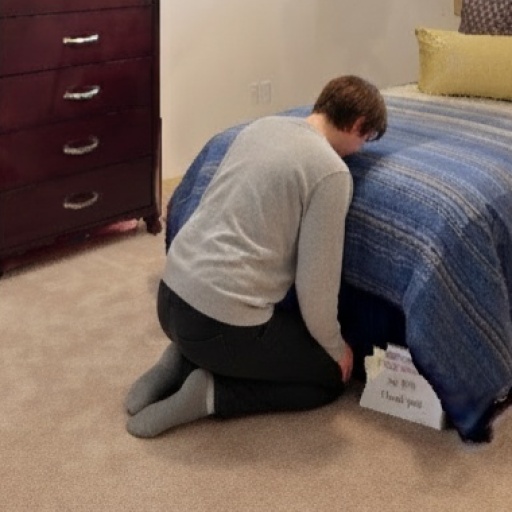}{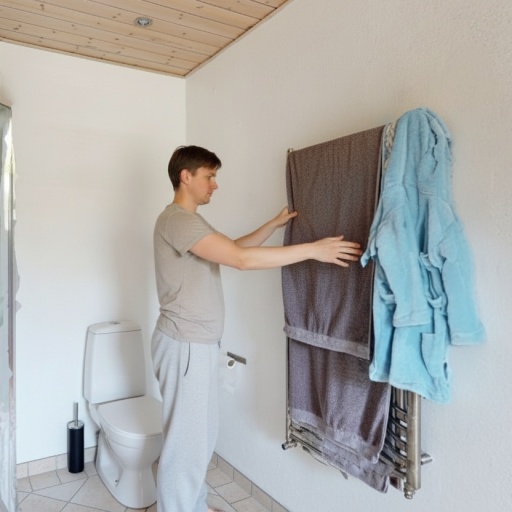}{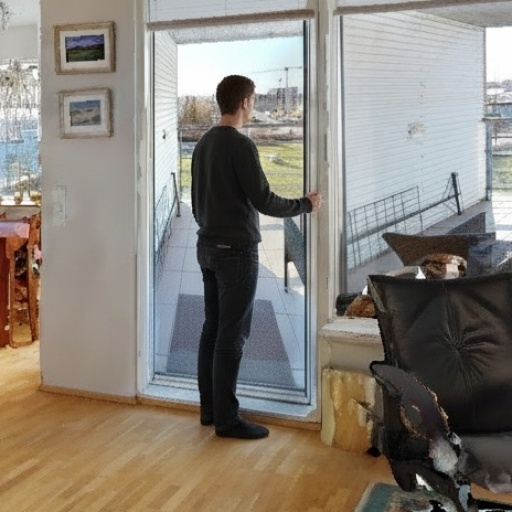}
    \vspace{\suppG}
    \suppdatarow{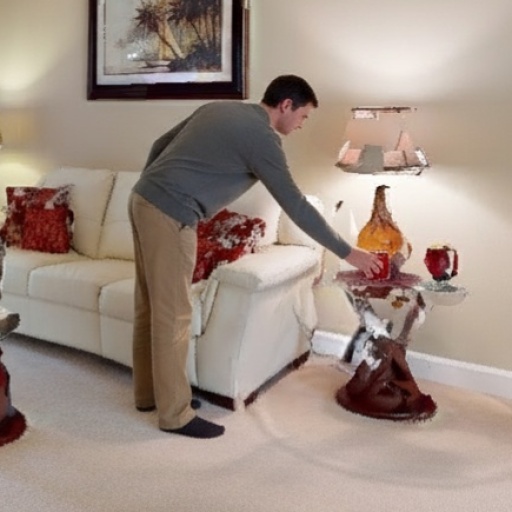}{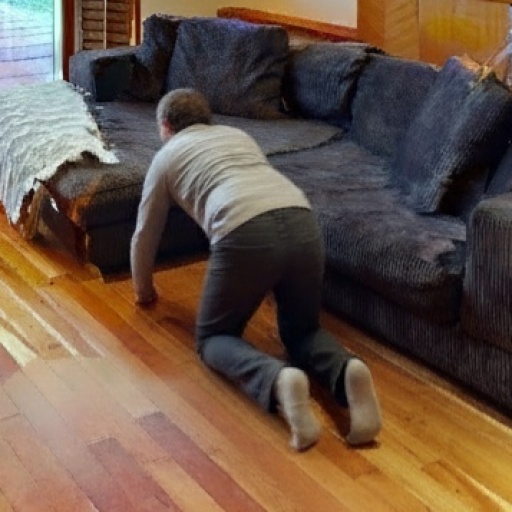}{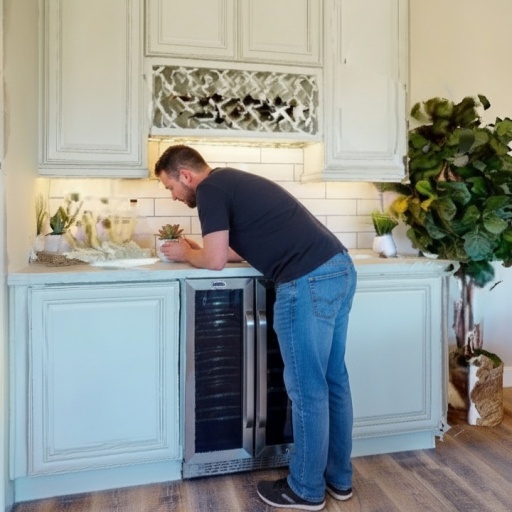}
    \vspace{\suppG}
    \suppdatarow{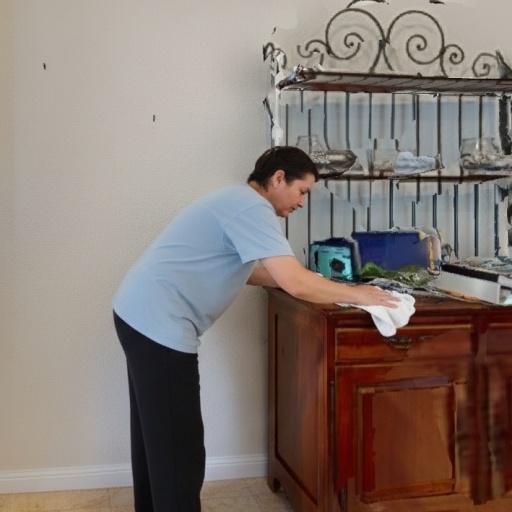}{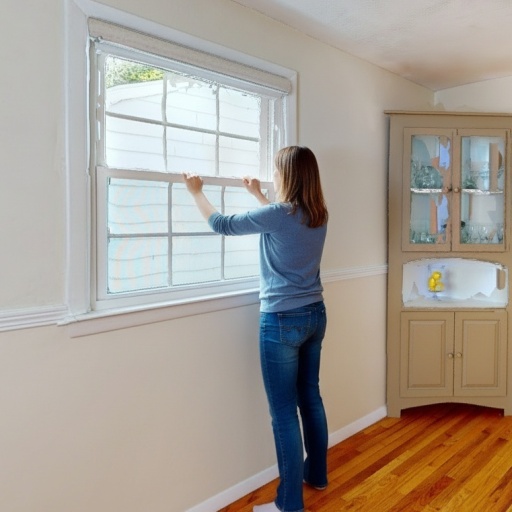}{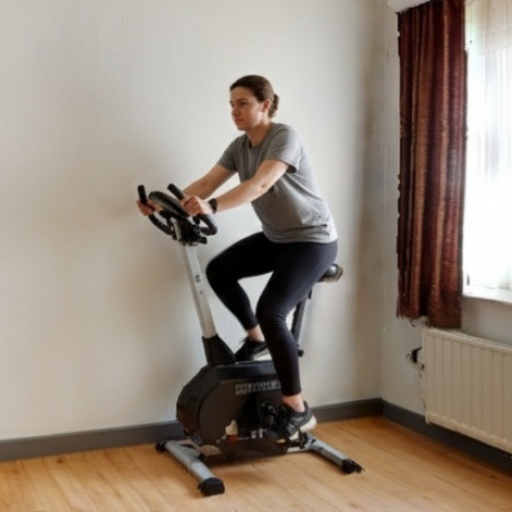}
    \vspace{\suppG}
    \suppdatarow{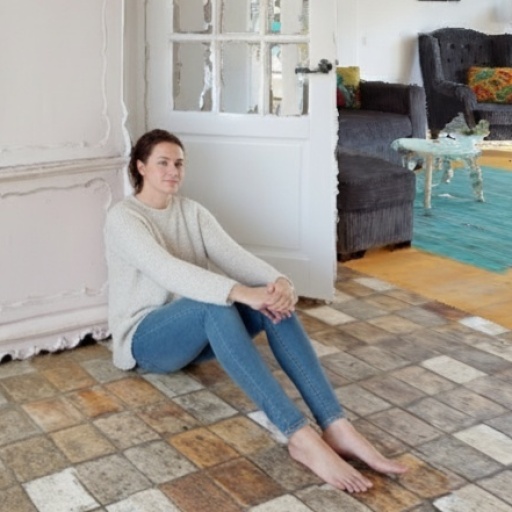}{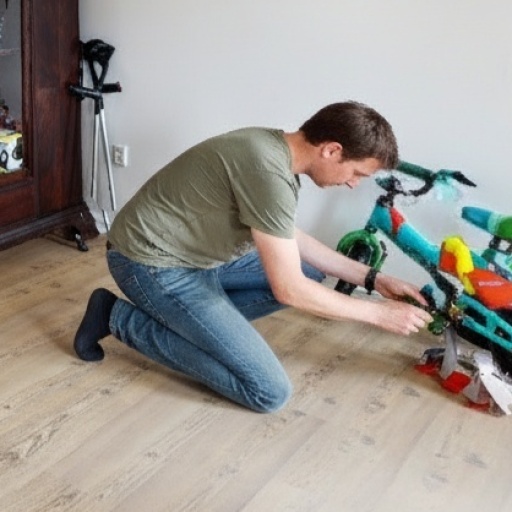}{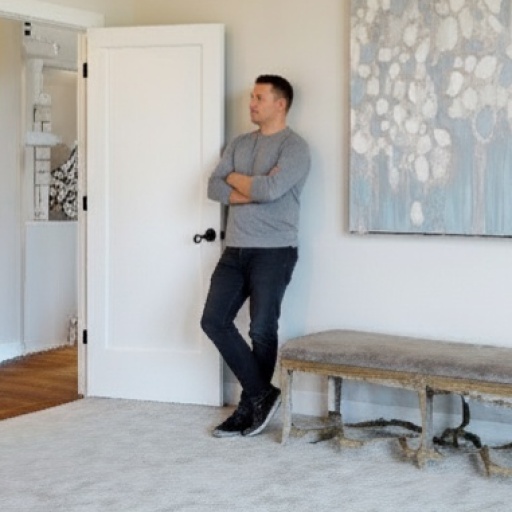}
    \vspace{\suppG}
    \suppdatarow{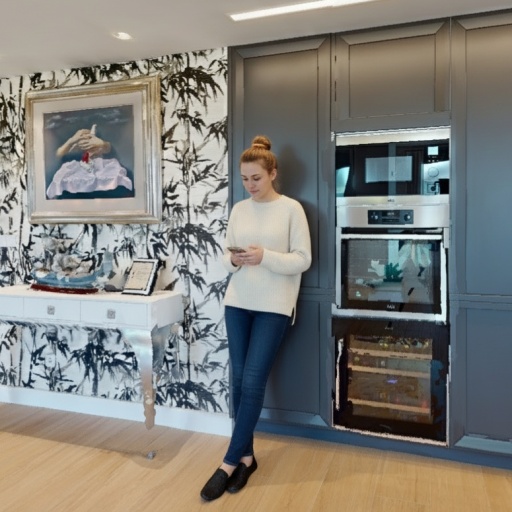}{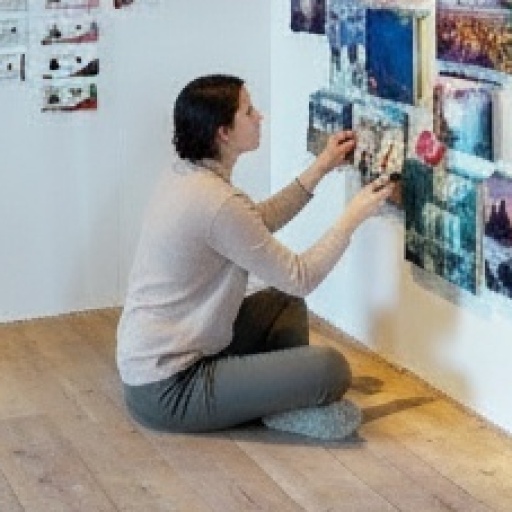}{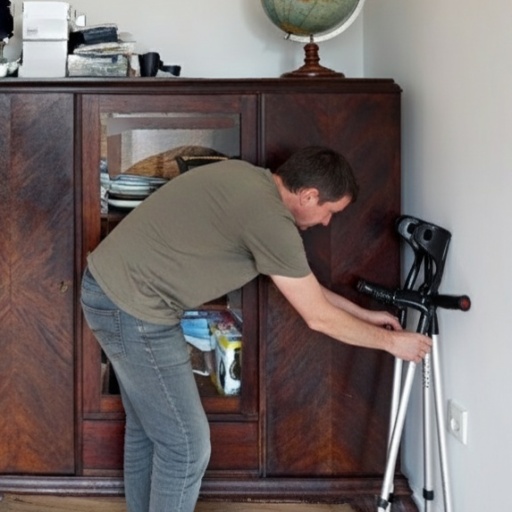}
    \vspace{\suppG}
    \suppdatarow{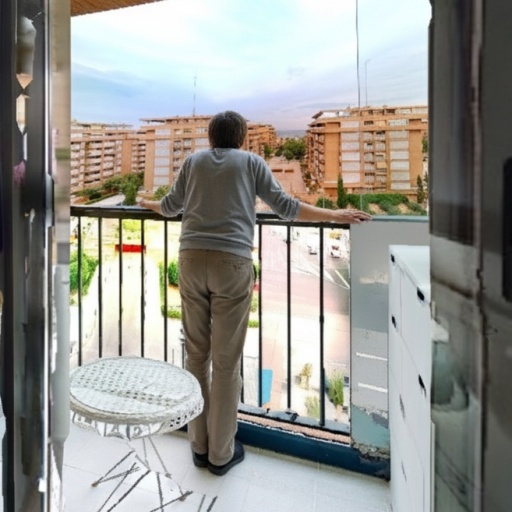}{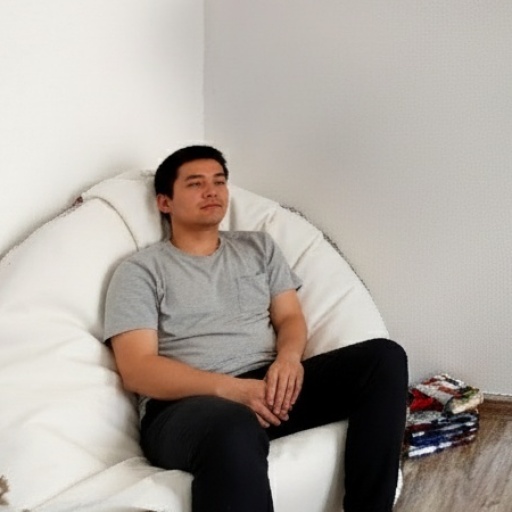}{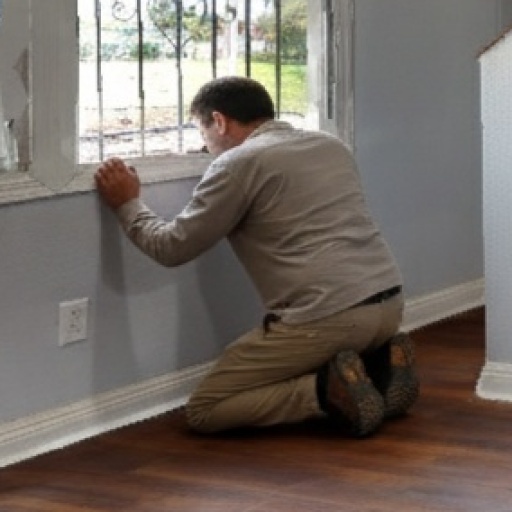}
  \end{minipage}
  \caption{Additional samples from \ourdata{}. Each pair shows the edited RGB image (left) and the corresponding lifted 3D human in the 3D scene (right).}
  \label{fig:suppl:data:3}
\end{figure*}

\end{document}